%% file: main.tex
\documentclass[conference]{IEEEtran}
\IEEEoverridecommandlockouts
\input{header.tex}

\newif\ifshowchanges
\showchangesfalse   

\newcommand{\rev}[1]{%
  \ifshowchanges
    {\color{blue}#1}%
  \else
    #1%
  \fi
}

\definecolor{darkred}{RGB}{190,30,45}
\definecolor{darkblue}{RGB}{2,117,188}
\definecolor{lightbrown}{RGB}{194,181,155}
\definecolor{darkbrown}{RGB}{114,102,88}

\begin{document}

\title{Simulation-Ready Cluttered Scene Estimation via Physics-aware Joint Shape and Pose Optimization}
\author{Wei-Cheng Huang$^1$, Jiaheng Han$^1$, Xiaohan Ye$^2$, Zherong Pan$^3$, and Kris Hauser$^1$%
\thanks{$^1$ Siebel School of Computing and Data Science, University of Illinois at Urbana-Champaign. $^2$ Department of Computer Science, The University of Hong Kong. $^3$ Meta Reality Labs.
}
\thanks{This project was partially supported by NSF Grant \#IIS-1911087 and \#2409661. All experiments, data collection, and processing activities were conducted by the University of Illinois Urbana-Champaign. Meta was involved solely in an advisory role.}
}

\maketitle
\begin{strip}
\centering
\vspace{-1.0in}
\includegraphics[width=\textwidth]{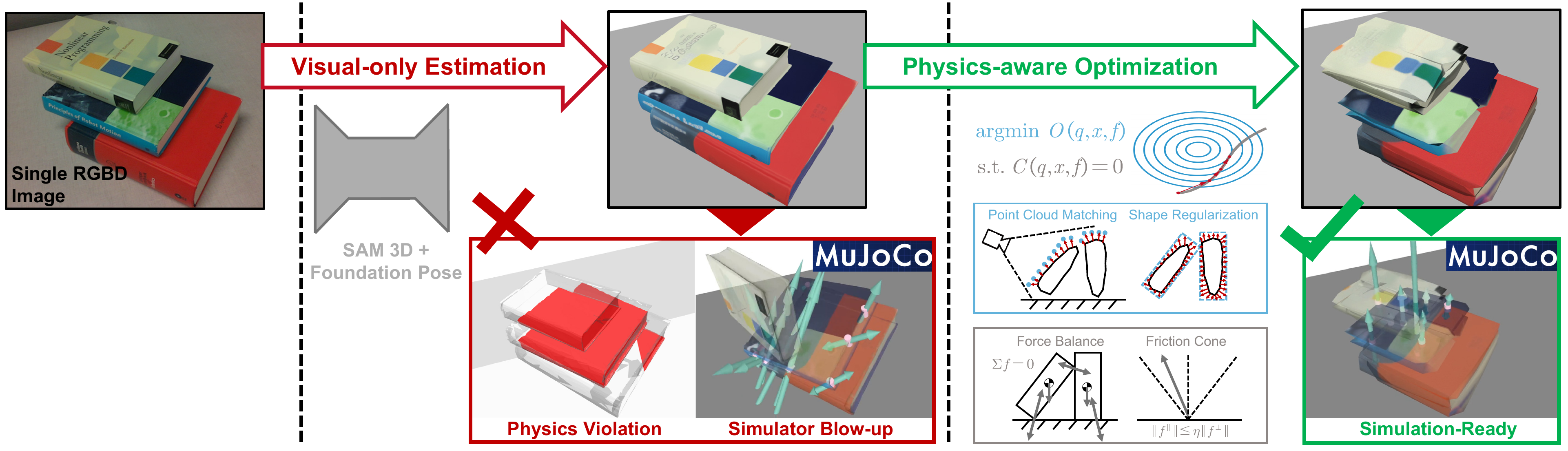}
\vspace{-0.2in}
\captionof{figure}{Given a single RGBD image observation of a cluttered scene, we use SAM3D and FoundationPose to derive an initial estimation of object shapes and poses. But these estimates can violate physical constraints and are not simulation ready (red). Our method jointly adjusts shape and pose parameters to enforce physics constraints while minimizing a perceptual loss, leading to simulation ready results (green).}
\label{fig:teaser}
\end{strip}

\begin{abstract}
Estimating simulation-ready scenes from real-world observations is crucial for downstream planning and policy learning tasks. Regretfully, existing methods struggle in cluttered environments, often exhibiting prohibitive computational cost, poor robustness, and restricted generality when scaling to multiple interacting objects. We propose a unified optimization-based formulation for real-to-sim scene estimation that jointly recovers the shapes and poses of multiple rigid objects under physical constraints. Our method is built on two key technical innovations. First, we leverage the recently introduced shape-differentiable contact model, whose global differentiability permits joint optimization over object geometry and pose while modeling inter-object contacts. Second, we exploit the structured sparsity of the augmented Lagrangian Hessian to derive an efficient linear system solver whose computational cost scales favorably with scene complexity. Building on this formulation, we develop an end-to-end \rev{Simulation-ready Physics-Aware Reconstruction for Cluttered Scenes (SPARCS)} pipeline, which integrates learning-based object initialization, physics-constrained joint shape-pose optimization, and differentiable texture refinement. Experiments on cluttered scenes with up to 5 objects and 22 convex hulls demonstrate that our approach robustly reconstructs physically valid, simulation-ready object shapes and poses. \rev{Project webpage: \href{https://rory-weicheng.github.io/SPARCS/}{\color{cyan}https://rory-weicheng.github.io/SPARCS/}}.
\end{abstract}

\IEEEpeerreviewmaketitle

\input{camera_ready/introduction}
\input{camera_ready/related}
\input{camera_ready/method}
\input{camera_ready/evaluation}
\input{camera_ready/conclusion}
\bibliographystyle{plainnat}
\bibliography{references}

\clearpage
\section*{Appendix}
In this appendix, we present the complete details of our methods, as well as several additional results. First in~\prettyref{appen:optimization_details}, we provide the complete detail of our ALM-based optimizer. Then in~\prettyref{appen:geometry_details}, we present details of our geometric processing algorithms for generating the valid initial guess. Finally, in~\prettyref{appen:extended_results}, we show more examples and baseline comparisons.

\begin{appendices}
\numberwithin{equation}{section} 
\input{camera_ready/Appendix_A_Optimization}
\input{camera_ready/Appendix_B_Preprocess}
\input{camera_ready/Appendix_C_Extended_Results}
\end{appendices}

\end{document}

%% file: header.tex
\usepackage{amsmath,amsfonts,amssymb,amsthm}
\newcommand\numberthis{\addtocounter{equation}{1}\tag{\theequation}}
\let\labelindent\relax
\usepackage{prettyref}
\usepackage{mathrsfs}
\usepackage{graphicx}
\usepackage{wrapfig}
\usepackage{subfig}
\usepackage{bbold}
\usepackage{tabu}
\usepackage{comment}
\usepackage{MnSymbol}
\usepackage{multirow}
\usepackage{booktabs}
\usepackage{enumitem}
\usepackage{algorithm}
\usepackage[table]{xcolor}
\usepackage[noend]{algpseudocode}
\usepackage{times}
\usepackage[numbers]{natbib}
\usepackage{multicol}
\usepackage[bookmarks=true]{hyperref}
\newcommand{\citewithauthor}[1]{\citeauthor{#1} \cite{#1}}
\usepackage{diagbox}
\usepackage{tabularx,booktabs}
\newcolumntype{Y}{>{\centering\arraybackslash}X}
\usepackage{cuted}   
\usepackage{caption} 

\newtheorem{property}{\TE{Property}}[section]

\algnewcommand{\IfThenElse}[3]{
  \State \algorithmicif\ #1\ \algorithmicthen\ #2\ \algorithmicelse\ #3}
  
\algnewcommand{\LineComment}[1]{\State \(\triangleright\) #1}
\algdef{SE}[DOWHILE]{Do}{doWhile}{\algorithmicdo}[1]{\algorithmicwhile\ #1}
\newcommand*{\colorboxed}{}
\def\colorboxed#1#{%
  \colorboxedAux{#1}%
}
\newcommand*{\colorboxedAux}[3]{%
  \begingroup
    \colorlet{cb@saved}{.}%
    \color#1{#2}%
    \boxed{%
      \color{cb@saved}%
      #3%
    }%
  \endgroup
}

\newrefformat{fig}{Figure~\ref{#1}}
\newrefformat{par}{Section~\ref{#1}}
\newrefformat{appen}{Appendix~\ref{#1}}
\newrefformat{sec}{Section~\ref{#1}}
\newrefformat{sub}{Section~\ref{#1}}
\newrefformat{table}{Table~\ref{#1}}
\newrefformat{ass}{Assumption~\ref{#1}}
\newrefformat{alg}{Algorithm~\ref{#1}}
\newrefformat{def}{Definition~\ref{#1}}
\newrefformat{thm}{Theorem~\ref{#1}}
\newrefformat{cor}{Corollary~\ref{#1}}
\newrefformat{lem}{Lemma~\ref{#1}}
\newrefformat{step}{Step~\ref{#1}}
\newrefformat{prop}{Property~\ref{#1}}
\newrefformat{ln}{Line~\ref{#1}}
\newrefformat{rem}{Remark~\ref{#1}}
\newrefformat{eq}{Equation~\ref{#1}}
\newrefformat{pb}{Problem~\ref{#1}}
\newrefformat{prob}{Problem~\ref{#1}}
\newrefformat{it}{Item~\ref{#1}}
\newrefformat{te}{Term~\ref{#1}}
\def\Eqref Eq:#1:{\eqref{eq:#1}}
\newrefformat{Eq}{Equation~\Eqref#1:}

\newcommand{\E}[1]{\mathbf{#1}}
\newcommand{\TE}[1]{\textbf{#1}}

\newcommand{\ResizedEq}[1]{\resizebox{0.85\hsize}{!}{$\begin{aligned}#1\end{aligned}$}}

\newcommand{\FPPR}[2]{{\partial{#1}}/{\partial{#2}}}

\newcommand{\TWO}[2]{\left(\setlength{\arraycolsep}{1pt}\begin{array}{cc}{#1}, & {#2}\end{array}\right)}
\newcommand{\TWOC}[2]{\left(\setlength{\arraycolsep}{1pt}\begin{array}{c}#1 \\ #2\end{array}\right)}

\newcommand{\THREEC}[3]{\left(\setlength{\arraycolsep}{1pt}\begin{array}{c}#1 \\ #2 \\ #3\end{array}\right)}

\newcommand{\FOURR}[4]{\left(\setlength{\arraycolsep}{1pt}\begin{array}{cccc}{#1}^T, & {#2}^T, & {#3}^T, & {#4}^T\end{array}\right)^T}
\newcommand{\FIVE}[5]{\left(\setlength{\arraycolsep}{1pt}\begin{array}{ccccc}{#1}, & {#2}, & {#3}, & {#4}, & {#5}\end{array}\right)}

\newcommand{\MTT}[4]{\left(\setlength{\arraycolsep}{1pt}\begin{array}{cc}#1 & #2 \\ #3 & #4\end{array}\right)}

\newcommand{\argmin}[1]{\underset{#1}{\text{argmin}}\;}

\newcommand{\ST}{\text{s.t.}}

\newcommand{\CH}{\text{CH}}


%% file: camera_ready/introduction.tex
\section{Introduction}
Scene estimation is a fundamental problem in robotics and embodied AI, particularly for real-to-sim transfer. An ideal scene estimator should reconstruct a simulation-ready environment from sparse observations such as images. Beyond perceptual fidelity, the estimated object shapes, poses, and physical properties must be physically consistent and directly usable within a physics simulator, which is critical for downstream tasks such as motion planning, model predictive control, and policy learning. This problem is rather challenging in cluttered scenes, where multiple objects interact through contact, and where accurate physical reasoning is essential for tasks such as robotic manipulation. Over the years, a wide range of scene estimation paradigms have been developed, including Bayesian inference~\cite{deng2021poserbpf}, deep learning~\cite{xiang2017posecnn}, and numerical optimization~\cite{zhang2021sipcc}. Among these, optimization-based scene estimator~\cite{zhang2021sipcc} offers a distinctive advantage for real-to-sim applications: they allow explicit incorporation of physical laws and constraints into the estimation process. By enforcing non-penetration, contact consistency, and equilibrium conditions, physics-based constraints can substantially regularize the solution space and reduce ambiguities.

Despite these advantages, a major challenge for optimization-based state estimators lies in the formulation of physics constraints, which introduces a large number of auxiliary variables, including normal and frictional contact forces as well as Lagrange multipliers. Most prior approaches jointly optimize all variables within a single large-scale nonlinear programming (NLP) formulation using off-the-shelf solvers~\cite{GilMS05,WaechterBiegler2006}. This monolithic strategy leads to computationally expensive problems that scale poorly to cluttered scenes with many interacting objects. To mitigate this complexity, practical methods such as~\cite{zhang2021sipcc} rely on heuristic contact-selection oracles, which are inherently brittle and may fail when contacts are missed. More fundamentally, to keep computational costs tractable, existing approaches assume known object geometries and restrict optimization to object poses. In contrast, scene estimation from sparse observations inherently requires jointly inferring both object shapes and poses, dramatically increasing the dimensionality of the decision space. The resulting proliferation of shape parameters renders existing optimization-based techniques computationally prohibitive and, in practice, intractable.

We attribute these limitations to a common root cause: existing optimization-based state estimators are not structure-aware~\citep{zhang2021sipcc,zhang2025simultaneous}. By formulating all variables within a monolithic NLP, they fail to exploit the structure of physics-constrained optimization. We therefore propose a structure-aware framework for real-to-sim scene estimation in cluttered environments---the first practical algorithm for numerical optimization in the joint shape-pose space. Our approach builds on the separating-plane-based shape-differentiable contact model (SDRS)~\citep{ye2025sdrs}, which eliminates normal contact forces as explicit variables by expressing them as functions of object pose. We adapt SDRS to quasistatic configuration optimization, reducing problem dimensionality, and show that the resulting augmented Lagrangian Hessian has a highly structured sparsity pattern that enables efficient solvers via Woodbury and Schur complement reductions. The proposed formulation is globally differentiable with respect to both shape and pose, enabling joint optimization under arbitrary contact conditions with shapes represented as unions of convex hulls. To improve robustness, we consider all potential contact pairs without heuristic contact selection~\cite{zhang2021sipcc}, while maintaining tractable computational cost through our reduced representation. A globally supported contact activation function~\cite{Ye2025Efficient} further mitigates vanishing gradients and constraint-qualification issues.

Building upon our joint optimization framework, we develop \rev{\textbf{S}imulation-ready \textbf{P}hysics-\textbf{A}ware \textbf{R}econstruction for \textbf{C}luttered \textbf{S}cenes, \textbf{SPARCS}}, that operates directly on a single RGBD image observation as illustrated in~\prettyref{fig:teaser}. 
We evaluate our method on a diverse set of cluttered benchmarks containing up to $5$ objects and $22$ convex hulls, where our method can robustly produce physically valid, simulation-ready reconstructions.

%% file: camera_ready/related.tex
\section{Related Work}
In this section, we review prior work on state estimation and \rev{static} scene understanding, focusing on their relevance to \rev{real-to-sim transfer for cluttered environments, physics-aware optimization, and differentiable physics.}

\paragraph{\rev{Sim-Ready} Scene Estimation}
Early scene estimation methods typically assume known object geometries and focus on recovering object poses from partial observations. Classical approaches~\cite{chetverikov2002trimmed,segal2009generalized} formulate rigid-body registration as geometric alignment, with later extensions to non-rigid~\cite{amberg2007optimal} and articulated~\cite{chang2008automatic} models. While geometrically well founded, these methods are brittle under occlusion and missing data, which are ubiquitous in cluttered real-world scenes. Learning-based approaches, such as PoseCNN~\cite{xiang2017posecnn} and FoundationPose~\cite{wen2024foundationpose}, improve robustness by leveraging learned priors, but remain purely perception-driven. \rev{With the growing interest in real-to-sim transfer for scaling up data for robotics, simulation-ready estimation shows increasing importance nowadays. Specifically for static scene, a sim-ready estimate requires the result be seamlessly imported into a physics simulator and remain at a stable configuration~\cite{song2018inferring, ni2024phyrecon}. Purely perception-driven approaches are ill-suited for such requirement due to the lack of explicit physics awareness or constraints}. More recent works~\cite{song2018inferring,ni2024phyrecon,bianchini2025vysics,yao2025cast,xia2025holoscene} incorporate physical reasoning through sampling-based optimization or physics-violation loss terms, but remain limited in scope: they assume a small set of fixed hypothesized object shapes~\cite{song2018inferring}, optimize the shape of only a single object~\cite{bianchini2025vysics}, require dense observation such as RGB video~\cite{ni2024phyrecon,xia2025holoscene}, or model collision constraints without enforcing full physical consistency~\cite{yao2025cast}.

\paragraph{Physics-aware Numerical Optimization}
To improve physical validity, prior work has integrated physics constraints into numerical optimization. Methods such as PhysPose~\cite{malenicky2025physpose} and Verefine~\cite{bauer2020verefine} encourage physically plausible configurations via penalties or post hoc simulation checks, but do not provide full physics-consistent optimization. A more principled approach~\cite{zhang2021sipcc} enforces non-penetration and force equilibrium constraints directly, but relies on heuristic contact-selection oracles that degrade robustness in cluttered scenes. Extensions to deformable objects~\cite{Hsu2022} further expand the scope of physics-aware estimation. Nevertheless, most existing methods assume known object geometry and focus exclusively on pose estimation, making them unsuitable for real-to-sim transfer where object shapes must also be recovered. We are aware of one prior work~\cite{bianchini2025vysics} incorporating both shape and pose reasoning but limited to a single object.

\paragraph{Differentiable Simulation \& Rendering}
Differentiable simulators and renderers~\cite{kato2020differentiable,newbury2024review} enable gradient-based optimization of physically grounded scenes and have been applied to large-scale state estimation~\cite{Xu-RSS-21,ma2021risp,zhu2025one}. However, most differentiable simulators target dynamic trajectories rather than quasistatic, force-balanced configurations typical of cluttered scenes. Moreover, non-smooth contact and visibility events violate the smoothness assumptions of NLP solvers. We build upon the globally differentiable SDRS contact model~\cite{ye2025sdrs} \rev{originally developed for dynamic simulation,} and reformulate it for quasistatic equilibrium, enabling smooth and scalable joint optimization of object shape and pose. This design directly supports simulation-ready reconstruction of cluttered real-world scenes for real-to-sim transfer.

%% file: camera_ready/method.tex
\section{\label{sec: methods}Methods}
In this section, we present our complete pipeline for simulation-ready cluttered scene reconstruction. We then focus on the details of our optimization-based scene estimator for a set of rigid bodies in the joint pose- and shape-space. 

\subsection{Problem Statement}
Following~\cite{zhang2021sipcc}, we express the problem as an equality-constrained NLP of the following form:
\begin{equation}
\begin{aligned}
\label{prob:NLP}
\argmin{q,x}\;O(q,x)\quad
\ST\;C(q,x)=0,
\end{aligned}
\end{equation}
and we propose a structured variant of Augmented Lagrangian Method (ALM) to find locally optimal solutions. We will omit function parameters below when confusion is unlikely. Here, $q$ and $x$ denote the vectors describing the poses and shapes of all objects, respectively, and the physics constraints are modeled as the globally differentiable equality constraint $C(q,x)=0$. The objective $O(q,x)$ is assumed to be an arbitrary globally differentiable function, allowing flexibility for various state-estimation tasks. For instance, $O$ could represent a differentiably rendered image-space loss~\cite{liu2019softras} or a point cloud registration loss~\cite{Aiger2008GlobalRegistration}. 

\begin{figure}[h]
\centering
\includegraphics[width=0.7\linewidth]{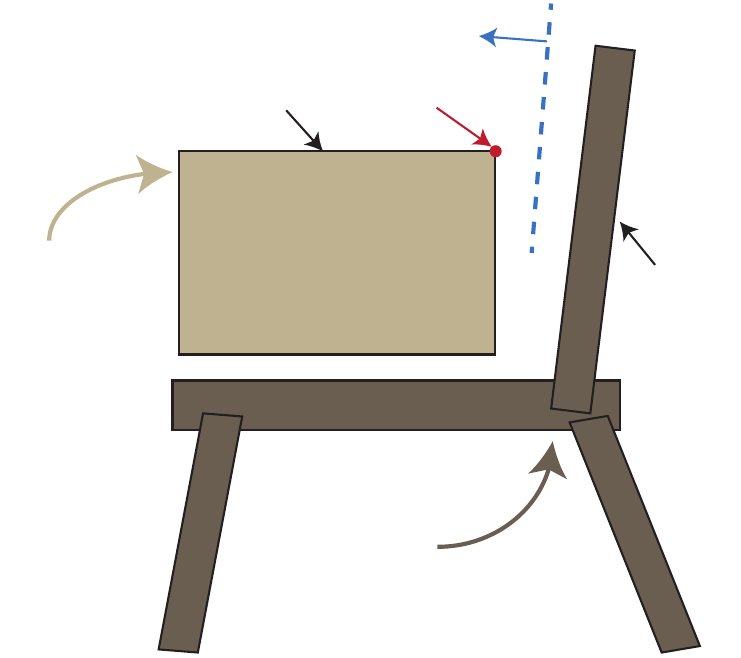}
\put(-135,130){\small$\text{Hull}_{ij}$}
\put(-25,85){\small$\text{Hull}_{i'j'}$}
\put(-95,130){\textcolor{darkred}{\small$X_{ijk}$}}
\put(-105,148){\textcolor{darkblue}{\small$\TWO{n}{d}_{iji'j'}$}}
\put(-175,90){\textcolor{lightbrown}{\small$\text{Body}_i$}}
\put(-100,25){\textcolor{darkbrown}{\small$\text{Body}_{i'}$}}
\caption{\label{fig:representation}Suppose we would like to model a box (light brown) put on a chair (dark brown). The box and the chair are the $i$th and $i'$th rigid bodies respectively, where the box is modeled as a single convex hull and the chair is modeled as the union of 4 convex hulls. Each convex hull is a polytope spanned by a set of vertices $X_{ijk}$ (red). Between the $ij$th convex hull on the box and the $i'j'$th convex hull on the chair modeling the back support, we introduce a separating plane $\TWO{n}{d}_{iji'j'}$ (blue) as a proxy for the contact model.}
\end{figure}

Our formulation builds on the recently proposed convex-hull-based contact model~\cite{ye2025sdrs}---a provably second-order differentiable contact model, enabling globally twice-differentiable operations under arbitrary changes in convex hull geometry. As illustrated in~\prettyref{fig:representation}, we represent a scene with $N$ rigid bodies, each modeled as a union of $M$ convex hulls, with each hull having $V$ vertices. Specifically, we denote $x_{ijk}\in\mathbb{R}^3$ as the $k$th vertex of the $j$th convex hull belonging to the $i$th rigid body, expressed in the body frame. Concatenating these vertices, we obtain $x\in\mathbb{R}^{N\times M\times V\times 3}$, which describes the shapes of all rigid bodies. To represent the poses of these bodies, we define the local-to-global transformation of the $i$th rigid body using a rotation matrix $R_i(\theta_i)=\exp[\theta_i]_\times$ and a translation vector $t_i\in\mathbb{R}^3$, where the rotation is parameterized via the Rodrigues formula with $\theta_i\in\mathbb{R}^3$ and $[\bullet]_\times$ being the cross-product matrix. Concatenating all $\theta_i$ and $t_i$ yields the vector $q$, which encodes the poses of all rigid bodies. It is known that the mapping $R_i(\theta_i)$ is smooth~\cite{hilgert2011structure}. The local-to-global transformation is defined as $X_{ijk}=R_ix_{ijk}+t_i$ and we omit the function parameters when the context is clear.

\input{camera_ready/pipeline}

\subsection{Optimization in Joint Shape-Pose Space}
It is well-known~\cite{solodov2009global} that the key requirement for finding a locally optimal and feasible solution of~\prettyref{prob:NLP} is that $C$ satisfies the Linear Independent Constraint Qualification (LICQ), i.e., \rev{the minimal eigenvalue of $\nabla C\nabla C^T$} is bounded away from zero for any $q$ and $x$. In practice, we use a custom version of ALM that iteratively minimizes the following augmented Lagrangian function, where we use Levenberg-Marquardt (LM) algorithm as the sub-problem:
\begin{equation}
\label{prob:ALM-subproblem}
\argmin{q,x}\;O(q,x)+\lambda^TC(q,x)+\frac{\rho}{2}\|C(q,x)\|^2,
\end{equation}
and we iteratively increase the Lagrangian multiplier $\lambda$ and the penalty coefficient $\rho$ according to~\cite{nocedal2006numerical}.

Our formulation in~\prettyref{prob:NLP} differs from prior work in several key ways. First, unlike complementarity constraint-based formulations~\cite{zhang2021sipcc,posa2014direct}, we eliminate auxiliary variables for normal contact forces, substantially reducing problem dimensionality and subproblem cost. Second, the formulation involves only equality constraints, so each subproblem reduces to solving a linear system rather than a general quadratic program, yielding significant speedups. We first analyze this frictionless case, then introduce friction as additional decision variables and present structured linear solvers to handle them efficiently.

\subsubsection{Physics Constraints Without Friction}
Let us consider the case where all objects experience only normal contact forces, without friction. Using the representation introduced in the previous section, the physics constraints reduce to minimizing the internal and external potential energy at the resting pose. The potential energy $\Psi(q,x)$ consists of two terms: the gravitational potential $\Psi_g$ and the collision potential $\Psi_c$, such that $\Psi = \Psi_g + \Psi_c$. The gravitational potential is formulated under the assumption that the mass is concentrated at the convex hull vertices~\cite{ye2025sdrs}, since a uniform mass distribution would be non-differentiable with respect to $x$. Under this assumption, we define:
\begin{align}
\Psi_g(q,x)=-\rho \sum_{i,j,k}\left<X_{ijk}, g\right>,
\end{align}
where $\rho$ and $g$ denote the density and gravitational acceleration, respectively. Clearly, $\Psi_g$ is twice-differentiable with all variables, which allow users to optimize object density and mass distributions. We then define the constraint as $C = \nabla_q \Psi$. If $\Psi$ is globally twice-differentiable, then $C$ is differentiable and satisfies the smoothness requirement.

Our second term is the collision potential. Compared with hard constraints for modeling collision~\cite{posa2014direct,zhang2021sipcc}, our method follows recent findings that using a primal-only, instead of primal-dual, interior-point formulation~\cite{Li2020IPC} provides a more robust approach. This is because interior-point methods maintain satisfaction of collision constraints, avoiding the vanishing gradient issue that can occur under deep penetration. \citewithauthor{ye2025sdrs} generalized the idea of~\cite{Li2020IPC} by defining a potential function between two general convex hulls. Specifically, two convex hulls are intersection-free if and only if there exists a separating plane between them~\cite{rockafellar-1970a}. Accordingly, we introduce the separating plane as an additional set of variables $\TWO{n}{d}_{iji'j'} \in \mathbb{R}^4$, which separates the $ij$th and $i'j'$th convex hulls. Here, $n_{iji'j'}$ is the plane normal and $d_{iji'j'}$ is the plane offset. The collision potential between the two convex hulls is then defined as $\bar{\Psi}_{iji'j'}(q,x,n_{iji'j'},d_{iji'j'})$:
\begin{equation}
\label{eq:normal}
\bar{\Psi}_{iji'j'}=
\begin{cases}
-\log(1-\|n_{iji'j'}\|)\\
\sum_k-\log(\left<n_{iji'j'},X_{ijk}\right>+d_{iji'j'})\\
\sum_{k'}-\log(-\left<n_{iji'j'},X_{i'j'k'}\right>-d_{iji'j'}),
\end{cases}
\end{equation}
The first term ensures that the plane normal has a magnitude no greater than $1$, while the second and third terms ensure that the two convex objects lie on opposite sides of the separating plane, thereby enforcing collision-free constraints. It has been shown that $\bar{\Psi}_{iji'j'}$ is well-defined and globally twice-differentiable. Although the definition of $\bar{\Psi}_{iji'j'}$ introduces the separating-plane variables $\TWO{n}{d}_{iji'j'}$, the function is strictly convex with respect to these variables. Therefore, we can eliminate them implicitly and define:
\begin{align}
\Psi_{iji'j'}(q,x) = \min_{n_{iji'j'},d_{iji'j'}} \bar{\Psi}_{iji'j'}(q,x,n_{iji'j'},d_{iji'j'}),
\end{align}
By the implicit function theorem, $\Psi_{iji'j'}$ remains globally twice-differentiable. Finally, the full collision potential is defined as $\Psi_c = \mu \sum_{i\neq i'} \sum_{j,j'} \Psi_{iji'j'}$, where $\mu$ is the complementarity gap. As shown in~\cite{Li2020IPC}, as $\mu \to 0$, $\Psi_c$ approximates hard collision-free constraints arbitrarily well. Further, the optimality of $\bar{\Psi}_{iji'j'}$ with respect to the separating plane $\TWO{n}{d}_{iji'j'}$ ensures that equal and opposite forces are applied on the two convex hulls, which essentially satisfy the Newton's third law. Finally, the function $\Psi=\infty$ when collision constraints are violated, in which case the solution will be rejected by the ALM subproblem solver to use a smaller search step size, thus ensuring the collision constraints are satisfied at every iteration. To improve efficiency, a clamped $\log$ function can be used to make the potential locally supported, so that $\Psi_{iji'j'}$ evaluates to zero when objects are far apart. In practice, thanks to the compact convex-hull representation, computing all pairs of collisions is sufficiently efficient, allowing us to use a globally supported $\log$ function. The global support is beneficial because it ensures non-zero gradients even for distant objects, which helps satisfy the LICQ condition. We summarize the well-behaved properties of our physics constraints below, which is formally proved in~\cite{ye2025sdrs}:
\begin{property}
Under frictionless setting, $C(q,x)$ is globally differentiable with respect to both $q$ and $x$. When $C(q,x)=0$, all objects are force and torque balanced while satisfying the Newton's third law.
\end{property}

\begin{figure}[h]
\centering
\includegraphics[width=0.7\linewidth]{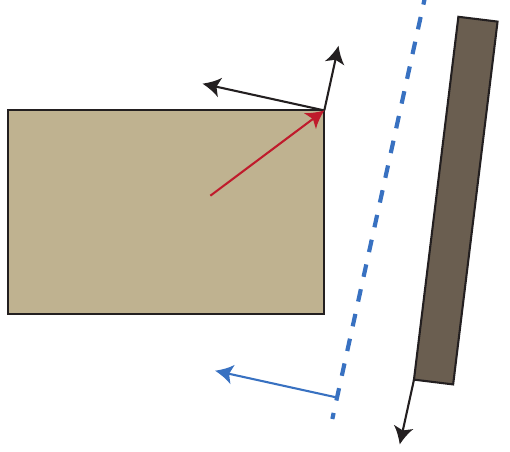}
\put(-115,80){\textcolor{darkred}{\small$X_{ijk}$}}
\put(-120,10){\textcolor{darkblue}{\small$\TWO{n}{d}_{iji'j'}$}}
\put(-130,135){\small$f^\perp_{ijk,i'j'}(q,x)$}
\put(-70,148){\small$f^\parallel_{ijk,i'j'}$}
\put(-30,5){\small$f^\parallel_{i'j'k',ij}$}
\caption{\label{fig:friction-representation}An illustration of our friction model, between the $ij$th and $i'j'$th convex hull. On each vertex, e.g. $X_{ijk}$, the normal force $f_{ijk,i'j'}^\perp$ is a function of $x$ and $q$ (\prettyref{eq:perp}) and the friction force $f_{ijk,i'j'}^\parallel$ is additional decision variables to be optimized. We follow the idea of SDRS contact model and use the separating plane as the proxy for contact modeling and each force $f_{ijk,i'j'}^\parallel$ applied on the $ij$th convex hull is counter-acted by $-f_{ijk,i'j'}^\parallel$ applied on the separating plane $\TWO{n}{d}_{iji'j'}$, and the case is the same for the $i'j'$th convex hull. We then model the separating plane as a physical object with zero mass, so that all the forces applied on it must be balanced.}
\end{figure}
\subsubsection{Incorporating Frictional Contacts}
Frictional contacts are more involved, since friction forces always dissipate kinetic energy and therefore do not admit a corresponding potential energy formulation. In this section, we first formulate our frictional contact model and then present our structure-aware algorithm for solving the ALM subproblem~\prettyref{prob:ALM-subproblem}. As illustrated in~\prettyref{fig:friction-representation}, we consider a pair of convex hulls. For each vertex $x_{ijk}$ (the case for $x_{i'j'k'}$ is symmetric), the normal contact force is given by:
\begin{align}
\label{eq:perp}
f_{ijk,i'j'}^\perp=\FPPR{\Psi_{iji'j'}}{X_{ijk}}.
\end{align}
In addition, by the friction cone constraint, the vertex may also experience tangential frictional forces $f_{ijk,i'j'}^\parallel$, which satisfy:
\begin{align}
\label{eq:cone-constraint}
\left<f_{ijk,i'j'}^\perp,f_{ijk,i'j'}^\parallel\right>=0\quad
\|f_{ijk,i'j'}^\parallel\|\leq\eta\|f_{ijk,i'j'}^\perp\|,
\end{align}
where $\eta$ is the friction coefficient of the cone. However, the friction cone constraints alone are not sufficient. We must also ensure that the collection of frictional forces applied on the two convex hulls satisfies both force and torque balance. To achieve this, we adopt the idea of~\cite{ye2025sdrs}, in which the separating plane is treated as a fictitious physical object with zero mass. While the separating plane has no direct physical impact on the system, it serves as a proxy for formulating and enforcing the balance of frictional forces. Concretely, for a frictional force applied on $x_{ijk}$, an opposing force is applied to the plane, resulting in an in-plane force of $-f_{ijk,i'j'}^\parallel$ and an in-plane torque of $-T_{iji'j'} X_{ijk}\times f_{ijk,i'j'}^\parallel$, where $T_{iji'j'}$ denotes the projection operator onto the plane normal space
and $n_{iji'j'}$ is the (unnormalized) separating plane normal, as defined in the normal collision potential (\prettyref{eq:normal}). Finally, since the separating plane has zero mass, it must remain in equilibrium to avoid unbounded accelerations. This yields the following in-plane force and torque balance conditions:
\begin{equation}
\begin{aligned}
\label{eq:balance-constraint}
&\sum_{k}f_{ijk,i'j'}^\parallel+\sum_{k'}f_{i'j'k',ij}^\parallel=0,\\
&T_{iji'j'}\left[\sum_{k}X_{ijk}\times f_{ijk,i'j'}^\parallel+
\sum_{k'}X_{i'j'k'}\times f_{i'j'k',ij}^\parallel\right]=0.
\end{aligned}
\end{equation}
Finally, we incorporate the frictional contact forces by defining the augmented potential energy and constraint function $\bar{C}$:
\begin{align}
\ResizedEq{\bar{C}(q,x,f_{iji'j'}^\parallel)=\nabla_q\left[\Psi(q,x)-\sum_{i\neq i'}\sum_{j,j'}\sum_k\langle X_{ijk},f_{ijk,i'j'}^\parallel\rangle\right].}
\end{align}
Again, we summarize the well-behaved properties of our augmented physical constraints below:
\begin{property}
When $\mu>0$, $\bar{C}(q,x,f_{iji'j'}^\parallel)$ is globally differentiable with respect to both $q$ and $x$. When $\bar{C}(q,x,f_{iji'j'}^\parallel)=0$, all objects are force and in-plane torque balanced while satisfying the Newton's third law.
\end{property}
We emphasize that an important difference from the frictionaless case is that we can only ensure torque balance for the components perpendicular to the plane normal~\cite{ye2025sdrs}. Unfortunately, the torque component along the plane normal direction is not balanced in general. But the violation to torque balance along this component is controlled by the complementarity gap $\mu$. By selecting a small value of $\mu$, the violation is negligible in practice.

Although the frictional contact forces are straightforward to formulate, incorporating these constraints would significantly increase the dimensions of the decision space. Indeed, we need to optimize a pair of per-vertex frictional forces $f_{iji'j'}^\parallel$ between each pair of rigid bodies, leading to the following constrained optimization:
\begin{align*}
&\argmin{q,x,f_{iji'j'}^\parallel}\;O(q,x)\numberthis\label{prob:ALM-friction}\\
&\ST\;
\begin{cases}
\bar{C}(q,x,f_{iji'j'}^\parallel)=0\\
\left<f_{ijk,i'j'}^\perp,f_{ijk,i'j'}^\parallel\right>=0\\
\|f_{ijk,i'j'}^\parallel\|\leq\eta\|f_{ijk,i'j'}^\perp\|\\
\sum_{k}f_{ijk,i'j'}^\parallel+\sum_{k'}f_{i'j'k',ij}^\parallel=0,\\
T_{iji'j'}\left[\sum_{k}X_{ijk}\times f_{ijk,i'j'}^\parallel+
\sum_{k'}X_{i'j'k'}\times f_{i'j'k',ij}^\parallel\right]=0.
\end{cases}
\end{align*}
The above optimized when handled using the ALM algorithm, which in turn requires large-scale linear system solves in the underlying LM algorithm---a major computational bottleneck. To tackle this issue, we notice that the additional complexity due to the frictional forces can be largely eliminated by utilizing the special sparsity pattern in the underlying Gauss-Newton Hessian matrix. First, we notice that the ALM subproblem takes the following form:
\begin{equation}
\begin{aligned}
\label{eq:ALM-subproblem-friction}
\argmin{q,x,f_{iji'j'}}\;O+\lambda^T\bar{C}+\frac{\rho}{2}\|\bar{C}\|^2+
\sum_{iji'j'}\Phi(q,x,f_{iji'j'}),
\end{aligned}
\end{equation}
where each term $\Phi(q,x,f_{iji'j'})$ includes the constraint (\prettyref{eq:cone-constraint} and~\prettyref{eq:balance-constraint}) and augmented Lagrangian terms due to frictions between each pair of convex hulls $ij$ and $i'j'$, with definitions deferred to appendix. Different pairs of convex hulls are coupled only in the term $\bar{C}(q,x,f_{iji'j'})$, so that the Gauss-Newton Hessian matrix takes the following form:
\begin{align}
H\triangleq\nabla^2O + \sum_{iji'j'}\nabla^2\Phi + \nabla\bar{C}^T\nabla\bar{C}.
\end{align}

\vspace{-12pt}
\begin{figure}[h]
\centering
\includegraphics[width=0.85\linewidth]{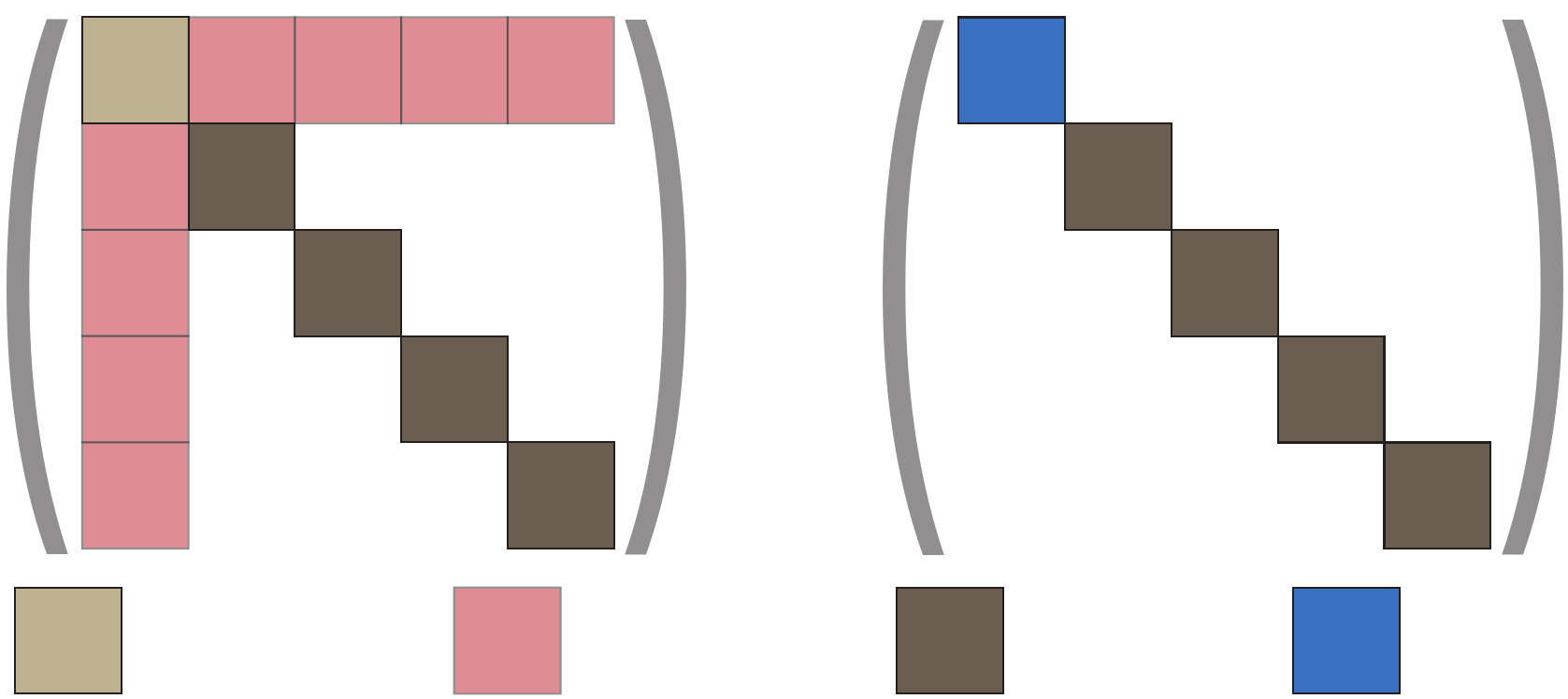}
\put(-195,6){\tiny{:$\setlength{\arraycolsep}{1pt}
\begin{array}{c}
\nabla C^T\nabla C+\\
\nabla^2O
\end{array}$}}
\put(-135,6){\tiny{:$\nabla_{iji'j',qx}^2\Phi$}}
\put(-74,6){\tiny{:$\nabla_{iji'j'}^2\Phi$}}
\put(-20,6){\tiny{:$\nabla^2O$}}
\put(-232,50){\small{$H=$}}
\put(-110,50){\small{$A=$}}
\caption{\label{fig:matrix}We illustrate the structure of matrix $H$ (left) and matrix $A$ (right). $A$ is block-diagonal and each block is small. $H$ is factored using the Woodbury matrix identity.}
\end{figure}

We first notice that $\nabla \bar{C}$ has a rank of at most $|q|$, which is the dimension of the configuration space. Therefore, we could use the Woodbury matrix identity to efficiently solve the linear system via~\prettyref{alg:woodbury}, where the related matrices are illustrated in~\prettyref{fig:matrix}. However,~\prettyref{alg:woodbury} still requires solving the linear system with matrix $A$ on the left hand side, which again involves all the decision variables. Fortunately, since the matrix $\nabla \bar{C}$ has been factored out of the matrix $A$, frictional forces between different pairs of convex hulls have been decoupled, allowing us to use the Schur complement solver to efficiently solve such linear systems. Specifically, given any righthand side $b$, we can decompose $b$ as:
\begin{align}
b=\FOURR{b_{qx}}{\cdots}{b_{iji'j'}}{\cdots},
\end{align}
where the first block corresponds to the first $|q|+|x|$ rows and the follow-up blocks each correspond to the rows for each pair of convex hulls $iji'j'$. Similarly, the Hessian of $\nabla^2\Phi(q,x,f_{iji'j'}^\parallel)$ has the following decomposition:
\begin{align}
\nabla^2\Phi=
\MTT
{\nabla_{qx}^2\Phi}{\nabla_{qx,iji'j'}^2\Phi}
{\nabla_{iji'j',qx}^2\Phi}{\nabla_{iji'j'}^2\Phi}
\end{align}
With the above notation, we can solve the linear system as outlined in~\prettyref{alg:schur}.
\begin{algorithm}[ht]
\caption{\label{alg:woodbury}Solve-H(b)}
\begin{algorithmic}[1]
\Require{$\nabla \bar{C},\nabla^2\Phi(q,x,f_{iji'j'}^\parallel),\nabla^2O$}
\Ensure{$H^{-1}b$}
\State $A\gets\nabla^2O + \sum_{iji'j'}\nabla^2\Phi$
\LineComment{We use the following Woodbury matrix identity}
\LineComment{$A^{-1}-A^{-1}\nabla \bar{C}^T(I+\nabla \bar{C}A^{-1}\nabla \bar{C}^T)^{-1}\nabla \bar{C}A^{-1}$}
\State $b\gets$Solve-A($b$)
\State $c\gets\nabla \bar{C}^T(I+\nabla \bar{C}A^{-1}\nabla \bar{C}^T)^{-1}\nabla \bar{C}b$
\State Return $b-$Solve-A($c$)
\end{algorithmic}
\end{algorithm}
\begin{algorithm}[ht]
\caption{\label{alg:schur}Solve-A(b)}
\begin{algorithmic}[1]
\Require{$\nabla^2\Phi(q,x,f_{iji'j'}^\parallel),\nabla^2O$}
\Ensure{$A^{-1}b$}
\State $A_{qx}\gets\nabla^2O$
\For{Each pair of convex hulls $iji'j'$}
\State $A_{qx}\gets A_{qx}-\nabla_{qx,iji'j'}^2\Phi
\nabla_{iji'j'}^2\Phi^{-1}
\nabla_{iji'j',qx}^2\Phi$
\State $b_{qx}\gets b_{qx}-\nabla_{qx,iji'j'}^2\Phi^T
\nabla_{iji'j'}^2\Phi^{-1}b_{iji'j'}$
\EndFor
\State $b_{qx}\gets A_{qx}^{-1}b_{qx}$
\For{Each pair of convex hulls $iji'j'$}
\State $b_{iji'j'}\gets
\nabla_{iji'j'}^2\Phi^{-1}
(b_{iji'j'}-\nabla_{iji'j',qx}^2\Phi b_{qx})$
\State Return $b$
\EndFor
\end{algorithmic}
\end{algorithm}

%% file: camera_ready/pipeline.tex
\subsection{Scene Estimation Pipeline}
Our joint optimizer uses a local gradient-based method that is prone to getting trapped in local minima without a high-quality initialization. To obtain a high-quality initial guess, we first extract the point cloud and then leverage the learning-based model SAM3D~\cite{chen2025sam} to guess initial object shapes from the point cloud. SAM3D allows us to extract the mesh and segment the point cloud for each mesh. The result is denoted as $N$ points clouds $\mathcal{P}_{1,\cdots,N}$ and $N$ corresponding meshes $\mathcal{M}_{1,\dots,N}$. While SAM3D also predicts object poses, we observe that these estimates are often inaccurate. Therefore, we refine the object poses using the FoundationPose model~\cite{wen2024foundationpose}, producing an initial guess for the pose vector $q$.
Next, for each mesh $\mathcal{M}_i$, we apply convex decomposition~\cite{wei2022coacd} to generate the vertices of convex hulls, yielding the initial guess for the vector $x$. These initial estimates are then adjusted to ensure penetration-free between different bodies and fed into our joint optimization to enforce the physical constraints (see our appendix for details). For convenience of presentation, we assume that each rigid body is represented by \rev{an equal number of} $M$ convex hulls, each with $V$ vertices, our implementation is fully general and can accommodate the case \rev{where each rigid body has different number of convex hulls and vertices per hull, as determined by the convex decomposition algorithm~\cite{wei2022coacd}, detailed in Appendix \ref{appen:geometry_details}.}

With the segmented point clouds $\mathcal{P}_i$, the extracted mesh $\mathcal{M}_i$, and the convex hulls, we can define differentiable objective function $O$ via the similar technique as the trimmed ICP~\cite{chetverikov2002trimmed}, where we select ICP terms to ensure monotonic objective value reduction over iterations, thus guaranteeing convergence. Specifically, for each convex hull vertex $X_{ijk}$ in world space, we already know that it belongs to the $i$th rigid body, so we search for the closest point on $\mathcal{M}_i$ as a continuous manifold to $X_{ijk}$, denoted as:
\begin{align}
\label{eq:closest-on-mesh}
p(X_{ijk})=\argmin{\rev{X\in\mathcal{M}_i}}\|\rev{X}-X_{ijk}\|.
\end{align}
We then fix $p(X_{ijk})$ and introduce a term $\|X_{ijk}(q,x)-p(X_{ijk})\|^2$ to regularize the convex hull shapes. Conversely, for \rev{each point $p_{i}$} belonging to the $i$th point cloud, we penalize its distance to the surface formed by union of convex hulls representing the $i$th rigid body, denoted as $\partial\cup_j\CH(X_{ij\bullet})$. We find the closest point to $\rev{p_{i}}$ on the union of convex hulls as a continuous manifold, denoted as:
\begin{align}
\label{eq:closest-on-convex-hull}
X(\rev{p_{i}})=\argmin{X\in\partial\cup_j\CH(X_{ij\bullet})}\|X-\rev{p_{i}}\|.
\end{align}
In practice, we use Manifold3d Library~\cite{manifold2025} to compute union of convex hulls and extract the surface into a triangle mesh, which then allows us to compute $X(\rev{p_{i}})$ as a point-to-mesh distance problem. The computed $X(\rev{p_{i}})$ must lie on the surface of a convex hull. Without a loss of generality, we can assume its the $j(\rev{p_{i}})$'s convex hull, i.e., $X(\rev{p_{i}})=\sum_{k}X_{ij(\rev{p_{i}})k}w_k$, with $w_k$ being the convex combination weights.
We then follow the idea of ICP and fix the convex combination weights $w_k$ so that $X(\rev{p_{i}})$ is a function of $x$ and $q$ only. We can then introduce the term $\|X(\rev{p_{i}})-\rev{p_{i}}\|^2$ to further regularize the convex hull shapes. Finally, we notice that the point-cloud can only regulate the object shape in visible areas. To further regulate object shapes in invisible areas, we utilize the SAM3D-recovered mesh $\mathcal{M}_i$. \rev{For each vertex $p_{i}\in\mathcal{M}_i$}, we use a similar technique to compute the closest point $X(\rev{p_{i}})$ and introduce the term $\|X(\rev{p_{i}})-\rev{p_{i}}\|^2$. In summary, our objective function is formulated as a sum of three types of terms weight by coefficients $w_{1,2,3}$:
\begin{align}
\label{eq:objective}
O(q,x)=
\begin{cases}
w_1\sum_{ijk}\|X_{ijk}-p(X_{ijk})\|^2+ & \text{Type I}\\
w_2\sum_{\rev{p_{i}}\in\mathcal{P}_i}
\|X(\rev{p_{i}})-\rev{p_{i}}\|^2+ & \text{Type II}\\
w_3\sum_{\rev{p_{i}}\in\mathcal{M}_i}
\|X(\rev{p_{i}})-\rev{p_{i}}\|^2 & \text{Type III}.
\end{cases}
\end{align}
Our formulation follows the idea of Hausdorff distance that penalizes symmetric distances. Our type I term regularize the distance between convex hull's vertex $X_{ijk}$ and $\mathcal{M}_i$. The remaining type II (resp. type III) term regularize the distance between the point cloud (resp. mesh) and the union of convex hull surface $\partial\cup_j\CH(X_{ij\bullet})$. We assume that the point cloud $\mathcal{P}_i$ is the direct observation and serves a strong guidance for correct object shapes, which must be emphasized as the main objective using a large weight $w_2$. On the other hand, the mesh $\mathcal{M}_i$ might be hallucinated by SAM3D~\cite{chen2025sam}, which is not necessarily matched exactly, but used as a shape prior via a small weight $w_3$.

\begin{figure}[ht]
\centering
\includegraphics[width=0.95\linewidth]{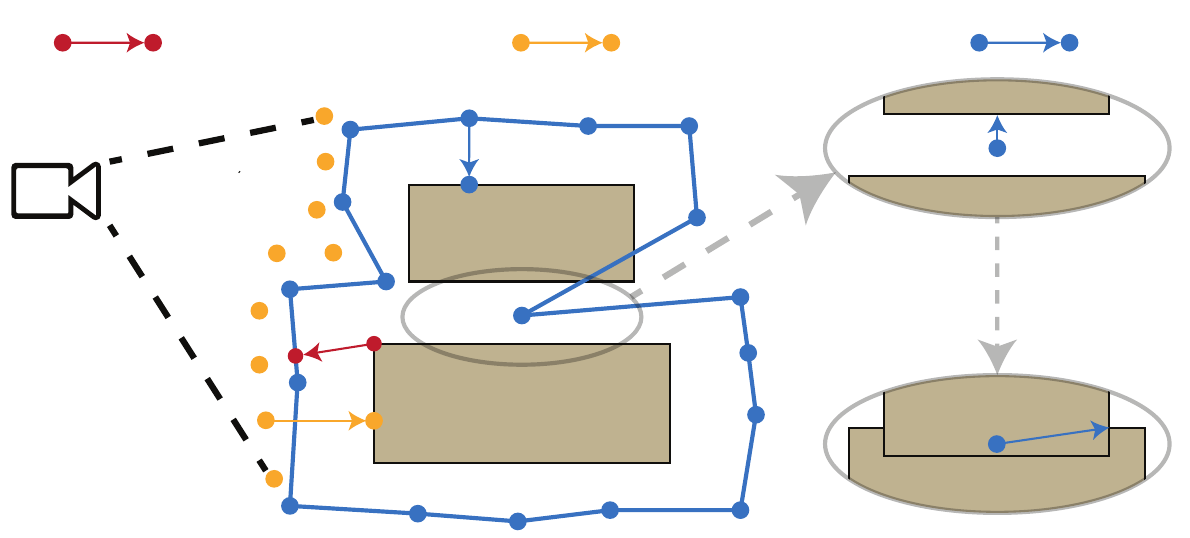}
\put(-247,98){\tiny$X_{ijk}$}
\put(-203,98){\tiny$p(X_{ijk})$}
\put(-225,105){\tiny{Type I}}
\put(-146,98){\tiny$\rev{p_{i}}$}
\put(-113,98){\tiny$X(\rev{p_{i}})$}
\put(-133,105){\tiny{Type II}}
\put(-54,98){\tiny$\rev{p_{i}}$}
\put(-19,98){\tiny$X(\rev{p_{i}})$}
\put(-42,105){\tiny{Type III}}
\put(-150,42){\small{(a)}}
\put(-60,57){{\small(b)}}
\put(-60,35){{\small(c)}}
\caption{\label{fig:objective}We illustrate the three types of objectives in~\prettyref{eq:objective}, regularizing the distance between convex hull vertex $X_{ijk}$ (red), the SAM3D-identified mesh vertex $\rev{p_{i}}\in\mathcal{M}_i$ (blue), and the point cloud $\rev{p_{i}}\in\mathcal{P}_i$ (yellow). Further, we highlight a case (a) where objective value can increase. Suppose our rigid body (light brown) consists of two disjoint convex hulls (b), the closest point to the blue vertex is the bottom surface of the top hull. After an update to hull vertices (c), the two hulls merge and the closest point is moved to the right boundary.}
\end{figure}
A critical drawback of the above procedure is the violation of NLP convergence guarantee. Note that our type I term $\|X_{ijk}-p(X_{ijk})\|^2$ adopts a standard treatment of ICP, where we can fix $p(X_{ijk})$ to solve ALM optimization, and then update $p(X_{ijk})$ via~\prettyref{eq:closest-on-mesh}. This procedure is guaranteed to converge~\cite{chetverikov2002trimmed} because each term monotonically decreases over iterations. However, our type II and type III terms $\|X(\rev{p_{i}})-\rev{p_{i}}\|^2$ are not guaranteed to decrease over iterations because the update in~\prettyref{eq:closest-on-convex-hull} can increase function values. This is essentially because the reference mesh $\mathcal{M}_i$ has fixed geometry but our union of convex hull can undergo shape changes. To rigorously ensure convergence, we introduce a heuristic technique inspired by~\cite{chetverikov2002trimmed}, where we selectively delete a subset of type II and type III terms that can increase function values. Specifically, we sort all type II and type III terms by the amount of function value increment after an update of $X(\rev{p_{i}})$ according to~\prettyref{eq:closest-on-convex-hull}:
\begin{align}
\label{eq:X_il-prev}
\rev{\Delta_{i}}=\|X(\rev{p_{i}})-\rev{p_{i}}\|^2-\|X(\rev{p_{i}})^\text{prev}-\rev{p_{i}}\|^2,
\end{align}
where $X(\rev{p_{i}})^\text{prev}=\sum_{k}X_{ij(\rev{p_{i}})k}w_k^\text{prev}$ is the point using convex combination weights from the previous iteration. We repeatedly delete the type II or type III term yielding the highest function value increase until the objective function $O$ is non-increasing. The entire procedure of closest point update and selective term deletion is performed after each ALM subproblem solve. The complete pipeline is outlined in our \rev{Appendix \ref{appen:optimization_details}} and illustrated in~\prettyref{fig:objective}. 

As an optional final step of our method, we can generate the color texture for each object by differentiable rasterization. Specifically, after the ALM optimization, we fix the object shape and pose. We then use Manifold3d Library~\cite{manifold2025} to convert the union of convex hulls into a triangle mesh. Next, we use xatlas~\cite{young2019xatlas} to generate the UV coordinates for each mesh. Finally, we use differentiable renderer~\cite{Laine2020diffrast} to minimize the difference between the SAM3D-predicted and the mesh-rendered image, with the texture map being the decision variables.

%% file: camera_ready/evaluation.tex
\begin{figure*}[t]
\centering
\setlength{\tabcolsep}{1px}
\begin{tabular}{ccccc}
\includegraphics[width=0.19\linewidth]{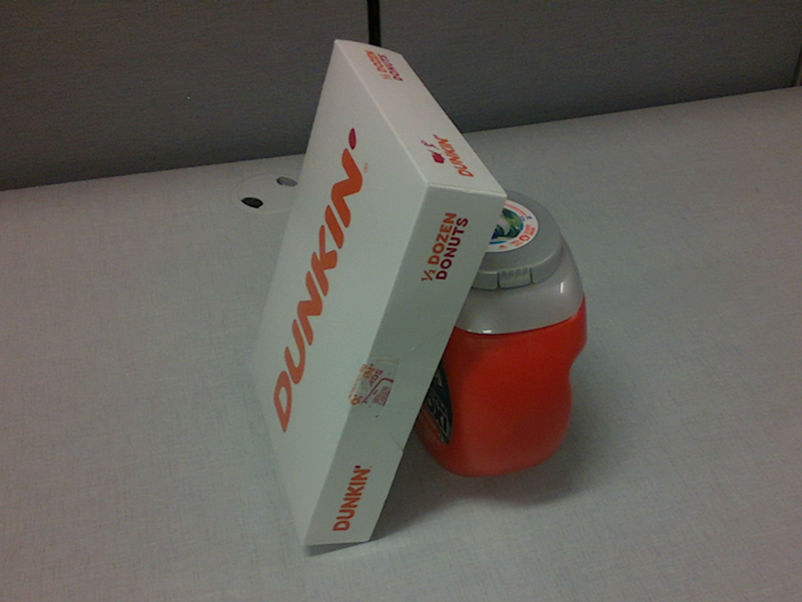}&
\includegraphics[width=0.19\linewidth]{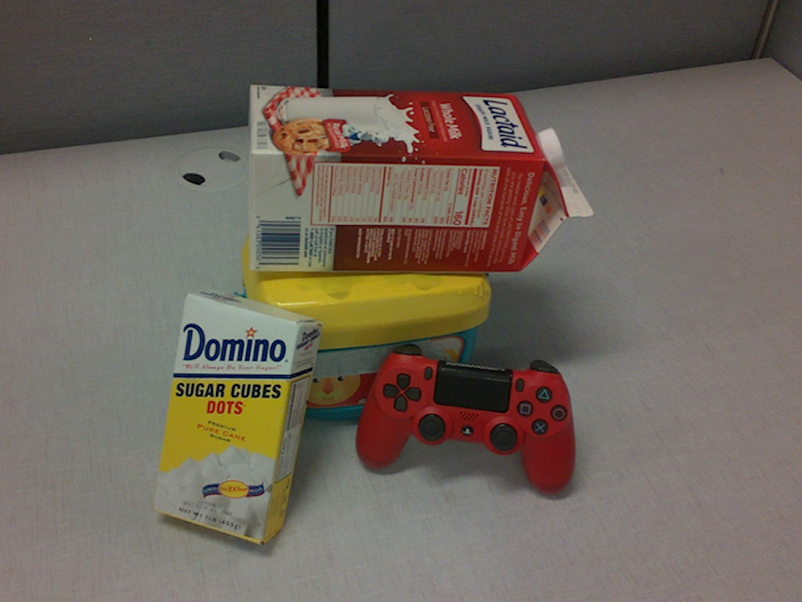}&
\includegraphics[width=0.19\linewidth]{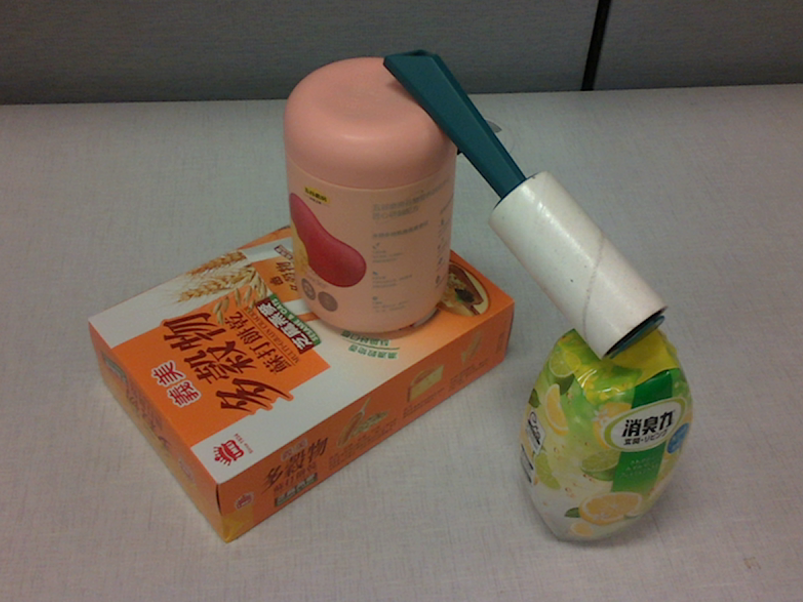}&
\includegraphics[width=0.19\linewidth]{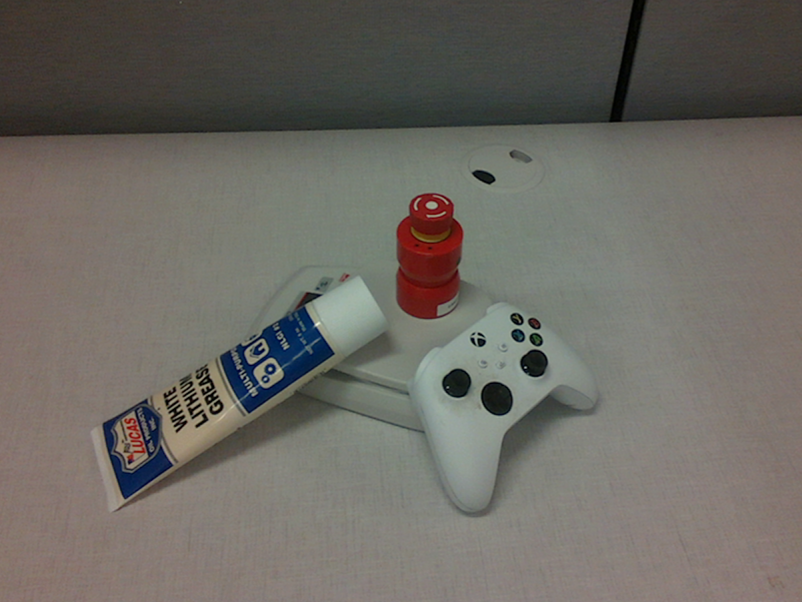}&
\includegraphics[width=0.19\linewidth]{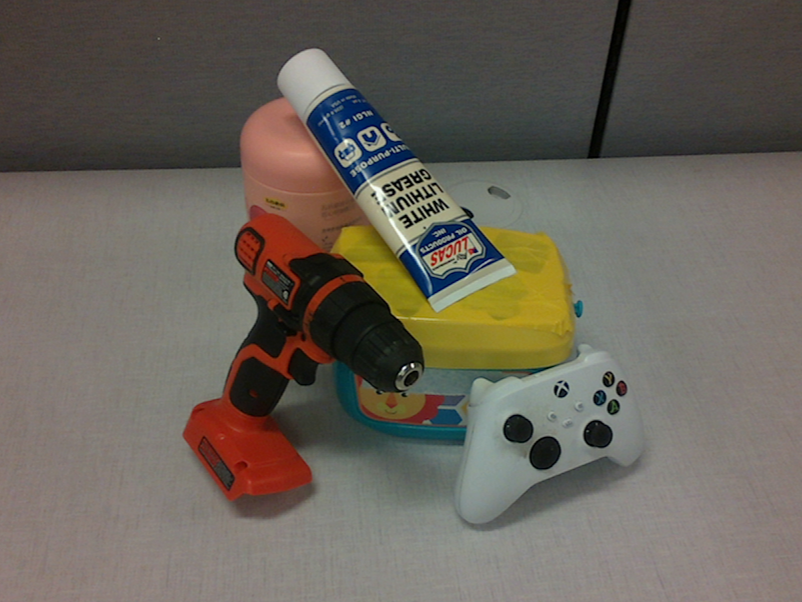}\\
\includegraphics[width=0.19\linewidth]{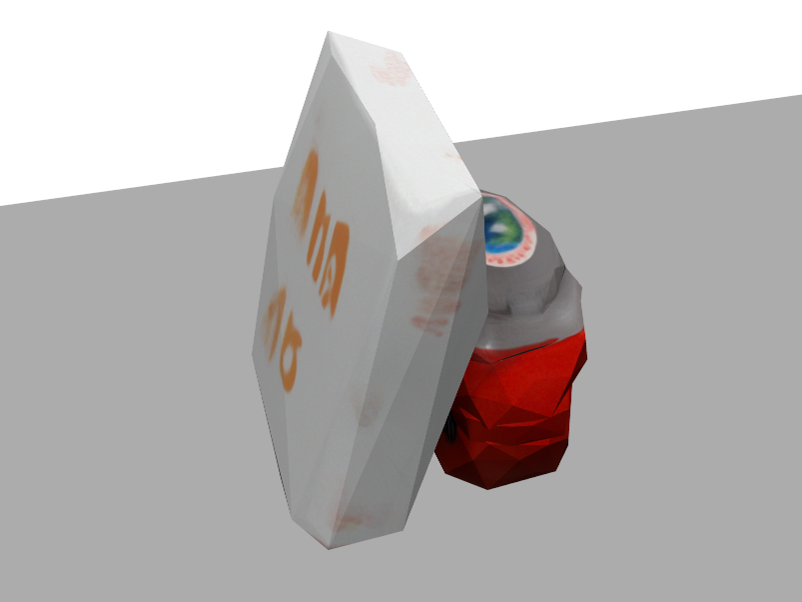}&
\includegraphics[width=0.19\linewidth]{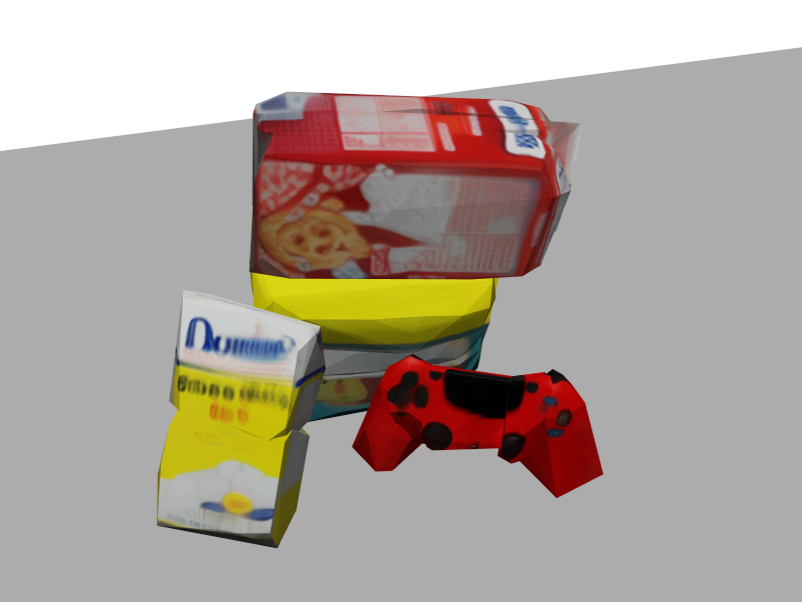}&
\includegraphics[width=0.19\linewidth]{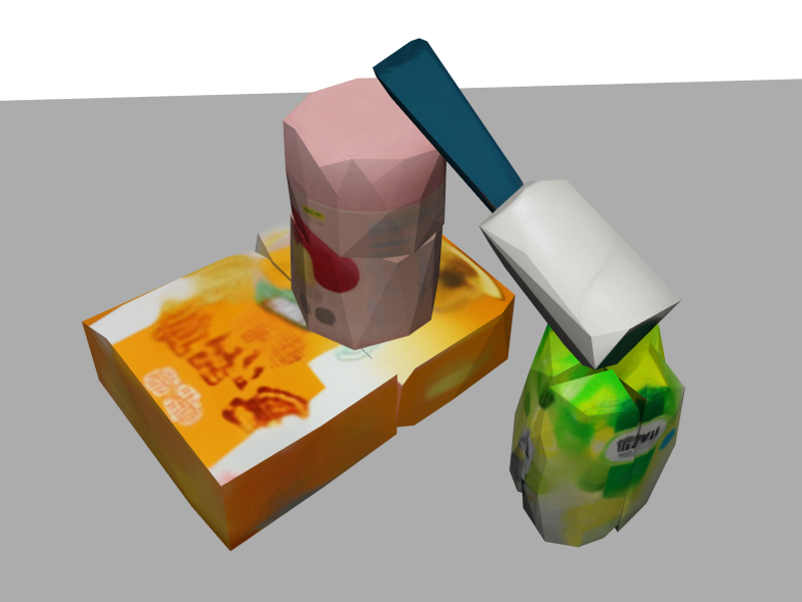}&
\includegraphics[width=0.19\linewidth]{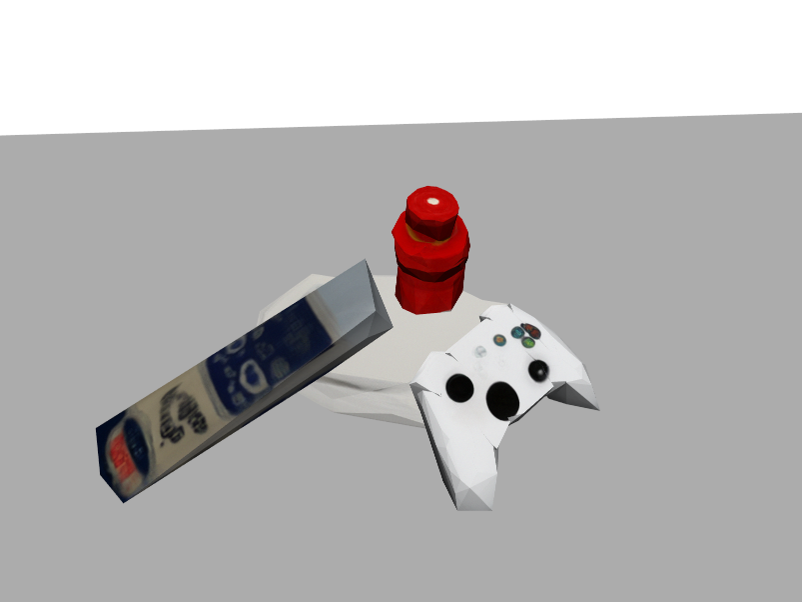}&
\includegraphics[width=0.19\linewidth]{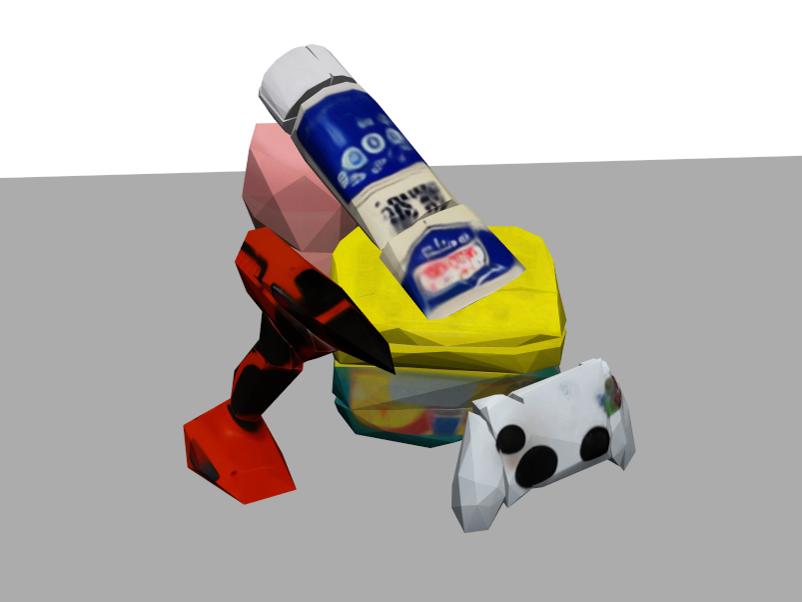}\\
\includegraphics[width=0.19\linewidth]{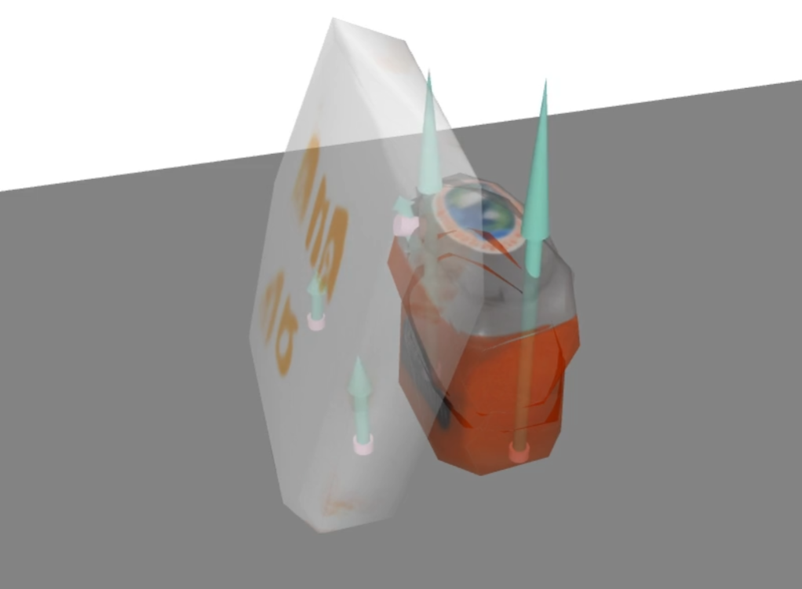}&
\includegraphics[width=0.19\linewidth]{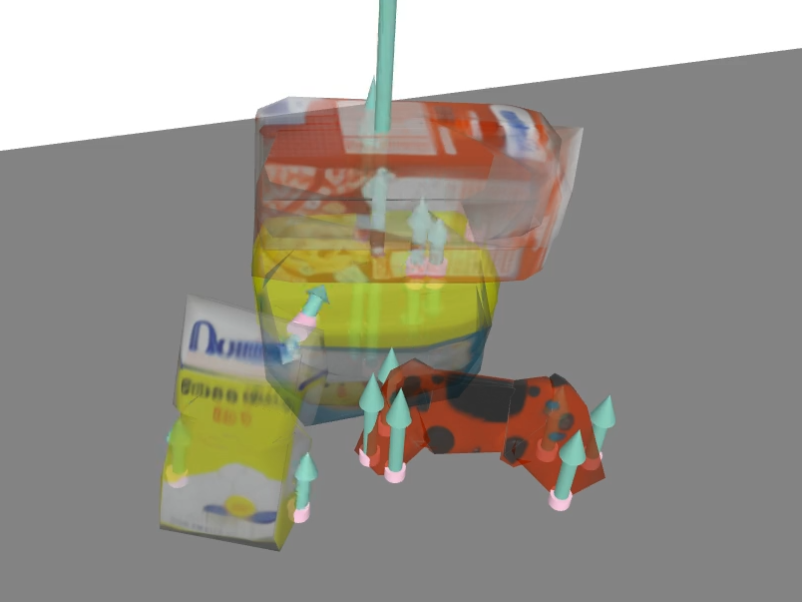}&
\includegraphics[width=0.19\linewidth]{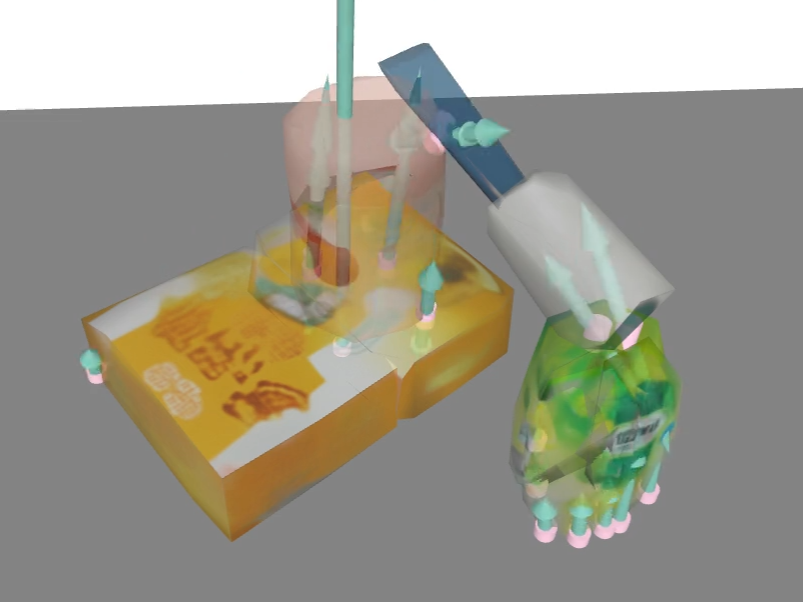}&
\includegraphics[width=0.19\linewidth]{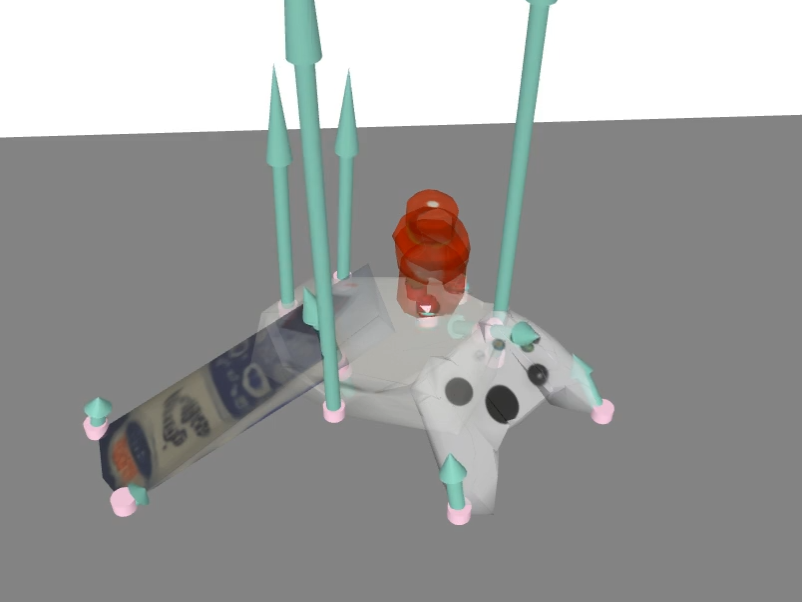}&
\includegraphics[width=0.19\linewidth]{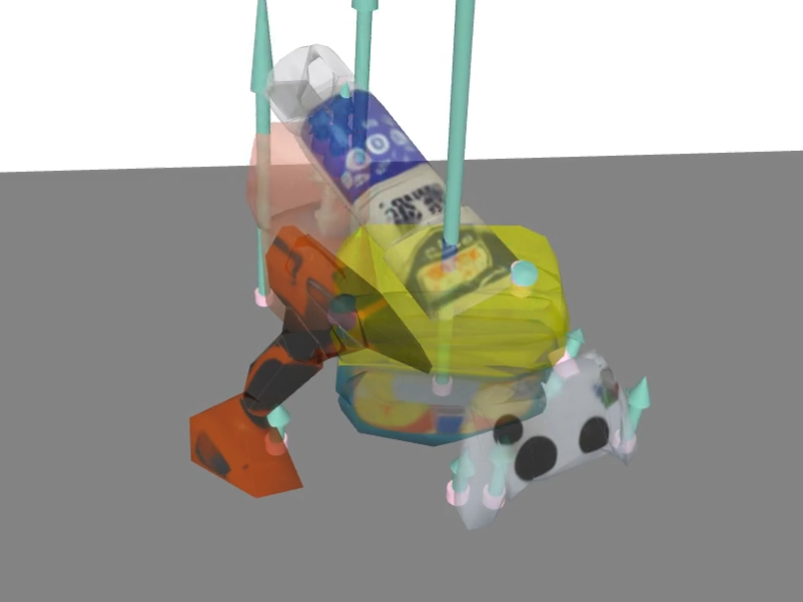}
\end{tabular}
\caption{\label{fig:benchmark}A visualization of our benchmarking scenarios. Top Row: The input single-view image observation. Middle Row: The estimated simulation-ready rigid bodies models. Bottom Row: Our estimated scenes achieve physical force equilibrium in MuJoCo~\cite{todorov2012mujoco}, showing simulated contact forces after settling.}
\end{figure*}
\section{Evaluation}
We have evaluated our method on a single Intel Core Ultra 9 285K CPU and a single GeForce RTX 5090 GPU with 32GB of memory. All the computations are done on CPU except for differentiable rendering based texture optimization. We use multi-threading to compute evaluate different rows of the Jacobian. We propose to use the Levenberg-Marquardt algorithm, a form of the Quasi-Newton's method to serve as our subproblem solver~\prettyref{prob:ALM-subproblem}. This is because~\prettyref{prob:ALM-subproblem} is essentially solving a least-square problem with $r(q,x)$ being the residual, due to the following rewrite:
\begin{align}
O+\lambda^TC+\frac{\rho}{2}\|C\|^2
\propto O+\frac{\rho}{2}\|C+\lambda/\rho\|^2
\triangleq\|r(q,x)\|^2,
\end{align}
and noting that our objective function $O$ (\prettyref{eq:objective}) also takes a least square form (A similar argument applies to the frictional constraint part, which is omitted here for brevity, \rev{and the full details are provided in Appendix \ref{appen:optimization_details}}). We terminate the subproblem solver when $\|r\|_\infty\leq\epsilon_r$ or $\|r^T\FPPR{r}{\TWO{x}{q}}\|_\infty\leq\epsilon_g$. We find that due to the physics constraint $C$ being very stiff, leading to the subproblem solver make very small progress over many iterations. We thus terminate the inner loop when the progress is less than $1\%$ over $20$ iterations. Finally, we terminate the outer ALM iteration when $\|C\|_\infty\leq\epsilon_C$ or the residual of the KKT condition does not improve by more than $1\%$ over consecutive ALM iterations. Through all our experiments, we choose parameters $\epsilon_r=10^{-6}, \epsilon_g= 10^{-2}, \epsilon_C=5 \times 10^{-4}$. 

\begin{table}[ht]
\begin{tabularx}{0.49\textwidth}{c *{4}{Y}}
\toprule
Scenario
 & \multicolumn{2}{c}{Max Kinetic Energy (J) $\downarrow$}  
 & \multicolumn{2}{c}{Max Drift Distance (cm) $\downarrow$}\\
\cmidrule(lr){2-3} \cmidrule(l){4-5}
  & Ours & SAM3D & Ours & SAM3D \\
\midrule
 1  &  $\E{7.19 \times 10^{-4}}$ &  $4.77 \times 10^{0}$ &  $\E{1.62}$ &  $59.41$ \\
 2  &  $\E{2.24 \times 10^{-3}}$ &  $5.68 \times 10^{0}$ &  $\E{0.83}$ &  $87.56$ \\
 3  &  $\E{1.12 \times 10^{-2}}$ &  $2.08 \times 10^{0}$ &  $\E{3.10}$ &  $31.06$ \\
 4  &  $\E{3.32 \times 10^{-3}}$ &  $2.32 \times 10^{0}$ &  $\E{0.73}$ &  $172.55$ \\
 5  &  $\E{4.36 \times 10^{-3}}$ &  $5.73 \times 10^{0}$ &  $\E{2.28}$ &  $51.69$ \\
\bottomrule
\end{tabularx}
\caption{\label{table:stability} We summarize the simulator stability by the amount of kinetic energy gain (left) during the first 1 second and drift distance (right) over the first 1 minute of simulation time.}
\end{table}
\begin{table}[ht]
\begin{tabularx}{0.49\textwidth}{c *{3}{Y}}
\toprule
Scenario
 & \multicolumn{3}{c}{PSNR$\uparrow$}  \\
\cmidrule(lr){2-4}
  & Ours vs. RGB & SAM3D vs. RGB & Ours vs. SAM3D \\
\midrule
 1  &  $17.92$ &  $\E{18.11}$ &  $20.16$ \\
 2  &  $\E{19.99}$ &  $18.99$ &  $22.32$ \\
 3  &  $\E{18.37}$ &  $17.34$ &  $18.70$ \\
 4  &  $\E{21.43}$ &  $21.11$ &  $23.95$ \\
 5  &  $20.15$ &  $\E{20.32}$ &  $20.17$ \\
\bottomrule
\end{tabularx}
\caption{\label{table:consistency} We profile the PSNR between the images rendered using our estimated shape and pose (Ours), SAM3D estimated initial guess, which is further adjusted by FoundationPose (SAM3D), and the Groudtruth RGBD image (RGB).}
\end{table}
\paragraph{Simulation-ready Results} All benchmarks are visualized in~\prettyref{fig:benchmark}. We evaluate our method on five cluttered tabletop scenes containing up to 5 rigid objects, represented by a total of 22 convex hulls. To assess the simulation-ready quality of our reconstructions, we forward the estimated object shapes and poses into the widely used MuJoCo physics simulator~\cite{todorov2012mujoco}. Under standard physical parameter settings (\rev{see Appendix \ref{appen:extended_results} for details on simulator setup}), our reconstructions remain in force equilibrium over 1 minute of simulation time. In contrast, the initial shape and pose estimates produced by SAM3D~\cite{chen2025sam} and FoundationPose~\cite{wen2024foundationpose} always contain severe inter-penetrations, causing simulation instability and failure, as reported in~\prettyref{table:stability}. We also try three most recent single-view scene reconstruction works \cite{Ardelean2025Gen3DSR, huang2025midi, Agarwal2026scenecomplete}, yet none of them can produce comparable result (as illustrated in Appendix \rev{\ref{appen:extended_results}}). 

We further compare visual fidelity against the initial SAM3D+FoundationPose estimates. As shown in~\prettyref{table:consistency}, our results achieve comparable PSNR, indicating that physical consistency is improved without sacrificing visual accuracy. \rev{We also conduct ablation studies on objective terms (detail illustrations in Appendix \ref{appen:extended_results}), showing that both the main term (Type II) and shape regularization terms (Type I and Type III) are necessary to achieve satisfactory estimation result.}

\begin{table}[ht]
\centering
\begin{tabular}{cccccc}
\toprule
Scenario & \#Hull & \#Vertex & \#ALM & \#LM & Wall Time (min)\\
\midrule
1 & 6 & 299 & 7 & 2536  & 46.1\\
2 & 16 & 795 & 6 & 2322 & 259.7\\
3 & 10 & 487 & 9 & 4529 & 224.2\\
4 & 12 & 597 & 7 & 3106 & 202.15\\
5 & 22 & 1099 & 7 & 2640 & 539.92\\
\bottomrule
\end{tabular}
\caption{\label{table:statistics}We summarize the statistics of our benchmarking scenarios. From left to right: the total number of convex hulls, the total number of vertices, number of ALM outer iterations, number of LM iterations, and the total computational time.}
\vspace{-10px}
\end{table}
\begin{figure}[ht]
\centering
\includegraphics[width=\linewidth]{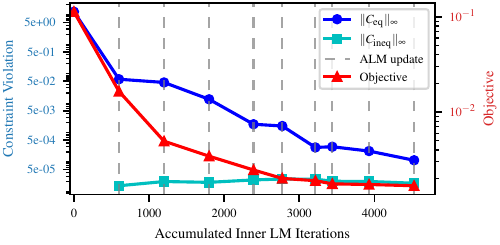}
\caption{\label{fig:convergence}The convergence history of a typical optimization procedure, which converges within $9$ ALM iterations.}
\vspace{-10px}
\end{figure}
\begin{figure}[ht]
\centering
\includegraphics[width=\linewidth]{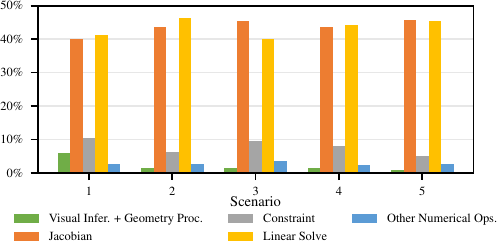}
\caption{\label{fig:performance}Our method involves Visual (SAM3D, FoundationPose) inference \& geometry process, constraint \& Jacobian evaluation, linear solve, and other operations. This figure shows the performance breakdown in percentage.}
\vspace{-10px}
\end{figure}

\begin{figure*}[t]
\centering
\setlength{\tabcolsep}{1px}
\begin{tabular}{cccc}
\includegraphics[width=0.24\linewidth]{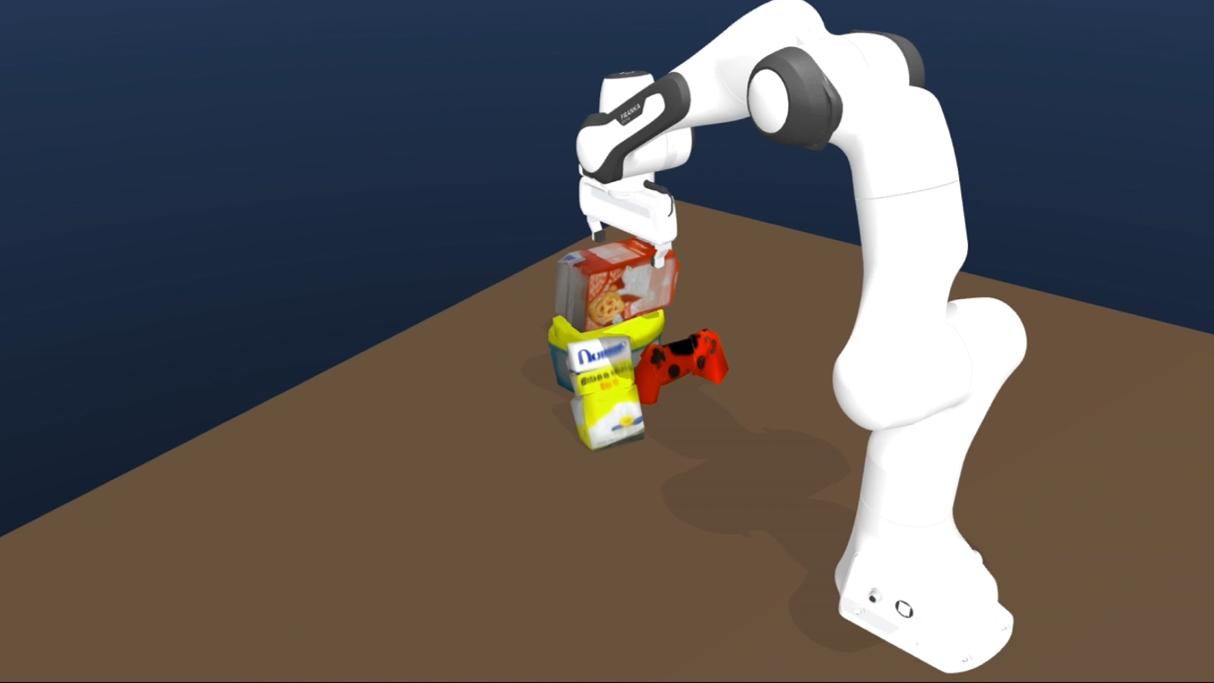}&
\includegraphics[width=0.24\linewidth]{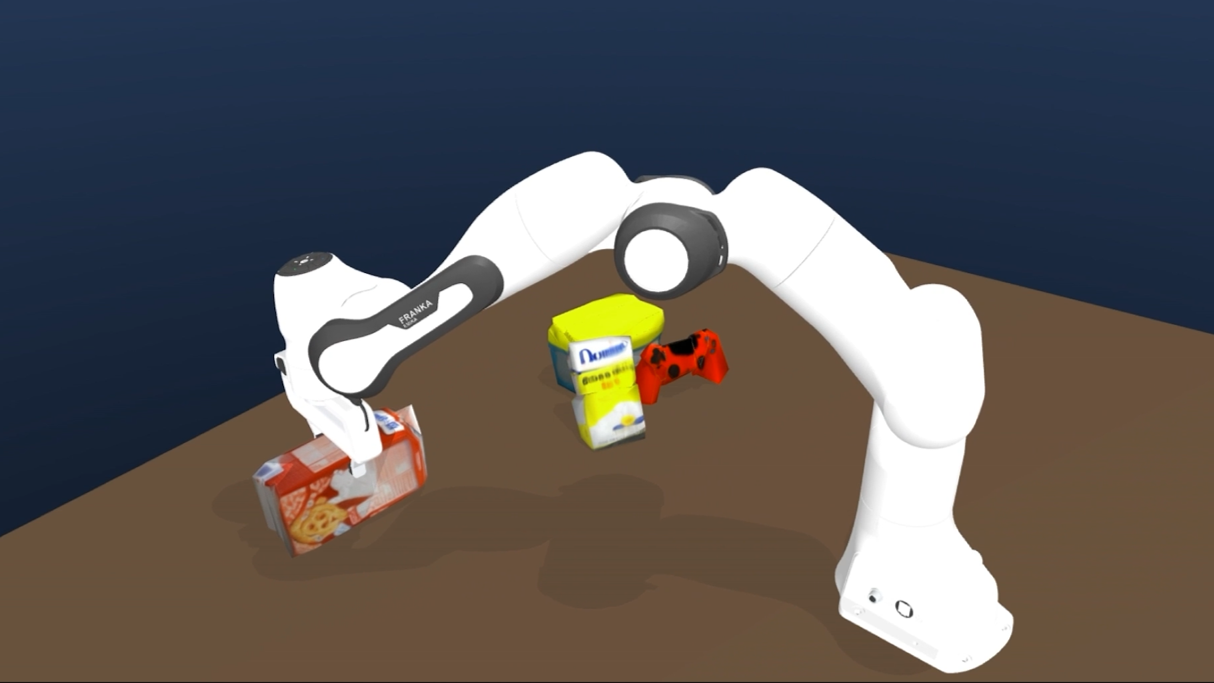}&
\includegraphics[width=0.24\linewidth]{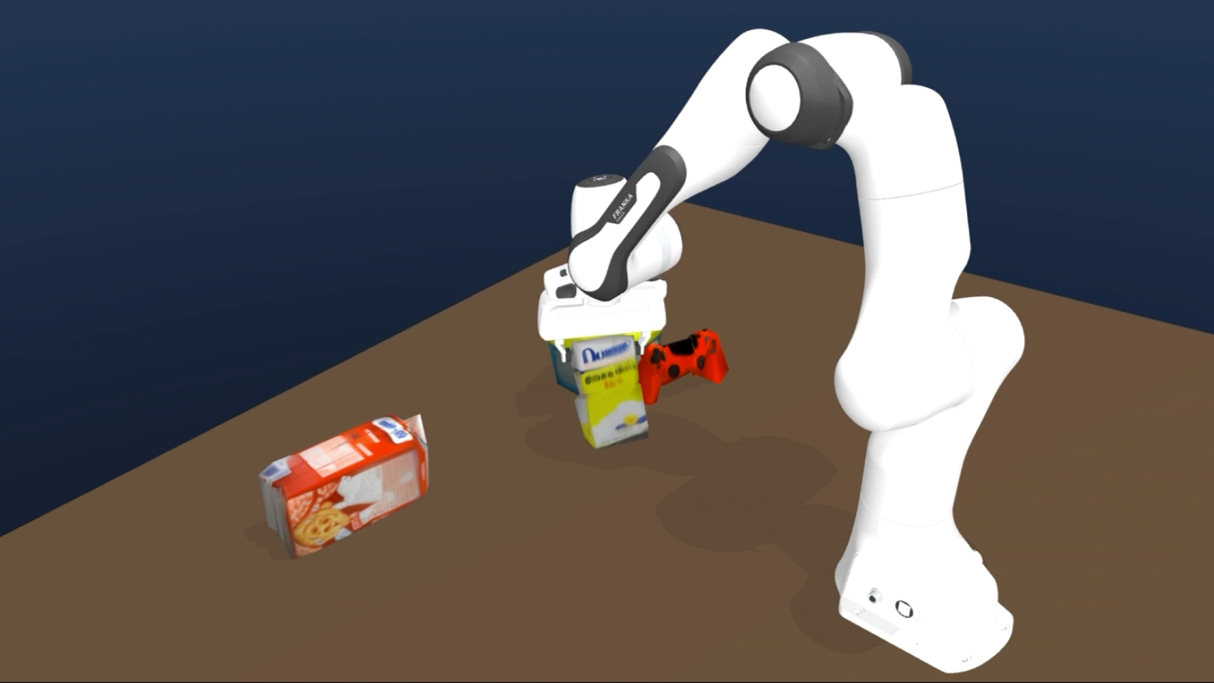}&
\includegraphics[width=0.24\linewidth]{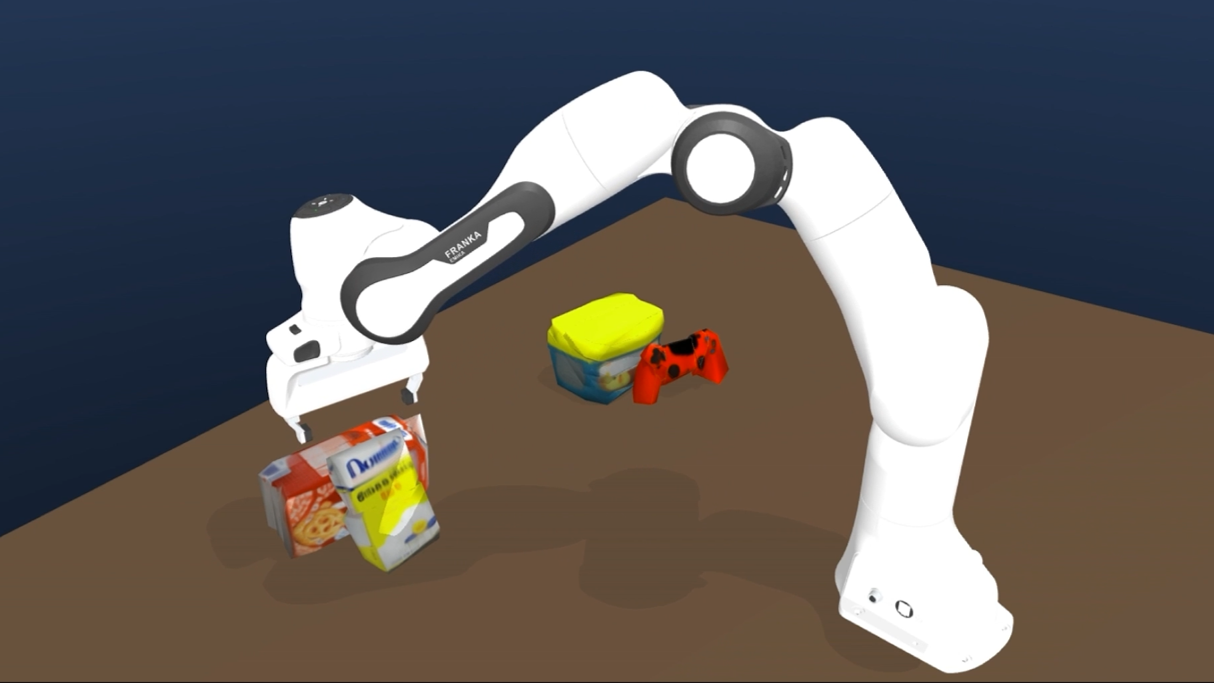}\\
\includegraphics[width=0.24\linewidth]{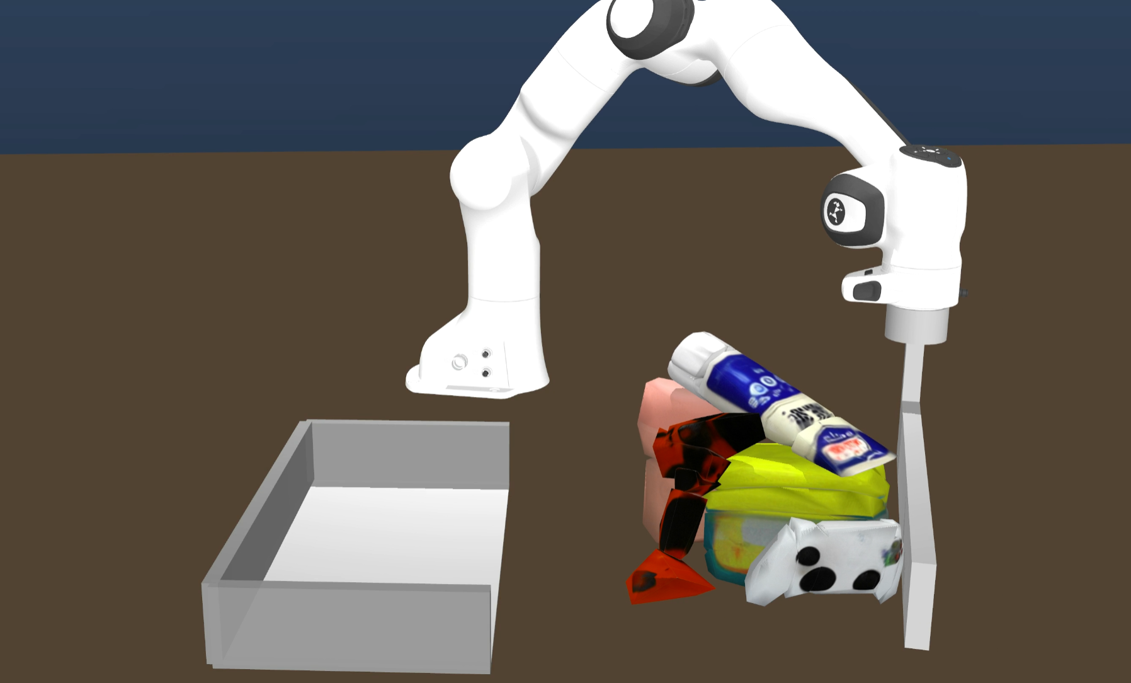}&
\includegraphics[width=0.24\linewidth]{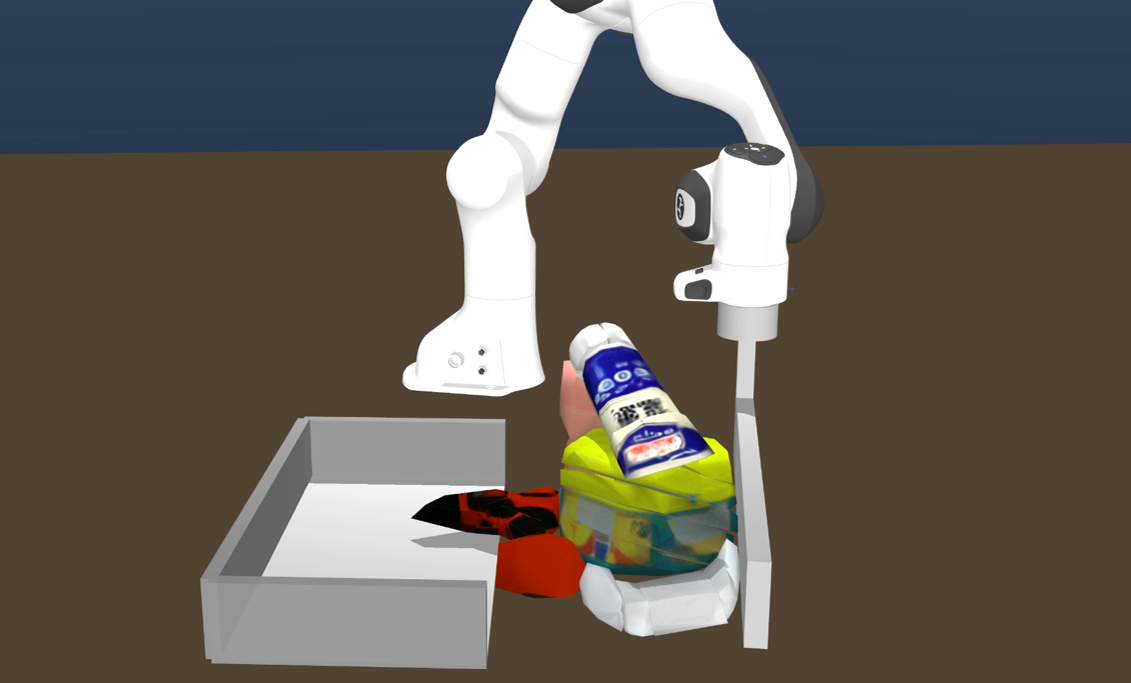}&
\includegraphics[width=0.24\linewidth]{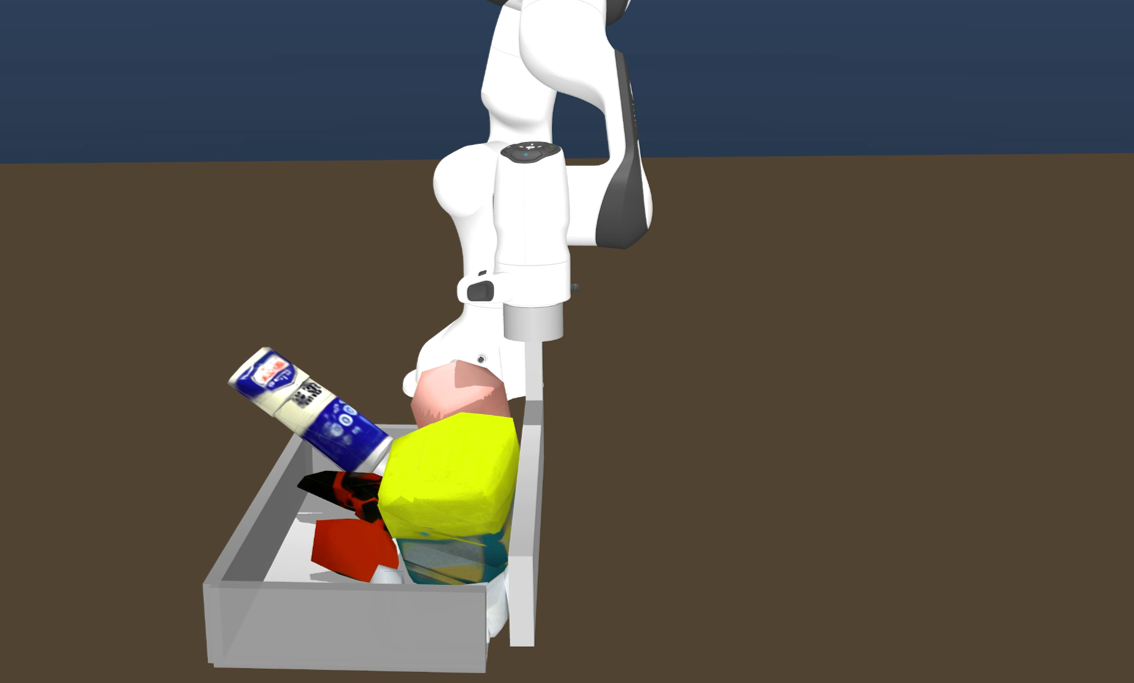}&
\includegraphics[width=0.24\linewidth]{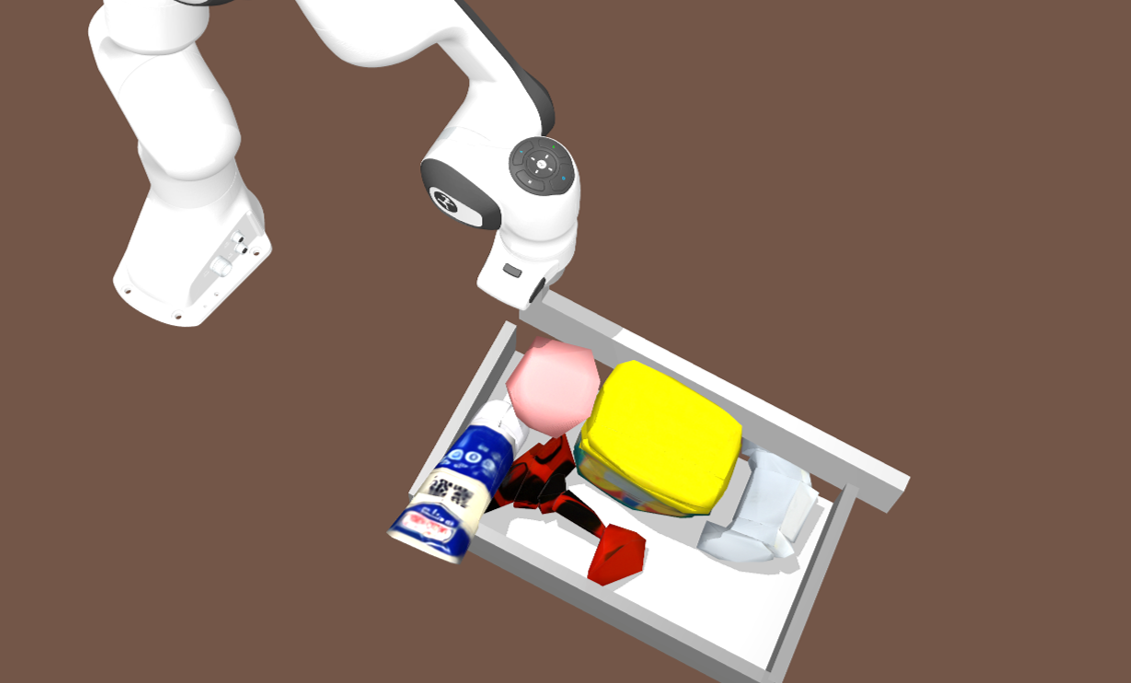}
\end{tabular}
\caption{\rev{\label{fig:manipulation task} Two demonstrations of conducting robotic manipulation tasks in MuJoCo simulation on the reconstructed scene.}}
\end{figure*}

\begin{table}[ht]
\centering
\begin{tabular}{ccccc}
\toprule
Scenario & \#Params & Woodbury (Alg.~\ref{alg:woodbury}) & Direct LU  & Speedup \\
\midrule
1 & 2751  & 0.475 & 0.681  & 1.43$\times$ \\
2 & 12045 & 3.162 & 19.713  & 6.23$\times$ \\
3 & 7425  & 1.318 & 7.226  & 5.48$\times$ \\
4 & 9075  & 1.814 & 8.085  & 4.45$\times$ \\
5 & 19932 & 7.342 & 63.909 & 8.71$\times$ \\
\bottomrule
\end{tabular}
\caption{\label{table:linear} The average cost comparison of solving the structured linear system (in seconds) using our method (\prettyref{alg:woodbury}) and direct LU factorization.}
\vspace{-10px}
\end{table}
\paragraph{Performance} Benchmark statistics are summarized in~\prettyref{table:statistics}. Our algorithm converges within 6-9 ALM iterations, with a representative convergence trajectory shown in~\prettyref{fig:convergence}. A detailed performance breakdown is provided in~\prettyref{fig:performance}. The dominant computational cost arises from repeated evaluations of the physics constraint $\bar{C}$ and its Jacobian, which require solving nested separating-plane optimizations between convex hulls. The second most expensive component is the linear solve. We further evaluate our structured linear solver (\prettyref{alg:woodbury}) and compare it against direct LU factorization in~\prettyref{table:linear}. The results demonstrate an overall speedup of up to $8.7\times$.

\rev{\paragraph{Downstream Applications} To illustrate the benefit of generating sim-ready scene estimate from images. We demonstrations two examples of conducting robotic manipulation tasks on the reconstructed scene in simulation. The first is a quasi-static pick and place task on Scenario 2 to rearrange the milk box and sugar box placement. The second is a dynamic pushing and sweeping task on Scenario 5 to relocate the clutter to target region. The screen shots of the tasks are shown in Figure \ref{fig:manipulation task}. For complete task executions, please check our supplementary video. These examples show the potential of our pipeline in generating large scale simulation scenes from real images for robotic manipulation data collection and policy training.}

%% file: camera_ready/conclusion.tex
\section{Conclusion}
We present \rev{SPARCS}: a real-to-sim scene estimation framework that reconstructs physically consistent, simulation-ready object shapes and poses from sparse observations in cluttered environments. Our core contribution is a joint physics-constrained optimization formulated directly in the coupled shape-pose space. Building on the novel SDRS contact model, we enforce quasistatic force equilibrium and develop a structure-aware linear solver that enables efficient and stable optimization for large-scale scenes with many interacting objects. Integrated into a pipeline with learning-based initialization and differentiable texture refinement, our method robustly produces physically valid, simulation-ready reconstructions \rev{that can be directly used in the downstream robotic manipulation tasks}.

Our method opens doors to several avenues of future work. The major limitation of our method is the high computational cost due to the extended decision variables parameterizing object shapes. We plan to utilize GPU to further reduce the computational overhead. \rev{In this work, parameters like total mass and friction coefficient are assigned manually as they are inherently ambiguous for static scenes without dynamic interaction data. Exploiting priors from large Vision-Language Models for these parameters could make our pipeline more comprehensive. Though our focus is on the most challenging setup of single-view input, our optimization formulation can directly generalize to multi-view guidance, with the more recent learning-based initializations \cite{li2026mv, zadaianchuk2026reconstruction} that support multi-view input. We emphasize that the single-view setting is the most challenging problem, and in highly cluttered scenes,
occlusions and complex contact reasoning always exist even
with multiple views. Finally}, the SAM3D estimated object shapes can be rather inaccurate in cluttered scenes with severe occlusions. In future works, we plan to enable image-guided end-to-end real-to-sim optimization \rev{and exploit other shape prior models to reduce the reliance on the full mesh-based initial guess}. 
%

%% file: camera_ready/Appendix_A_Optimization.tex
\section{\label{appen:optimization_details}Optimization Details}
\subsection{Complete ALM formulation}
In this section, we expand the frictional ALM subproblem in \prettyref{eq:ALM-subproblem-friction} by explicitly defining the per-contact augmented term $\Phi(q,x,f_{iji'j'})$, and illustrate our complete algorithmic flow to solve the ALM problem. Recall that for each interacting convex-hull pair $\TWO{ij}{i'j'}$ and each vertex $X_{ijk}$, the normal force $f^\perp_{ijk,i'j'}$ is a function of shape $x$ and pose $q$ according to \prettyref{eq:perp}, and we introduce per-vertex tangential friction force variables $f^\parallel_{ijk,i'j'}$. We use the vector $f_{iji'j'}^\parallel$ to denotes all force terms $\{f^\parallel_{ijk,i'j'}\}_k$ stacked together. For brevity, we denote $z \triangleq \FIVE{q}{x}{\cdots}{f_{iji'j'}^\parallel}{\cdots}$. With such notation, our formulation consists of 4 kinds of frictional contact constraints for our scene estimation problem:
\begin{enumerate}[leftmargin=*]
\item Force equilibrium for objects:
\begin{equation*}
\ResizedEq{
C_{\text{equi}}(z) \triangleq \bar{C}(z)
=\nabla_q\left[\Psi(q,x)-\sum_{i\neq i'}\sum_{j,j'}\sum_k\langle X_{ijk},f_{ijk,i'j'}^\parallel\rangle\right]=0,}
\end{equation*}
\item Orthogonality:
\begin{equation*}
C_{\text{orth}}^{iji'j'}(z) \triangleq \THREEC{\vdots}{\left< f^\perp_{ijk,i'j'},\, f^\parallel_{ijk,i'j'}\right>}{\vdots}= 0 ,
\end{equation*}
\item Friction cone:
\begin{equation*}
C_{\text{cone}}^{iji'j'}(z) \triangleq
\THREEC{\vdots}{\|f^\parallel_{ijk,i'j'}\| - \eta\|f^\perp_{ijk,i'j'}\|}{\vdots} \le 0,
\end{equation*}
\item Tangential force equilibrium for separating plane:
\begin{equation*}
\ResizedEq{
C_\text{plane}^{iji'j'}(z) \triangleq 
\TWOC{
\sum_{k} f^\parallel_{ijk,i'j'} + \sum_{k'} f^\parallel_{i'j'k',ij}
}
{
T_{iji'j'}\left[\sum_{k} X_{ijk}\!\times f^\parallel_{ijk,i'j'} + \sum_{k'} X_{i'j'k'}\!\times f^\parallel_{i'j'k',ij}\right]}=0
}.
\end{equation*}
\end{enumerate}
Note that we redefine $\bar{C}$ as $C_{\text{equi}}$ for more conveniently distinguishing different types of constraints. We then compactly group the constraints as follows:
\begin{align*}
C_{\text{eq}}(z)&\triangleq
\THREEC{C_{\text{equi}}(z)}
{\{C_{\text{orth}}^{iji'j'}(z)\}_{ij,i'j'}}
{\{C_{\text{plane}}^{iji'j'}(z)\}_{ij,i'j'}}\\
C_{\text{ineq}}(z)&\triangleq
\{C_{\text{cone}}^{iji'j'}(z)\}_{ij,i'j'}.
\end{align*}

To handle inequality constraint in the Augmented Lagrangian formulation, following \cite{bertsekas1997nonlinear}, we introduce the element-wise clamp operation by defining $[u]_+ \triangleq \max(u,0)$ and $\hat{C}_{\text{ineq}}(z) \triangleq [C_{\text{ineq}}(z)]_+$. Let $\lambda_{\text{eq}}$ and $\lambda_{\text{ineq}}$ denote Lagrange multipliers for equality and inequality constraint, respectively. We further define the corresponding penalty parameter be $\rho_\text{eq}>0$ and $\rho_\text{ineq}>0$. (We define similarly for constraint groups $\TWO{\lambda_\text{equi}}{\rho_\text{equi}}$,  $\TWO{\lambda_\text{orth}^{iji'j'}}{\rho_\text{orth}^{iji'j'}}$ etc.). The complete Augmented Lagrangian subproblem takes the following form:
\begin{align}
\label{prob:full_alm}
\argmin{z}
\begin{cases}
O(z)+\\ 
\lambda_{\text{eq}}^T C_{\text{eq}}(z) + \frac{\rho_{\text{eq}}}{2}\|C_{\text{eq}}(z)\|^2+\\
\lambda_{\text{ineq}}^T \hat{C}_{\text{ineq}}(z) + \frac{\rho_{\text{ineq}}}{2}\|\hat{C}_{\text{ineq}}(z)\|^2.
\end{cases}
\end{align}
For each interacting pair $\TWO{ij}{i'j'}$ of convex hulls, define the pairwise equality and inequality constraint:
\begin{equation*}
C_{\text{eq}}^{iji'j'}(z)\triangleq
\TWOC{C_{\text{orth}}^{iji'j'}(z)}
{C_{\text{plane}}^{iji'j'}(z)}
\qquad
\hat C_{\text{ineq}}^{iji'j'}(z)\triangleq [C_{\text{cone}}^{iji'j'}(z)]_+.
\label{eq:Cpair_def}
\end{equation*}
Then \prettyref{prob:full_alm} can be equivalently written as:
\begin{equation*}
\mathcal{L}(z)= O(z) +\lambda_\text{equi}^T C_\text{equi}(z)+\frac{\rho_\text{eq}}{2}\|C_\text{equi}(z)\|^2+\sum_{ij,i'j'}\Phi^{iji'j'}(z),
\label{eq:AL_phi}
\end{equation*}
which is exactly \prettyref{eq:ALM-subproblem-friction} in our main paper with the per-pair augmented term:
\begin{align}
\ResizedEq{
\Phi^{iji'j'}(z)\triangleq
\begin{cases}
(\lambda_{\text{eq}}^{iji'j'})^T C_{\text{eq}}^{iji'j'}(z)
+\frac{\rho_{\text{eq}}}{2}\|C_{\text{eq}}^{iji'j'}(z)\|^2+\\
(\lambda_{\text{ineq}}^{iji'j'})^T C_{\text{ineq}}^{iji'j'}(z)
+\frac{\rho_{\text{ineq}}}{2}\|\hat C_{\text{ineq}}^{iji'j'}(z)\|^2.
\end{cases}}
\label{eq:phi_def_clean}
\end{align}

Still, since our visual objective takes a least-square form, i.e., we can rewrite $O(z)$ by finding $r_O(z)$ satisfying $\|r_O(z)\|^2 = O(z)$, solving the complete ALM \prettyref{prob:full_alm} is equivalent to minimizing $\|r(z)\|^2$ with:
\begin{equation*}
r(z) \triangleq 
\THREEC{r_O(z)}
{\sqrt{\rho_{\text{eq}}}\Big(C_{\text{eq}}(z)+\lambda_{\text{eq}}/\rho_{\text{eq}}\Big)}
{\sqrt{\rho_{\text{ineq}}}\Big(\hat{C}_{\text{ineq}}(z)+\lambda_{\text{ineq}}/\rho_{\text{ineq}}\Big)},
\end{equation*}
and we solve the least-square problem using the LM algorithm with our structure aware linear solver (\prettyref{alg:woodbury} in the main paper) that exploit the low rank and pairwise sparsity. The ALM solving process is summarized in~\prettyref{alg:alm_friction_complete_appendix} below.

\begin{algorithm}[ht]
\caption{ALM-based Joint Shape and Pose Optimization}
\label{alg:alm_friction_complete_appendix}
\begin{algorithmic}[1]
\Require{Initial guess $z\triangleq (x,q,f_{iji'j'}^\parallel)$; initial multiplier $\lambda_{\text{eq}}, \;\lambda_{\text{ineq}}$; initial penalty $\rho_{\text{eq}},\rho_{\text{ineq}}\in(0,\infty)$; 
$\gamma_{\text{eq}},\gamma_{\text{ineq}}\in(0,1)$;
$\beta_{\text{eq}},\beta_{\text{ineq}}\in(1,\infty)$}
\Ensure{Locally optimal $z$}

\State $O^{\text{prev}}\gets\infty$
\While{Not converged}
    \LineComment{ICP-type closest point update}
    \For{Each convex hull vertex $X_{ijk}$}
    \State Compute and fix $p(X_{ijk})$ (\prettyref{eq:closest-on-mesh})
    \EndFor
    \For{Each $\rev{p_{i}}\in\mathcal{P}_i$ and $\rev{p_{i}}\in\mathcal{M}_i$}
    \State Compute $X(\rev{p_{i}})$ (\prettyref{eq:closest-on-convex-hull})
    \State Compute $\rev{\Delta_{i}}$ by fixing $w_k$ (\prettyref{eq:X_il-prev})
    \EndFor
    \State Evaluate $O(z)$ (\prettyref{eq:objective})
    \LineComment{Heuristic to ensure function value decrease}
    \While{$O(z)>O^\text{prev}$}
    \State Find $X(\rev{p_{i}})$ with largest $\rev{\Delta_{i}}$
    \State Exclude $\|X(\rev{p_{i}})-\rev{p_{i}}\|^2$ from $O$
    \EndWhile
    \State $z^{\text{prev}}\gets z$
    \LineComment{LM method with linear solver (\prettyref{alg:woodbury})}
    \State Solve~\prettyref{prob:full_alm} to update $z$
    \LineComment{Update multiplier as in \cite{bertsekas1997nonlinear}}
    \State $\lambda_{\text{eq}} \gets \lambda_{\text{eq}} + \rho_{\text{eq}}\, C_{\text{eq}}$, $\lambda_{\text{ineq}} \gets \lambda_{\text{ineq}} + \rho_{\text{ineq}}\, \hat{C}_{\text{ineq}}$
    \LineComment{Schedule penalty as in \cite{bertsekas1997nonlinear}}
    \If{$\|C_{\text{eq}}(z)\|_\infty\geq \gamma_{\text{eq}}\|C_{\text{eq}}(z)\|_\infty$}
        \State $\rho_{\text{eq}}\gets \beta_{\text{eq}}\,\rho_{\text{eq}}$
    \EndIf
    \If{$\|\hat{C}_{\text{ineq}}(z)\|_\infty\geq \gamma_{\text{ineq}}\|\hat{C}_{\text{ineq}}(z)\|_\infty$}
        \State $\rho_{\text{ineq}}\gets \beta_{\text{ineq}}\,\rho_{\text{ineq}}$
    \EndIf

    \State $O^{\text{prev}}\gets O(z)$
\EndWhile
\end{algorithmic}
\end{algorithm}

We use a unified optimization parameter setting across all the scenarios illustrated in the main paper and~\prettyref{appen:extended_results}. We set visual objective weights to be $w_1 = 2 \times 10^{-2}, w_2= 10^{-1}, w_3 = 2 \times 10^{-2}$, the complementarity gap is chosen to be a small value $\mu=5\times10^{-5}$ and the frictional coefficient $\eta=1$. We terminate the subproblem LM solver when $\|r(z)\|_\infty\leq\epsilon_r$ or $\|r^T\FPPR{r}{z}\|_\infty\leq\epsilon_g$. We find that, due to the physics constraint being very stiff, leading to the subproblem solver occasionally make very small progress over many iterations. We thus terminate the inner loop when the progress is less than $1\%$ over $20$ iterations. Finally, we terminate the outer ALM iteration when $\|C_\text{eq}\|_\infty\leq\epsilon_C{_\text{eq}}$ and $\|C_\text{ineq}\|_\infty\leq\epsilon_C{_\text{ineq}}$ or the residual of the KKT condition does not improve by more than $1\%$ over consecutive ALM iterations. Through all our experiments, we choose parameters $\rho_\text{eq}^\text{init}=10^{-2}, \rho_\text{ineq}^\text{init}=2, \lambda_\text{eq}^{\text{init}}=\lambda_\text{ineq}^{\text{init}}=0, \gamma_\text{eq}=\gamma_\text{ineq}=0.25, \beta_\text{eq}=8, \beta_{\text{ineq}}=4, \epsilon_r=10^{-6}, \epsilon_g= 10^{-2}, \epsilon_C{_\text{eq}} = \epsilon_C{_\text{ineq}} =5 \times 10^{-4}$.

\begin{figure}[ht]
\centering
\includegraphics[width=\linewidth]{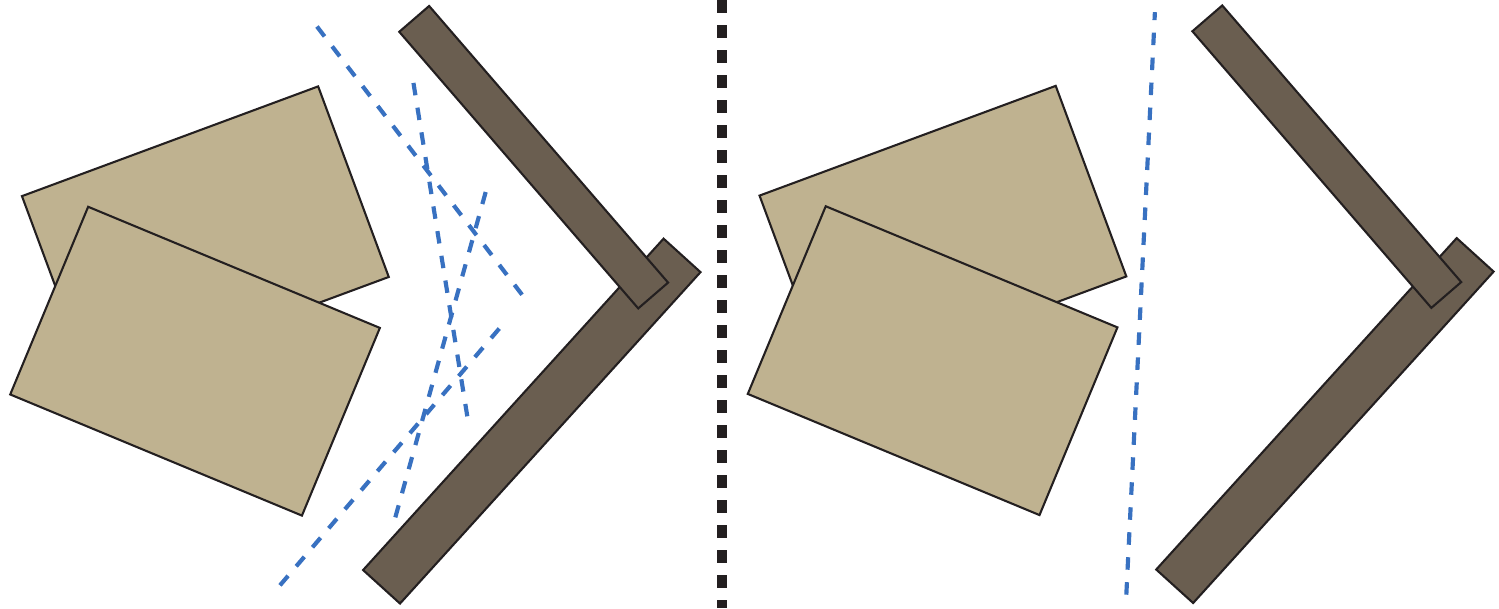}
\caption{\label{fig: plane aggregation}Two non-convex objects separated using SDRS contact model, each consisting of two convex hulls. Left: Without aggregation, we need 4 separating planes. Right: With aggregation, we only introduce a single separating plane.}
\end{figure}
\subsection{Separating Plane Aggregation}
Our scene estimation problem is inherently ill-posed: we seek to recover complete 3D object shapes from a single RGB-D observation. Addressing this challenge requires introducing a large number of decision variables, which in turn gives rise to numerous feasible local minima. As a result, the optimizer is free to converge to arbitrary solutions, particularly in occluded regions where shape evidence is absent. In practice, this freedom is often undesirable and commonly manifests as noisy or irregular geometry on occluded contact surfaces.
To mitigate this issue, we propose separating plane aggregation, a simple yet effective regularization technique that encourages the recovery of smooth object shapes. Although the contact model of~\cite{ye2025sdrs} conceptually inserts a separating plane between each pair of convex hulls, this assumption is not strictly necessary. In fact, a separating plane can be introduced between arbitrary non-convex shapes, in which case SDRS implicitly treats each shape as its convex hull.
Leveraging this observation, we introduce a single separating plane between each pair of interacting objects, rather than multiple planes between all pairs of their constituent convex hulls, as shown in Figure \ref{fig: plane aggregation}. This aggregation effectively regularizes occluded contact regions by encouraging them to align with a common separating plane, resulting in smoother and more physically plausible contact surfaces. 
We emphasize that, even with the separating plane aggregation, we are still representing each object as multiple convex hulls and the vertices of each convex hull is treated as separate decision variables to be optimized. This is necessary to match arbitrarily non-convex object shapes.

%% file: camera_ready/Appendix_B_Preprocess.tex
\section{\label{appen:geometry_details}Visual Inference \& Geometry Process}
In this section, we explain the details of the visual inference pipeline and geometry processing for initialization of our physics-aware optimization.
\begin{figure*}[t]
\centering
\setlength{\tabcolsep}{1px}
\begin{tabular}{ccccc}
\includegraphics[width=0.19\linewidth]{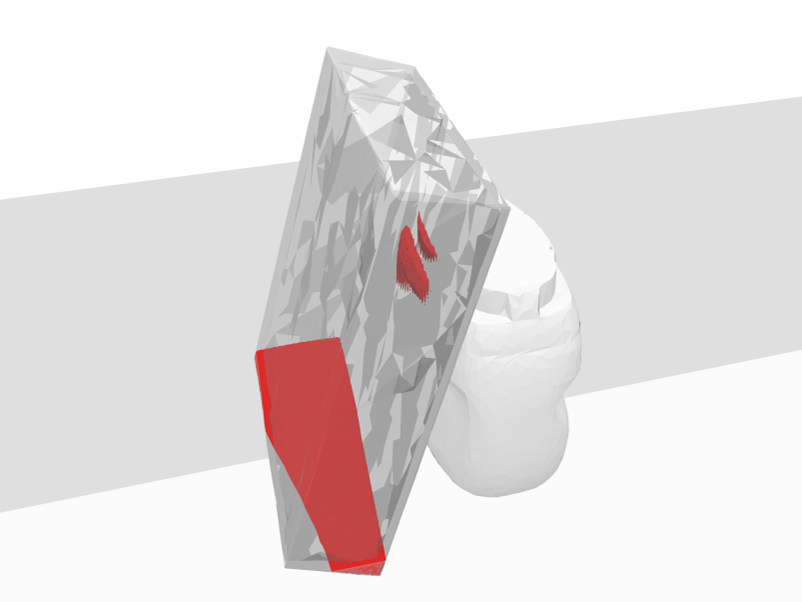}&
\includegraphics[width=0.19\linewidth]{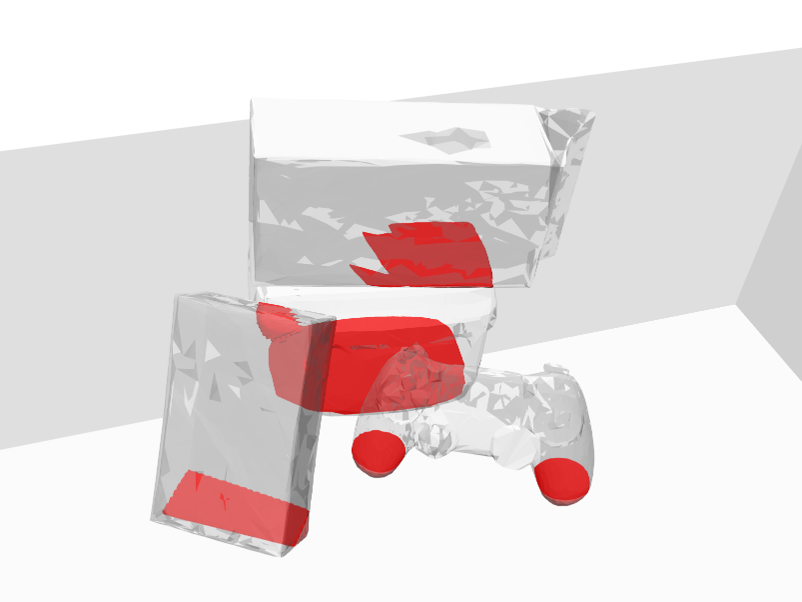}&
\includegraphics[width=0.19\linewidth]{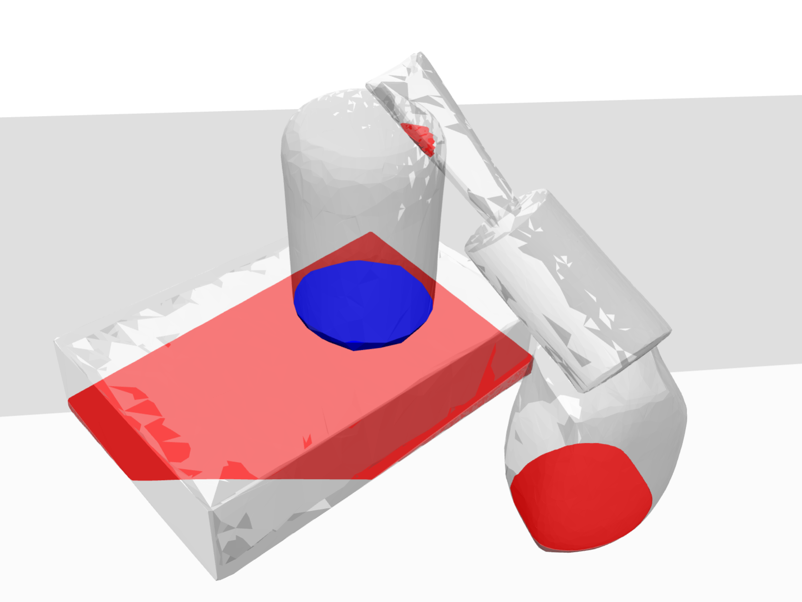}&
\includegraphics[width=0.19\linewidth]{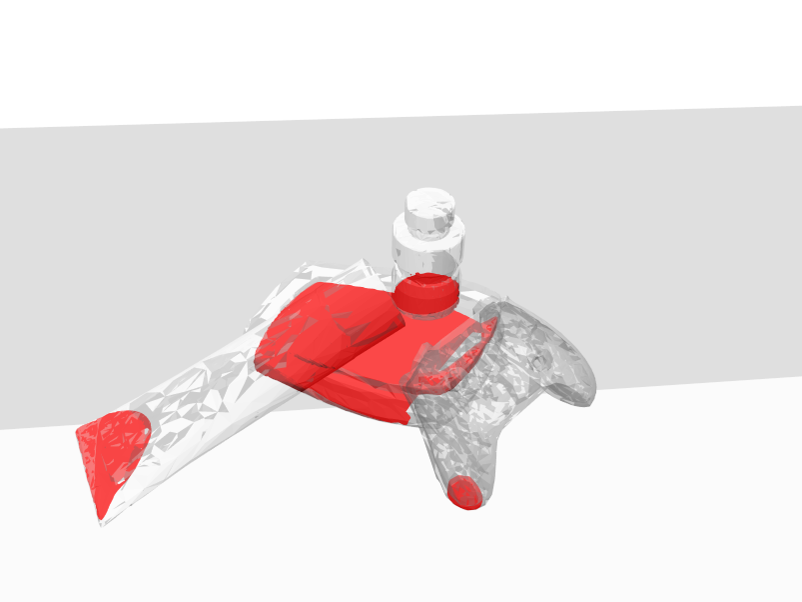}&
\includegraphics[width=0.19\linewidth]{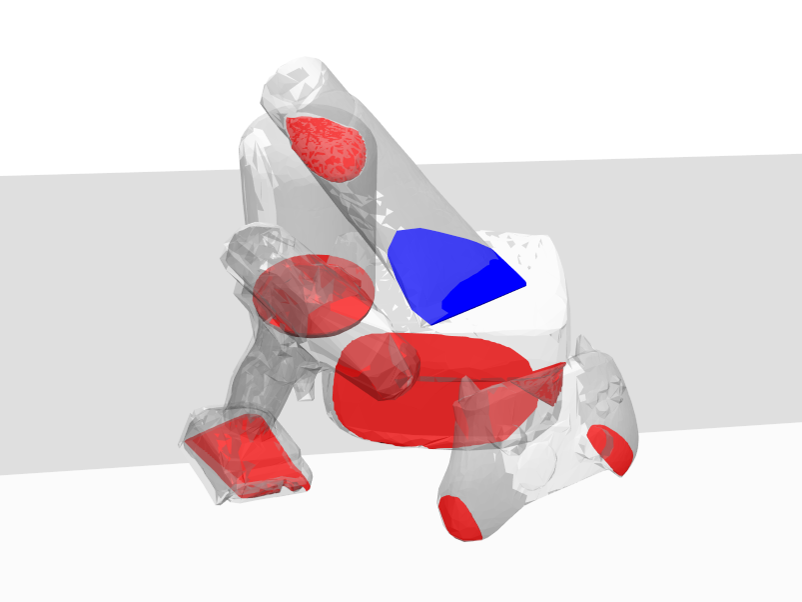}\\
\includegraphics[width=0.19\linewidth]{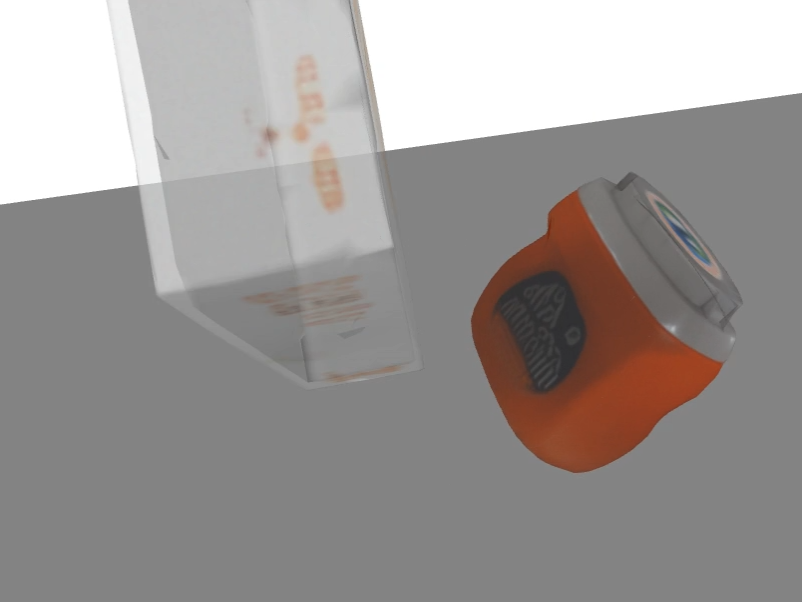}&
\includegraphics[width=0.19\linewidth]{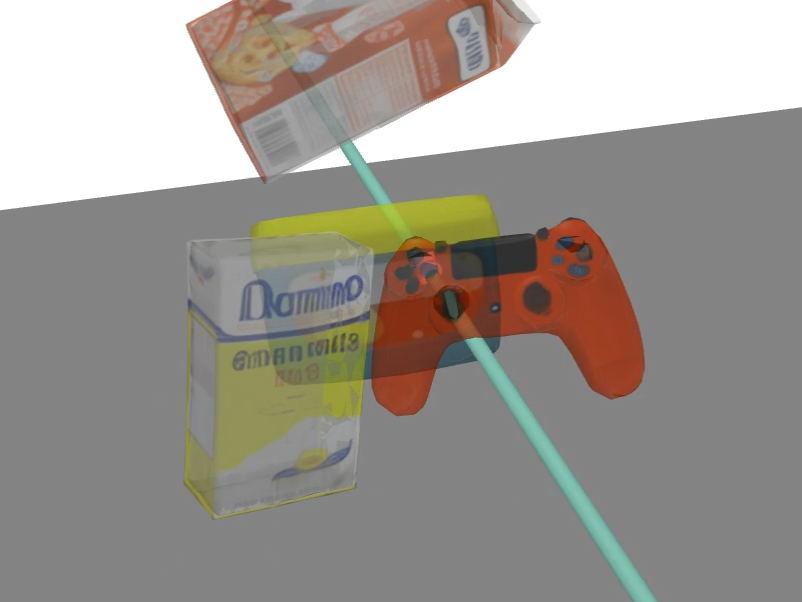}&
\includegraphics[width=0.19\linewidth]{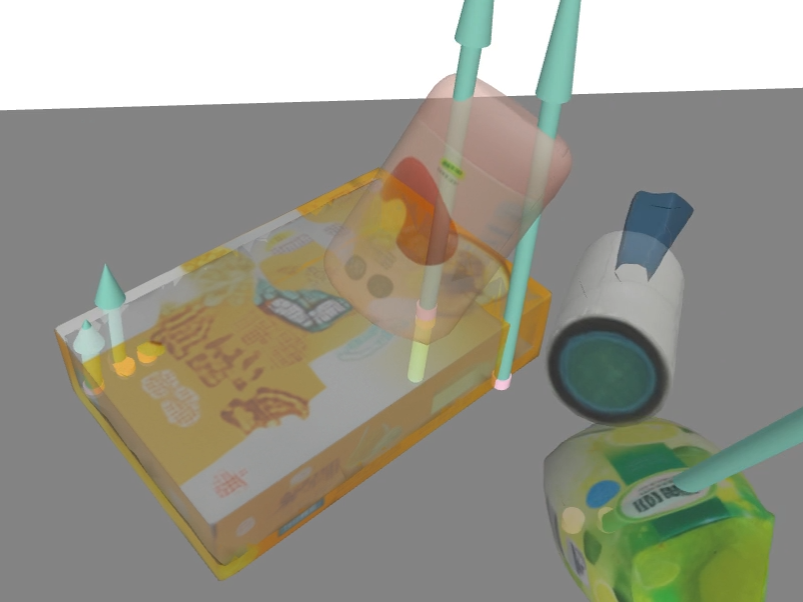}&
\includegraphics[width=0.19\linewidth]{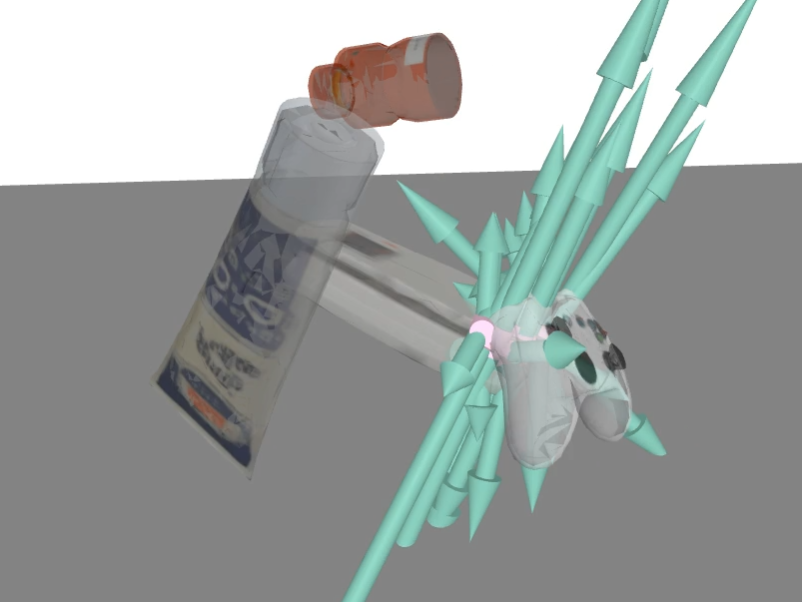}&
\includegraphics[width=0.19\linewidth]{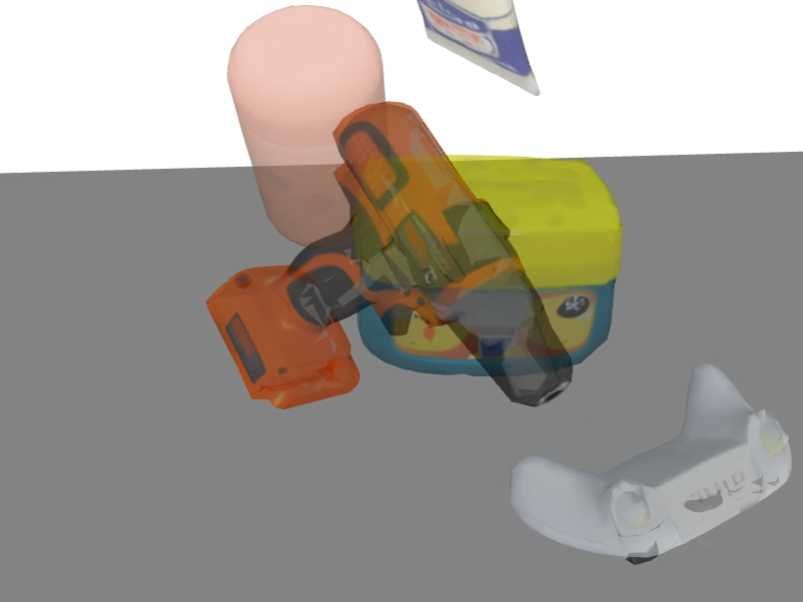}\\
\end{tabular}
\caption{\label{fig:violations_blow_up}We illustrate various violations to physical constraints induced by visual inference. Top: penetrations (red) and floating objects (blue) always occur as SAM3D and FoundationPose do not explicitly enforce physics constraints. Bottom: These violations always lead to simulation blow up in MuJoCo.}
\end{figure*}

\subsection{\label{sec:SAM3Ddetails}Visual Inference}
Given an RGBD image of the scene, we first use SAM2~\cite{ravi2024sam2} to generate the mask for each object in the scene from a user selected bounding box. We then use SAM3D~\cite{chen2025sam} to generate an initial geometry for each object, conditioned on the RGBD image and the mask. The output mesh might not be watertight and has excessively high resolution, which slows down geometric processing. Therefore, we simplify and fix the mesh using PyMeshFix~\cite{attene2010lightweight} and re-textured the mesh using xatlas~\cite{young2019xatlas} and Nvdiffrast~\cite{Laine2020diffrast}. As mentioned in the main content, the poses given by SAM3D are oftentimes inaccurate. To further correct them, we use FoundationPose~\cite{wen2024foundationpose} to register the textured mesh to the RGBD image.

The point clouds derived from the RGBD camera are prone to noisy outliers. We use the SAM3D generated mesh and the registered pose to filter the noisy segmented point cloud used in the optimization. For each object $i$, we remove any point with segmentation label $i$ if its distance to the mesh of the object $i$ is greater than $0.01$m. We include the table in the scene as a static object, i.e. it's pose and shape are not optimized but its contact interactions are accounted in constraint. We fit the upper surface of the table using the observed point cloud. The gravitational direction is assumed to be aligned with the normal of the table upper surface. We further define the body frame of each object using PCA. For each object, we compute its initial translation by the center of its vertices. The initial orientation is defined by PCA analysis extracting principal axes.

\begin{figure}[t]
\centering
\setlength{\tabcolsep}{1px}
\begin{tabular}{cccc}
\includegraphics[width=0.48\linewidth]{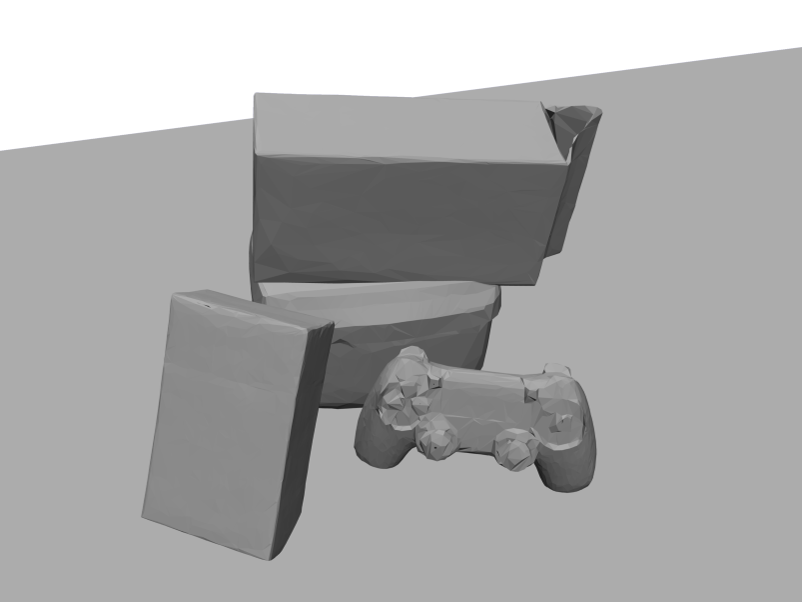}&
\includegraphics[width=0.48\linewidth]{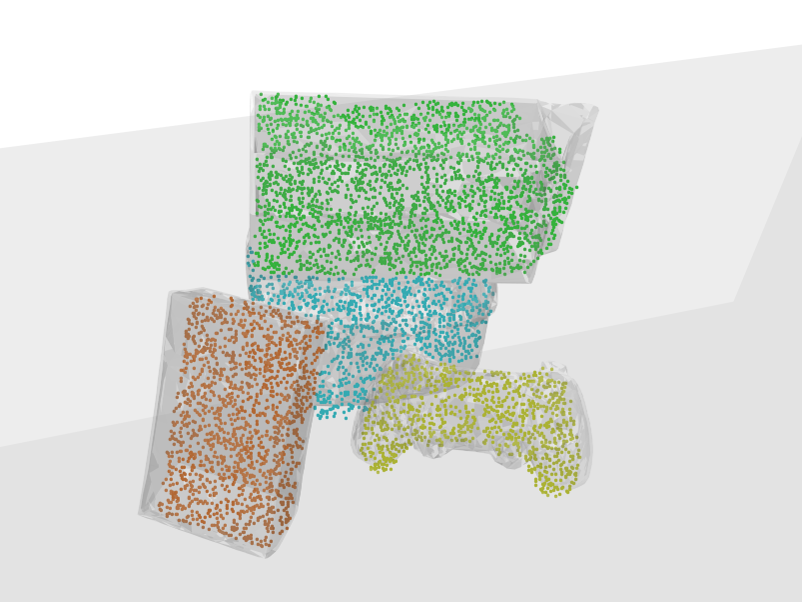}\\
\includegraphics[width=0.48\linewidth]{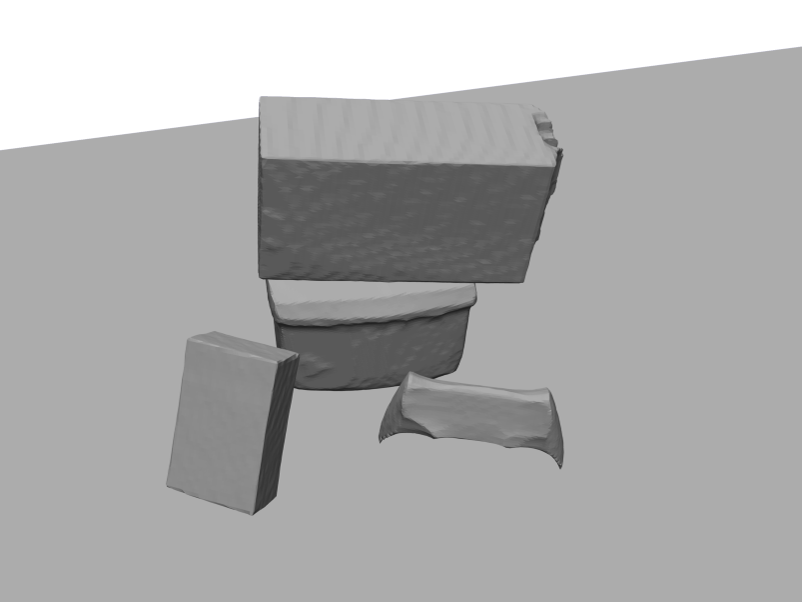}&
\includegraphics[width=0.48\linewidth]{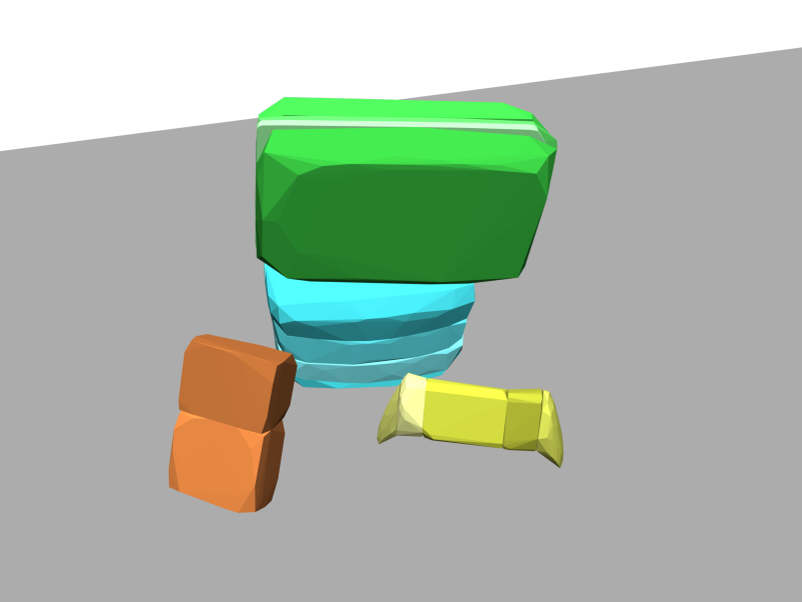}
\end{tabular}
\caption{\label{fig:shrink_mesh} An illustration of our visual inference and geometry process pipeline. Top Left: Initial estimation of each object mesh; Top Right: Filtered point cloud; Bottom Left: Penetrating free mesh after shrinkage; Bottom Right: Convex decomposition.}
\end{figure}

\begin{figure*}[t]
\centering
\setlength{\tabcolsep}{1px}
\begin{tabular}{ccccc}
\includegraphics[width=0.19\linewidth]{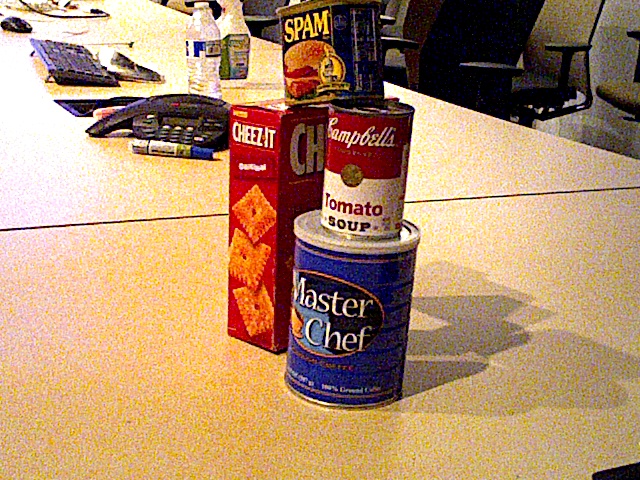}&
\includegraphics[width=0.19\linewidth]{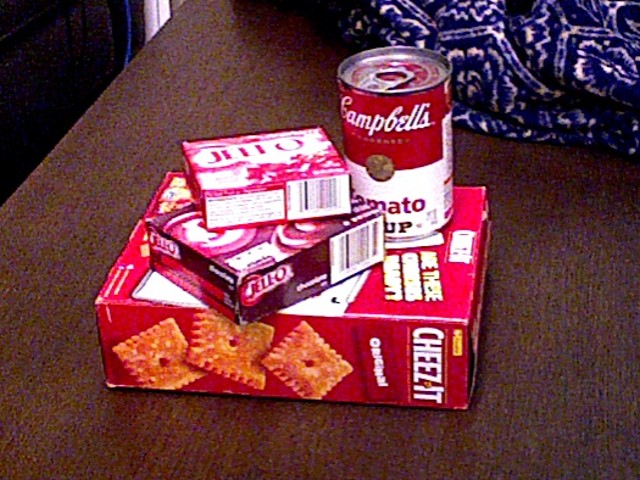}&
\includegraphics[width=0.19\linewidth]{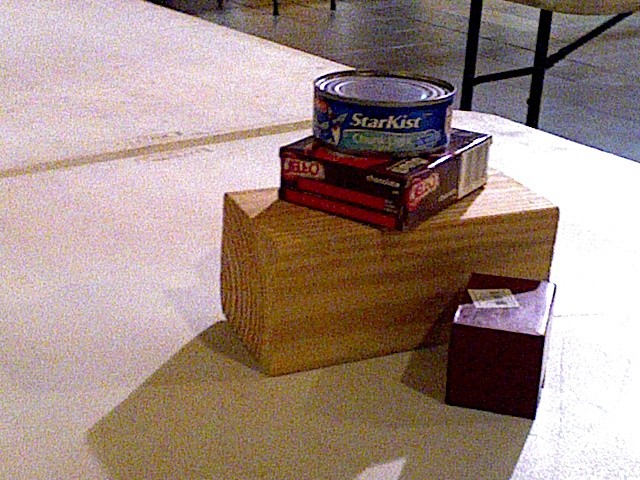}&
\includegraphics[width=0.19\linewidth]{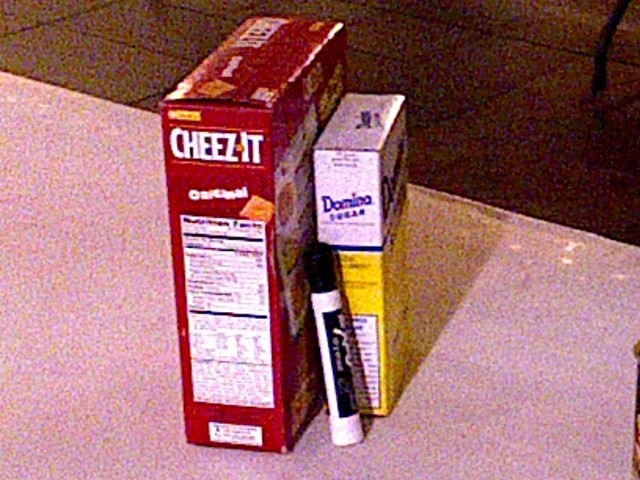}&
\includegraphics[width=0.19\linewidth]{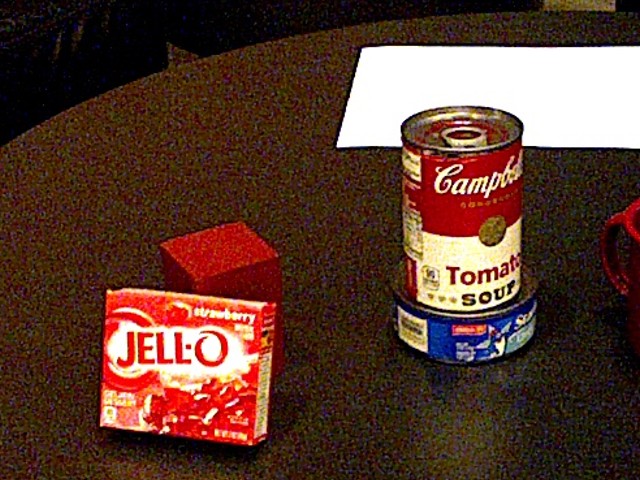}\\
\includegraphics[width=0.19\linewidth]{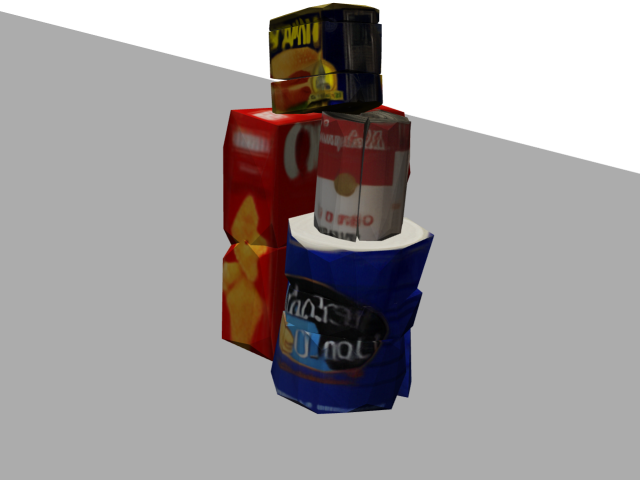}&
\includegraphics[width=0.19\linewidth]{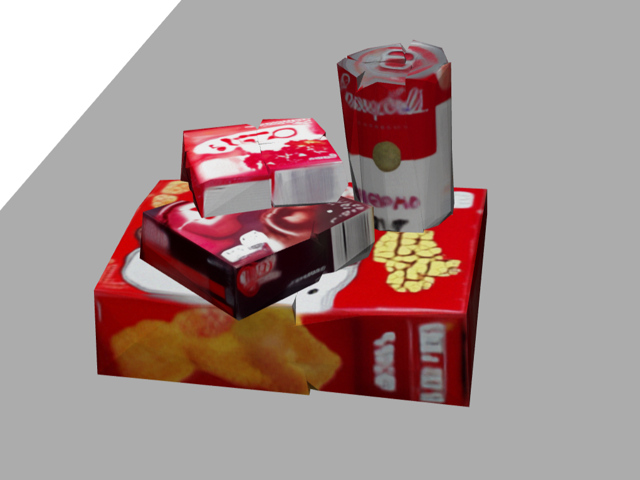}&
\includegraphics[width=0.19\linewidth]{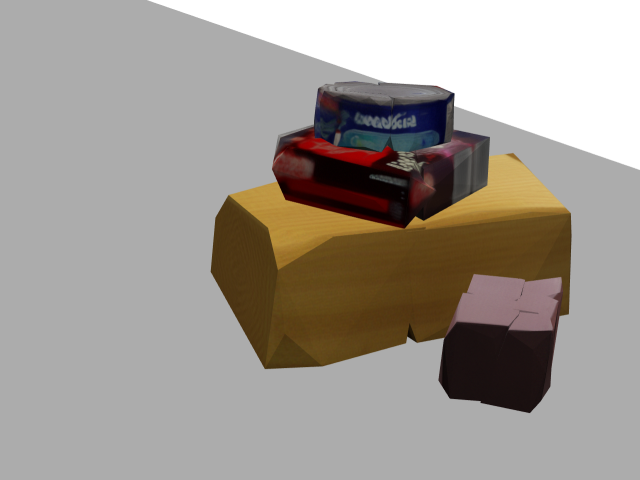}&
\includegraphics[width=0.19\linewidth]{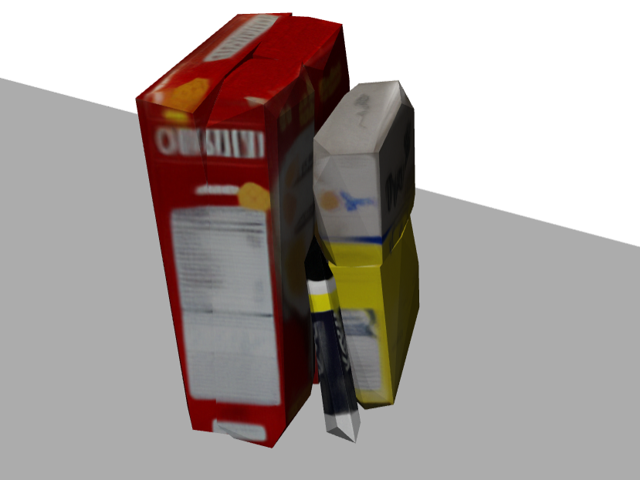}&
\includegraphics[width=0.19\linewidth]{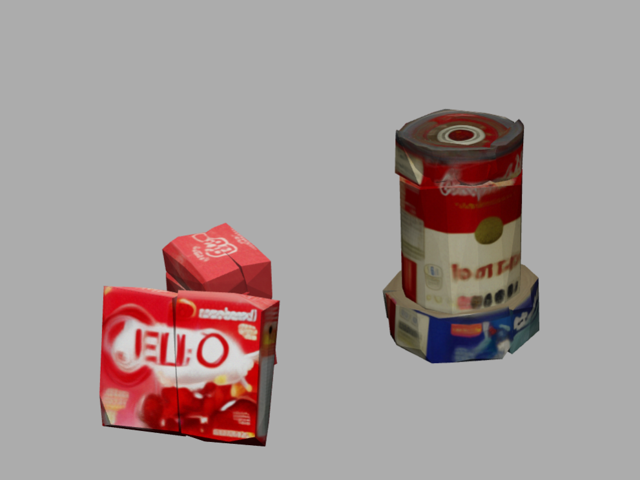}\\
\includegraphics[width=0.19\linewidth]{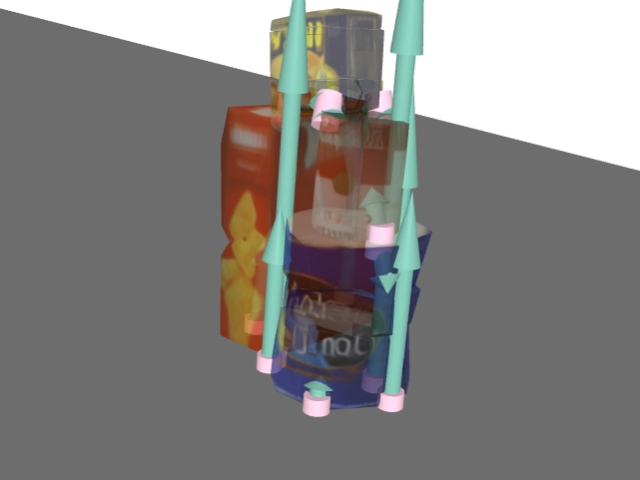}&
\includegraphics[width=0.19\linewidth]{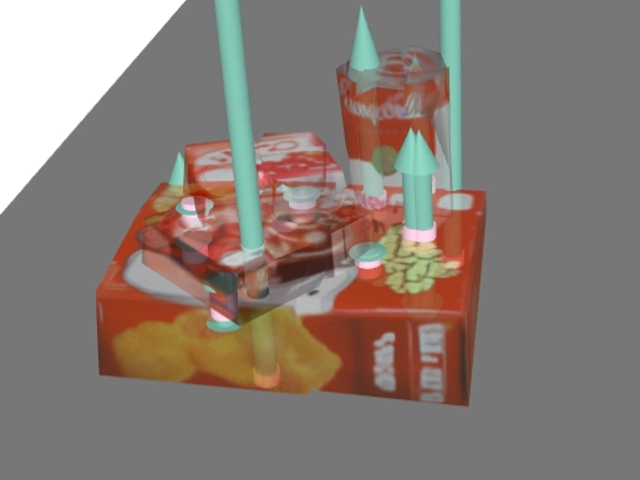}&
\includegraphics[width=0.19\linewidth]{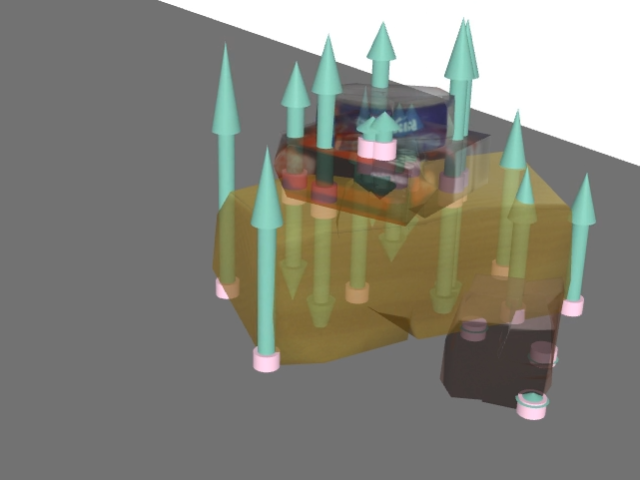}&
\includegraphics[width=0.19\linewidth]{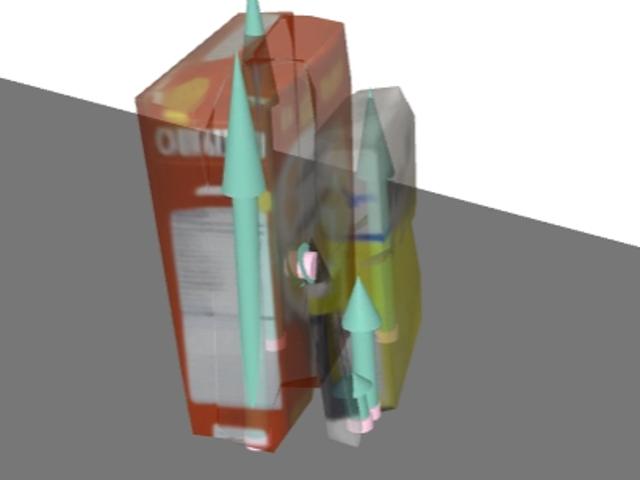}&
\includegraphics[width=0.19\linewidth]{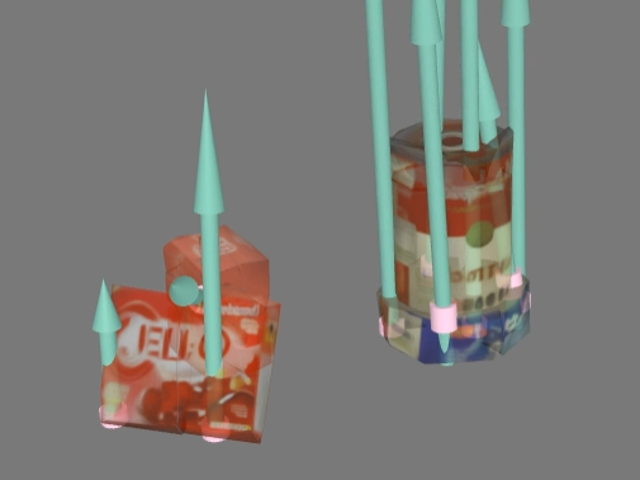}
\end{tabular}
\caption{\label{fig:extended_examples} A visualization of our additional examples on scenes from the YCB-V dataset. Top Row: The input single-view image observation. Middle Row: The estimated simulation-ready rigid bodies models. Bottom Row: Our estimated scenes achieve physical force equilibrium in MuJoCo~\cite{todorov2012mujoco}, where we visualized the contact forces.}
\end{figure*}

Since mass is ambiguous from only RGBD input, we manually assign the total mass of each object. For each object, we compute the volume of the mesh given by SAM3D and the total mass of the object is set to be the product of density and volume. The density of the objects across all examples is set to be $800$ kg/m$^3$ (the average density for materials like wood and plastic). Note that the assigned per-vertex mass is just an initial guess and our optimizer is allowed to fine-tune the mass and inertia properties.

\subsection{Penetration-free Initialization}
The SDRS contact model we adopt is based on the interior point principle, which requires a strictly feasible, i.e., penetration-free initialization. Unfortunately, this is not guaranteed by our visual inference as shown in~\prettyref{fig:violations_blow_up}. Therefore, we adopt the following process to resolve penetrations for initialization of the optimization process. For each colliding mesh pair in the initial estimate, we shrink both objects by extracting the isosurface with the SDF value $-\delta$, where $\delta>0$ is the minimum shrinkage distance for the object pair to become penetration-free. To make this process more robust for thin objects, the SDF is computed in the normalized space, where the object bounding box is normalized to $[0,1]^3$.

After the penetration is resolved, we use CoACD~\cite{wei2022coacd} to decompose the mesh of each object into unions of convex hulls. CoACD provides various options to control the fidelity of the decomposed result. For performance consideration, we set \texttt{max-convex-hull}$=5$, \texttt{max-ch-vertex}$=50$, so that each body is constrained to be decomposed into at most 5 hulls, with each hull having at most 50 vertices. The concavity \texttt{threshold}$=0.05$ which is the recommended default value to capture geometry fidelity, other parameters are set to default values as well. An illustration of the visual inference and geometry process pipeline is shown in~\prettyref{fig:shrink_mesh}.

%% file: camera_ready/Appendix_C_Extended_Results.tex
\begin{figure*}[t]
\centering
\setlength{\tabcolsep}{1px}
\begin{tabular}{cccccc}
\includegraphics[width=0.19\linewidth]{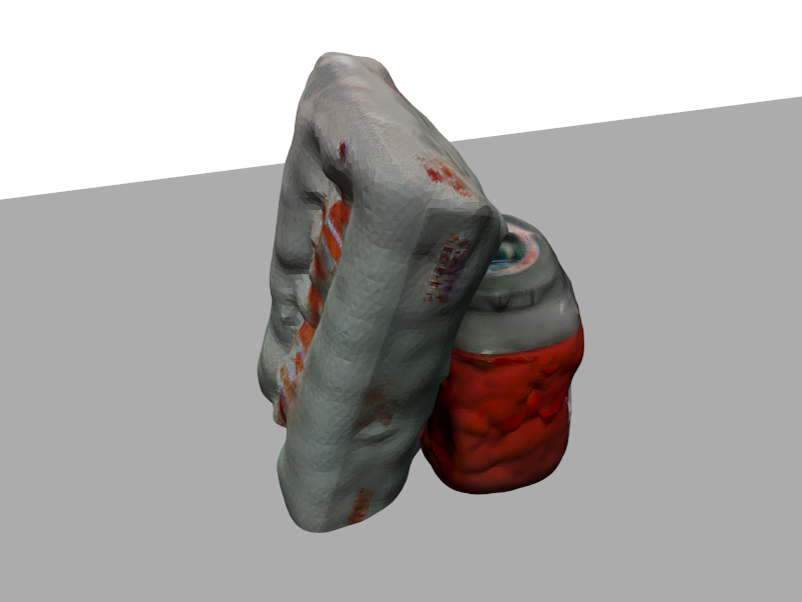}&
\includegraphics[width=0.19\linewidth]{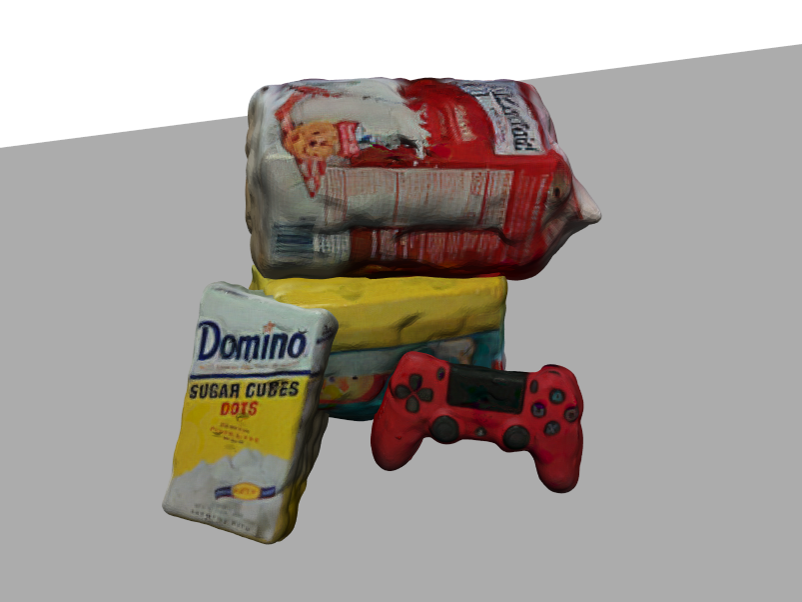}&
\includegraphics[width=0.19\linewidth]{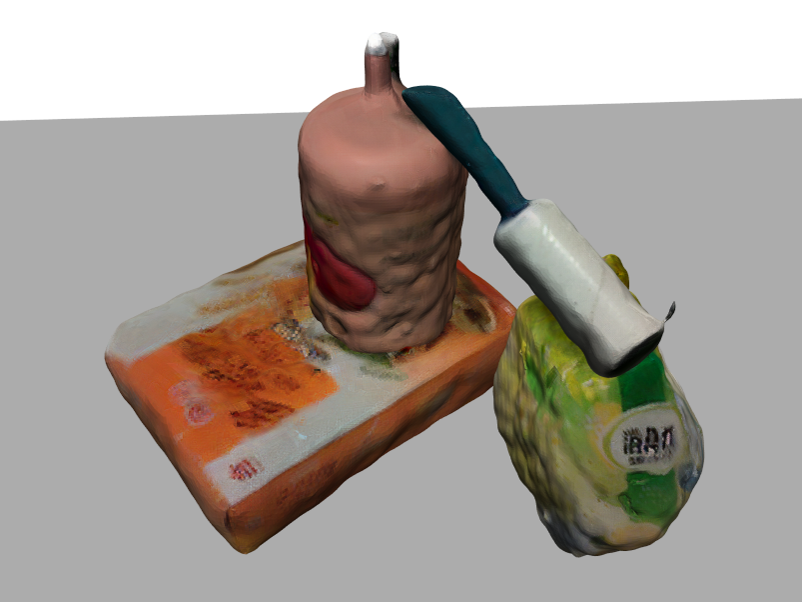}&
\includegraphics[width=0.19\linewidth]{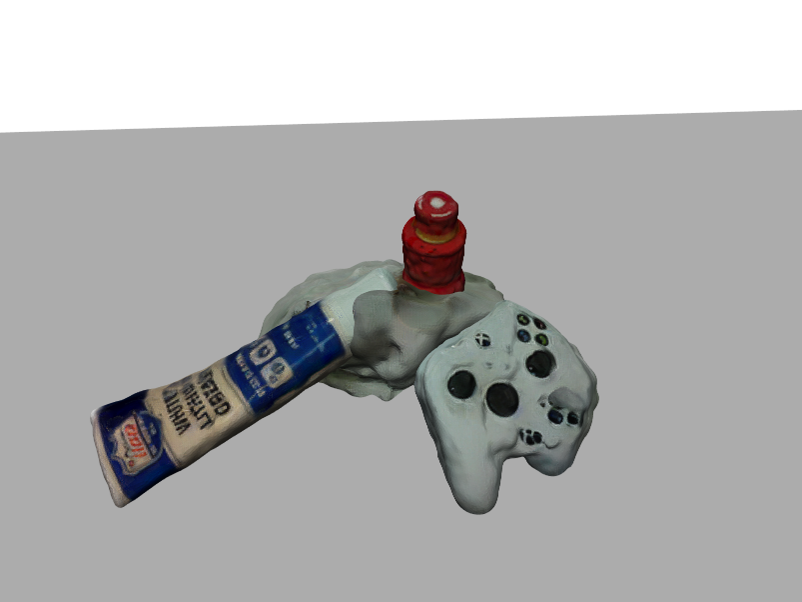}&
\includegraphics[width=0.19\linewidth]{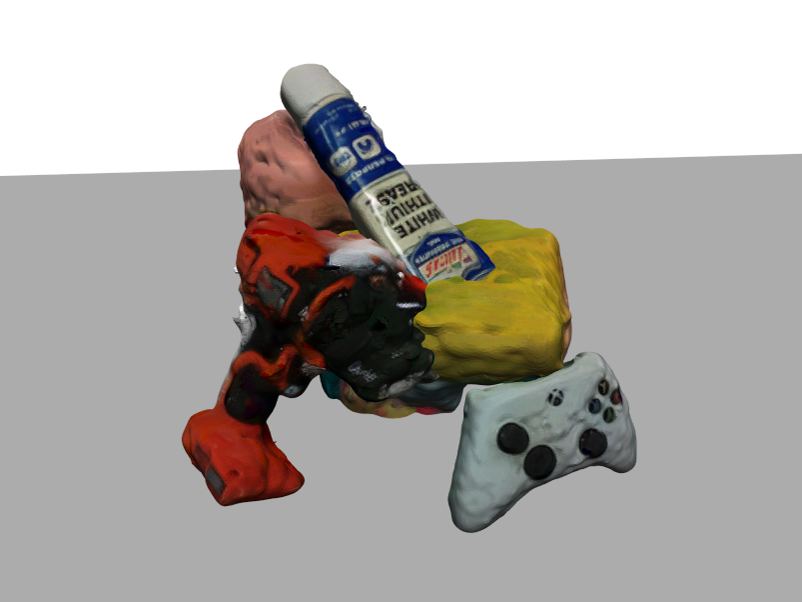}\\
\includegraphics[width=0.19\linewidth]{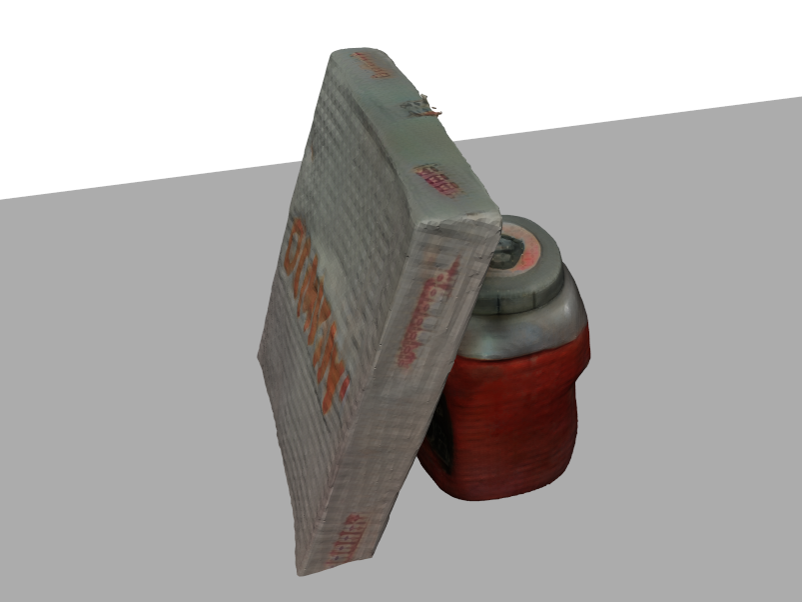}&
\includegraphics[width=0.19\linewidth]{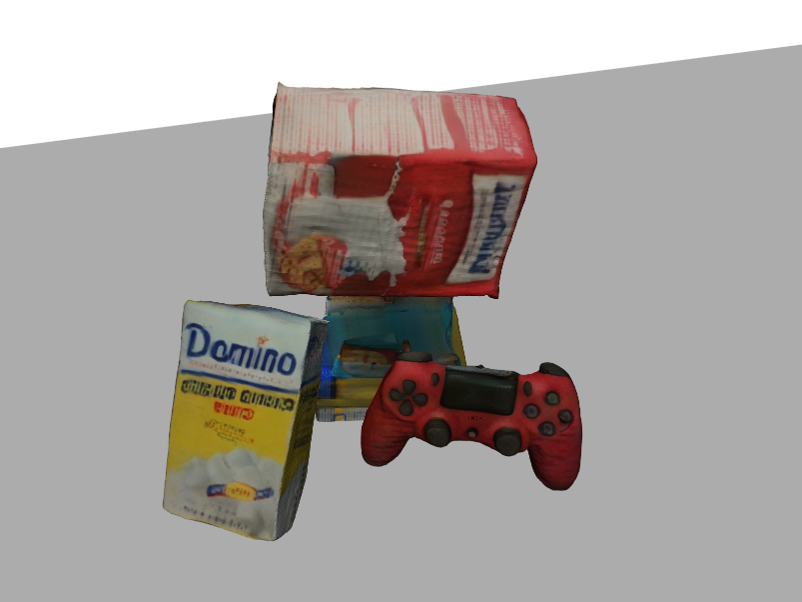}&
\includegraphics[width=0.19\linewidth]{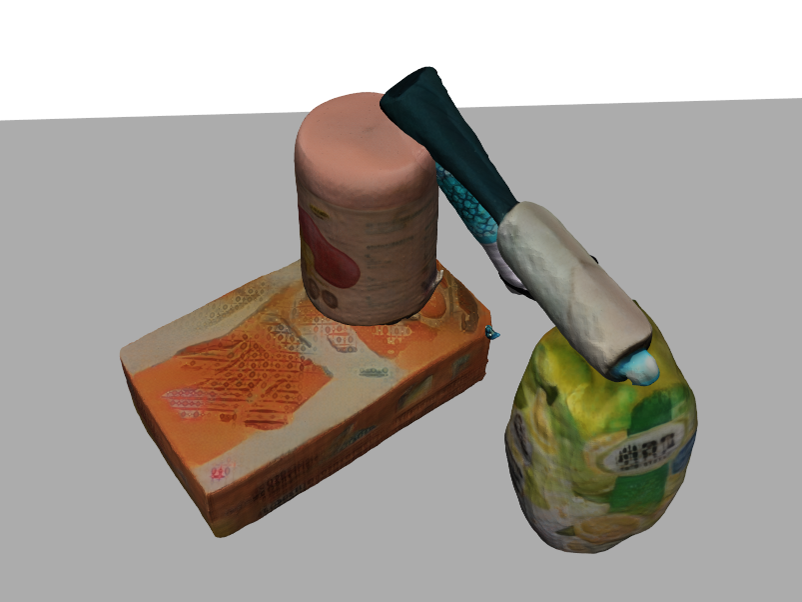}&
\includegraphics[width=0.19\linewidth]{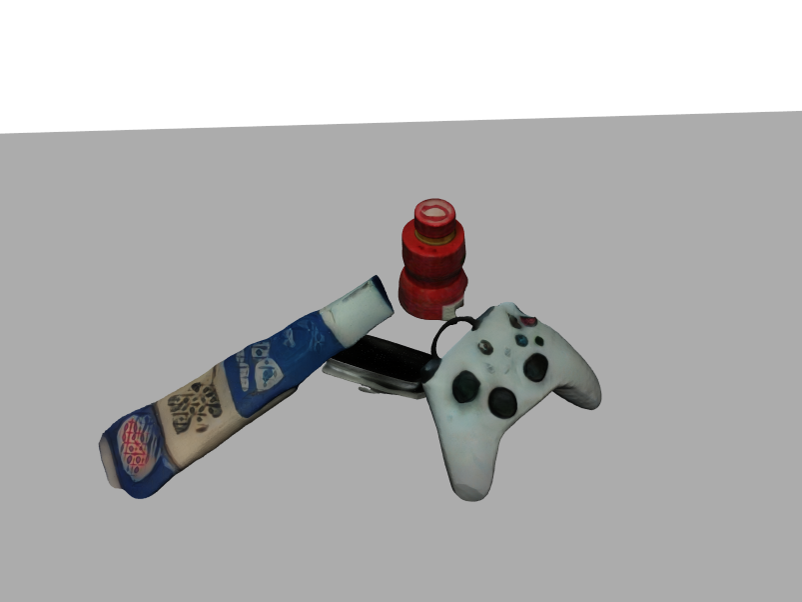}&
\includegraphics[width=0.19\linewidth]{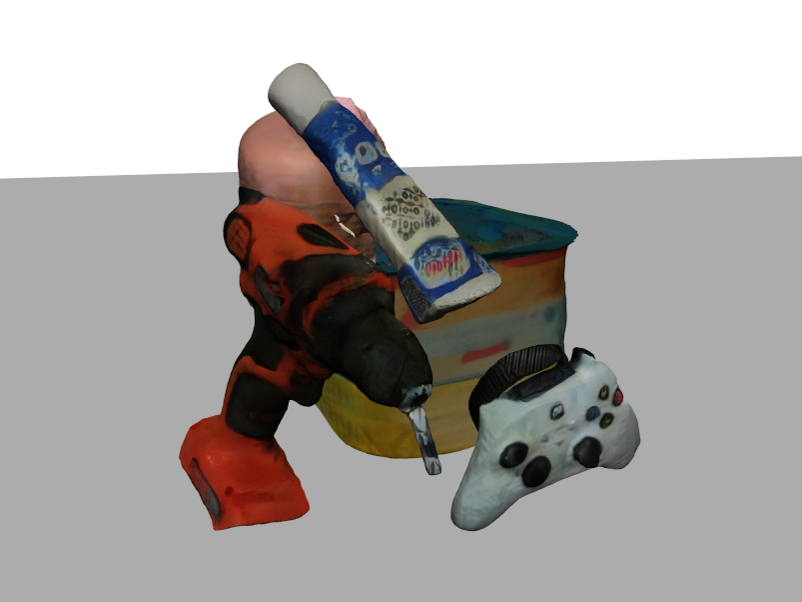}\\
\includegraphics[width=0.19\linewidth]{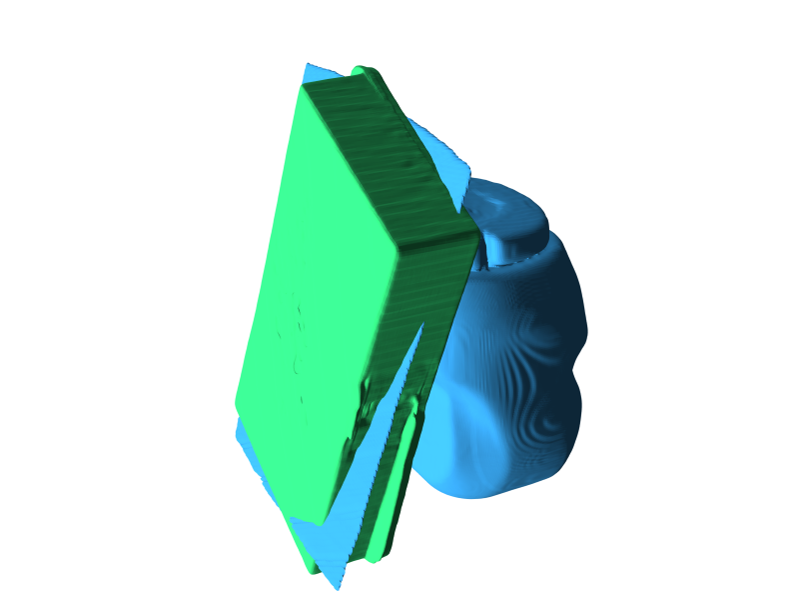}&
\includegraphics[width=0.19\linewidth]{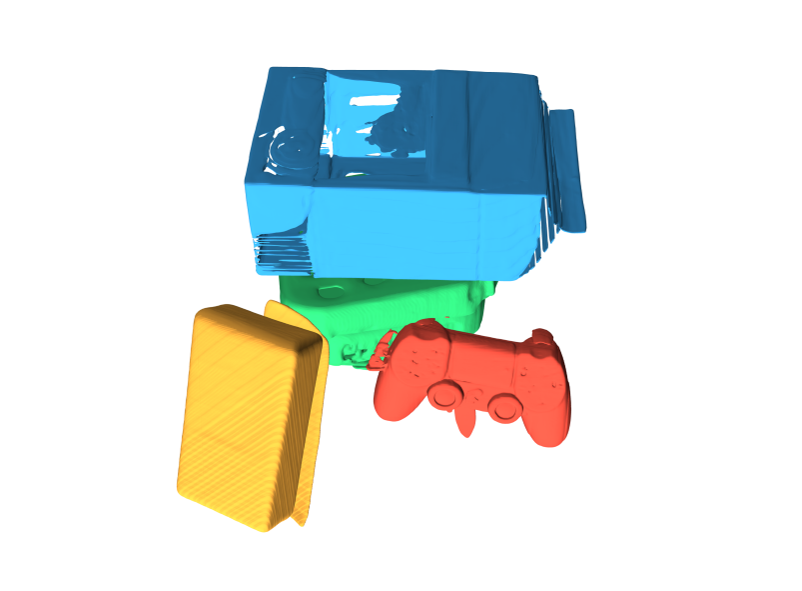}&
\includegraphics[width=0.19\linewidth]{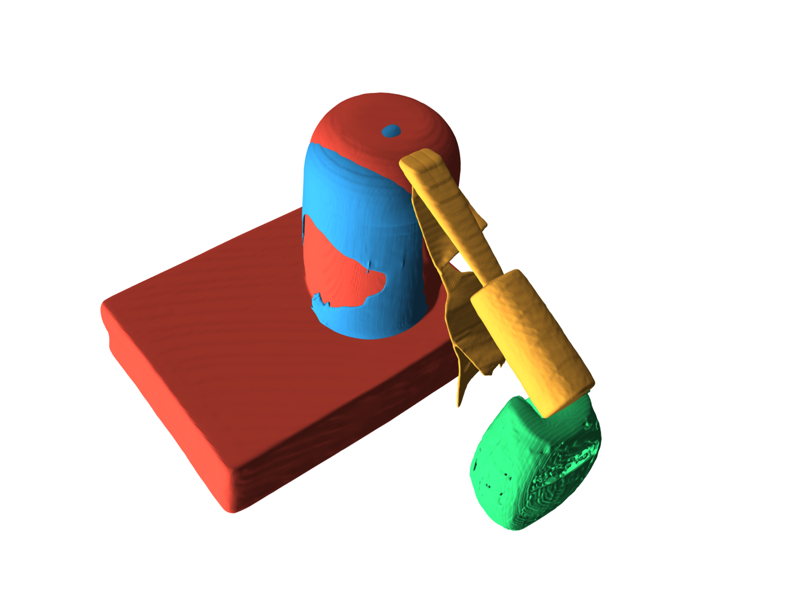}&
\includegraphics[width=0.19\linewidth]{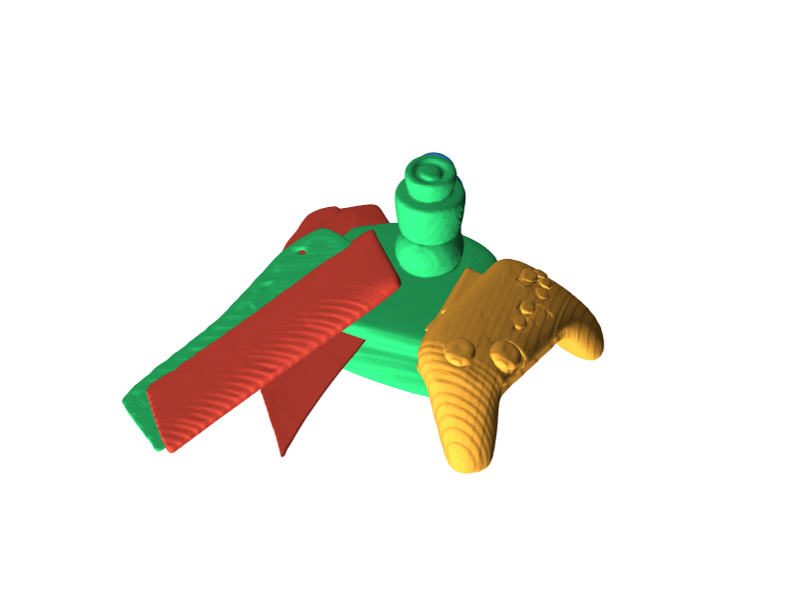}&
\includegraphics[width=0.19\linewidth]{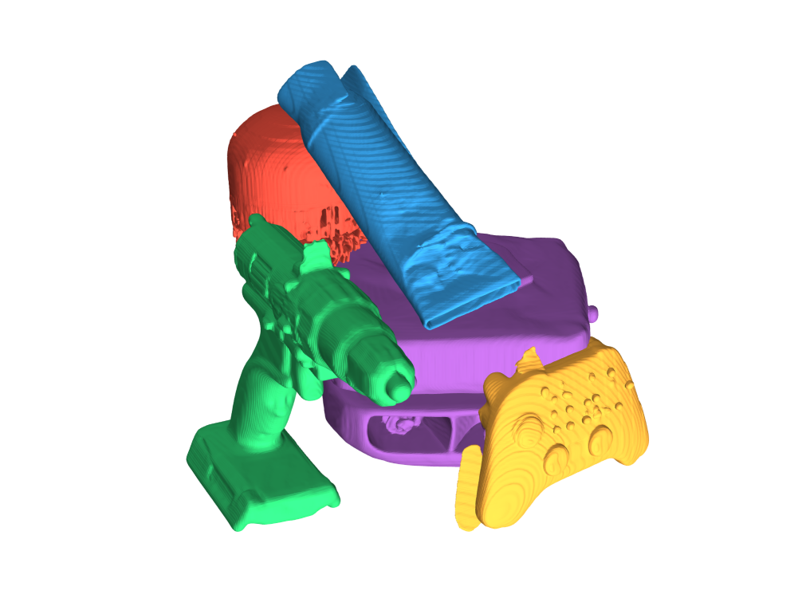}\\
\end{tabular}
\caption{\label{fig:baseline}A visualization of the baseline methods. Top row: Gen3DSR~\cite{Ardelean2025Gen3DSR}, middle row:  SceneComplete~\cite{Agarwal2026scenecomplete}, bottom row: MIDI~\cite{huang2025midi} (with different reconstructed objects visualized in different colors)}
\end{figure*}
\section{\label{appen:extended_results}Additional Results \& Baeline Comparisons}
In this section, we present our detailed MuJoCo simulator setup and present additional results.

\subsection{MuJoCo Simulation Setting}
To set up a simulation-ready scene in MuJoCo \cite{todorov2012mujoco}, we specify the simulator parameters and the states and properties of rigid bodies as follows.

\subsubsection{Simulator Parameter Setting}
For a complete list of simulator related parameters, we refer the readers to the official documentation. Here we explain the critical settings that differ from the default values: the simulator \texttt{timstep}$=10^{-3}$, and the \texttt{integrator} is set to ``implicit'' to enhance simulation stability for the cluttered scene. Moreover, MuJoCo's contact model by default induces gradual slippage phenomenon and \texttt{noslip\_iterations}$=10$ is set to prevent such slippage artifact. For the collision detection and contact related settings, we turn on \texttt{multi\_ccd} to enable the collision detection pipeline to return multiple contact points for a single collision geometry, the \texttt{cone} option is set to ``elliptic'' to match our friction cone formulation and we use \texttt{condim}$=4$, which provide additional torsional friction around contact normal. This is also recommended by the official documentation to prevent simulation instability and drifting in multi contact scenario. We set the \texttt{friction} (coefficient) to be $[1,0.005]$ to match our optimization parameter setting and only provide minimal torsional friction (default value).  

\subsubsection{Baseline Processing} 
For comparison, we simulate the scene estimated using various visual inference pipelines, including the SAM3D + FoundationPose pipeline and two baselines below. The results of these scene estimators are using mesh-based object representation. For each of the estimated rigid bodies in the scene, we replace their collision mesh by a high-fidelity convex decomposition using CoACD~\cite{wei2022coacd}. This is for fair comparison and recommended by the official documentation of MuJoCo. The visual geometry is the textured geometry returned by the pipelines. The mass and inertia tensor is auto calculated by MuJoCo using it's internal scheme with our assigned density. Internally MuJoCo performs volume based mass calculation as we do and we confirm that the numerical difference is negligible. The pose is also set based on the output of the pipelines. An example of the simulation result is shown in \prettyref{fig:violations_blow_up} bottom row.

\subsubsection{Optimized Result Processing} 
We discuss details for loading results from our optimization result to MuJoCo. For each of the rigid bodies in the scene, it's collision geometry is the unions of convex hulls extracted from the optimization result $x$, while for visual geometry, we used the triangular mesh merged from $x$ using Manifold3d Library~\cite{manifold2025} with the texture optimized by differentiable rendering~\cite{Laine2020diffrast}. The inertia tensor is set based on our mass model instead of the default auto-calculation done by MuJoCo, and we set the poses of the objects according to our optimization result $q$.

\subsection{Additional Examples}
In addition to the five self-created cluttered scenes illustrated in our main paper, we provide additional five examples to show the generality of our pipeline. Specifically, we collect five single view images from the YCB-V dataset~\cite{xiang2017posecnn}, where the selected scenarios are cluttered with complicated contact relationship. We use the same parameter setting described above. We include the performance statistic of these five examples in~\prettyref{table:extended example statistics} and the results are visualized in~\prettyref{fig:extended_examples}.
\begin{table}[ht]
\centering
\begin{tabular}{ccccccc}
\toprule
Scen. & \#Hull & \#Vertex  & \#Params & \#ALM & \#LM & Wall Time\\
\midrule
6 & 14 & 698 & 10590 & 7  & 2844 & 200.11\\
7 & 14 & 695 & 10545 & 7 & 1643 & 148.48\\
8 & 14 & 694 & 10485 & 7 & 1936 & 140.58 \\
9 & 6 & 299 & 3678 & 6 & 2480 & 36.13\\
10 & 15 & 746 & 11310 & 6 & 1083 & 104.48\\
\bottomrule
\end{tabular}
\caption{\label{table:extended example statistics}The statistics of the additional examples. From left to right: the total number of convex hulls, the total number of vertices, number of parameters in linear solve, number of ALM outer iterations, number of LM iterations, and the total computational time in minutes.}
\vspace{-10px}
\end{table}


\subsection{Baseline Comparisons}
Here we include additional comparison with three most recent single-view scene reconstruction methods, including Gen3DSR~\cite{Ardelean2025Gen3DSR}, SceneComplete~\cite{Agarwal2026scenecomplete} and MIDI~\cite{huang2025midi}. We directly use the open source implementation of the works with their default model, parameter, and weight settings, while making the following adjustments for our examples.
First, we observe that Gen3DSR and SceneComplete use their own automatic segmentation and masking pipeline through entity/background segmentation or vision language model prompt. These pipelines often suffer from segmentation failure such as missing object, mixing multiple objects into a single one, when facing the highly cluttered and occluded scene. For fair comparisons, we enhance their pipeline using the same well-segmented masks that are created and used in our pipeline with user bounding box selection. MIDI has the same user bounding box selection interface as we do, hence we direct use its native interface.
In addition, these baselines rarely take into considerations the static environment reconstruction, i.e. table-top surface in our examples, because they are not physically-aware. Gen3DSR can reconstruct environment, yet often falsely reconstructs the table as another bulky object (e.g. a huge bed) in the scene. SceneComplete does not reconstruct the environment, it only focuses on the objects in the scene. Since these two works take RGBD image as input, we can manually assign the table top surface fitted from the point cloud for them as we do for our pipeline. However, MIDI takes only RGB image as input, and it does not provide or estimate camera information. It is also unable to segment or reconstruct the table. Therefore, the reconstructed result is ambiguous in depth and object size, and we are unable to align and set up the static scene for it. Moreover, we are unable to get texture of the scene for MIDI due to computation resource limit.
We try our best to align the reconstruction result for these baselines and show them in~\prettyref{fig:baseline}. Regretfully, all of these baselines generate meshes that have severe penetrations, leading to immediate simulator blow up.

\rev{
\subsection{Ablation Study}
Here we include the ablation study on the objective terms. In \prettyref{eq:objective} defining the objective function $O(q,x)$, where term Type II is the primary term to match visual observation with large weight, Type I and III use Hausdorff/Chamfer distance for shape regularization with small, equal weight. We demonstrate the necessity of Type I and III terms using the illustration in \prettyref{fig:ablation} below.
\begin{figure}[H]
    \centering
    \includegraphics[width=0.43\linewidth]{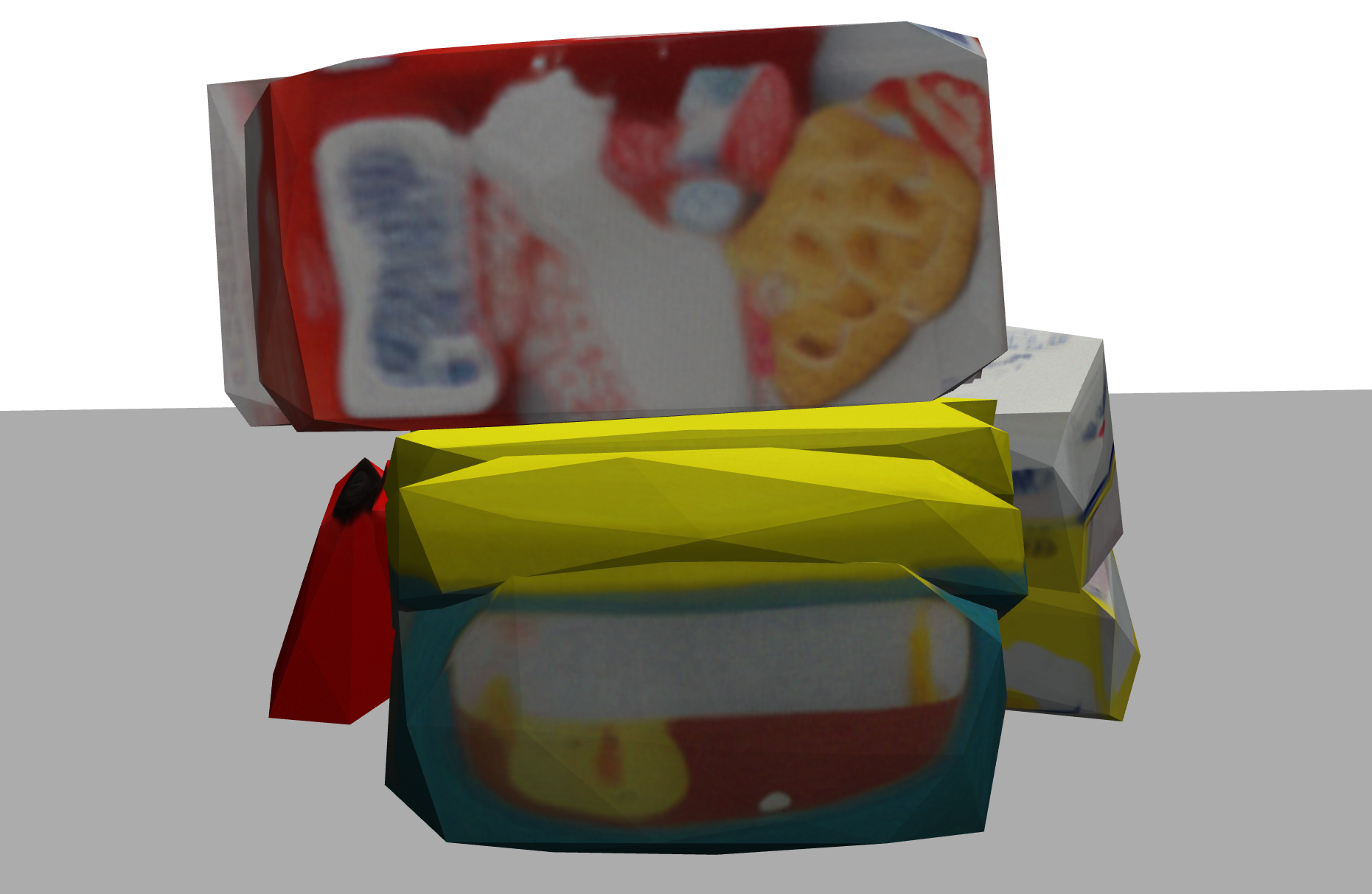}
    \hspace{2pt}
    \includegraphics[width=0.46\linewidth]{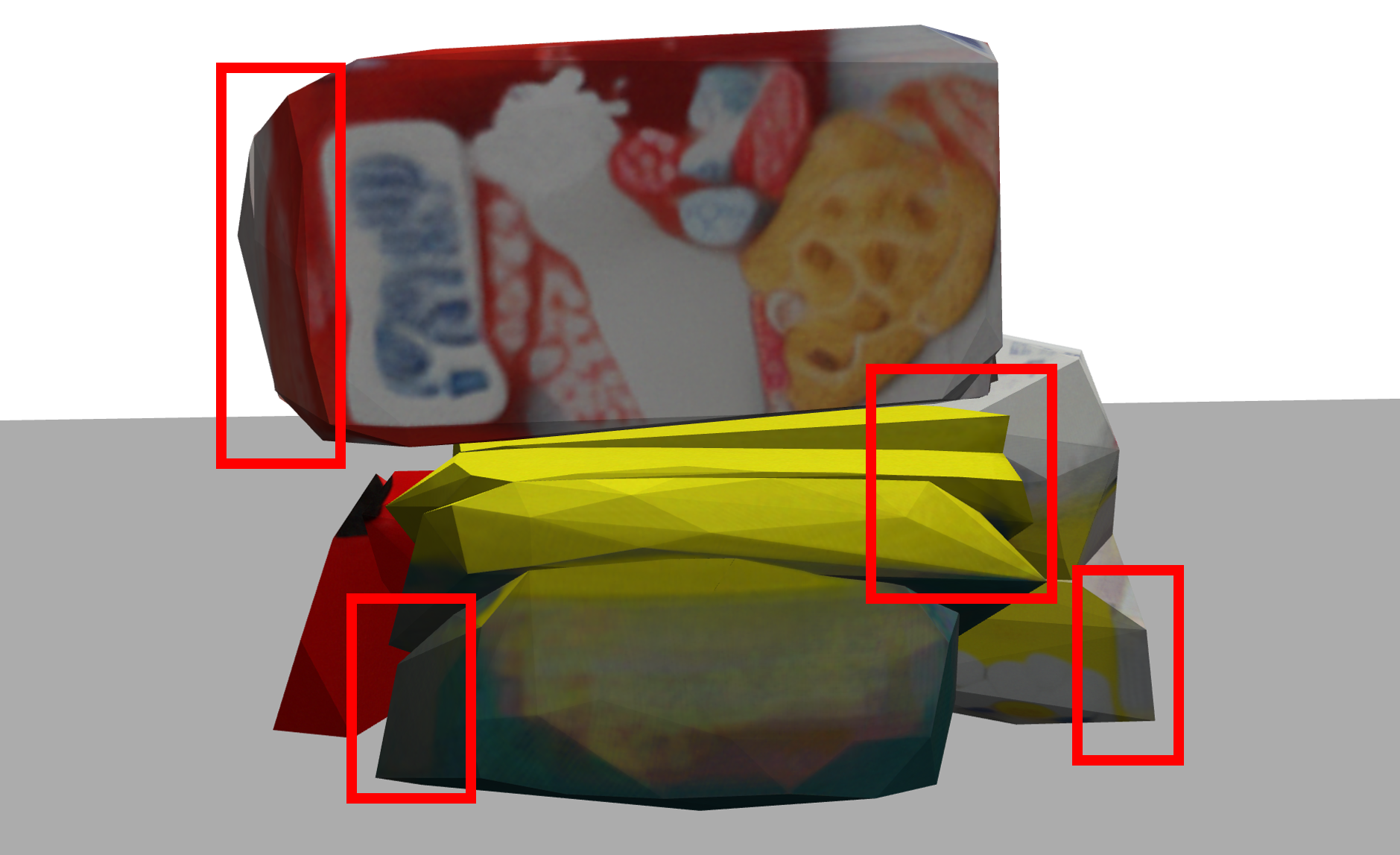}
    \vspace{-4pt}
    \includegraphics[width=0.45\linewidth]{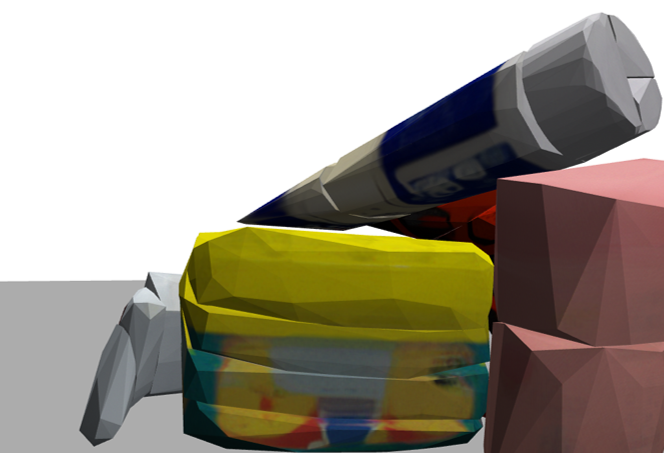}
    \hspace{2pt}
    \includegraphics[width=0.45\linewidth]{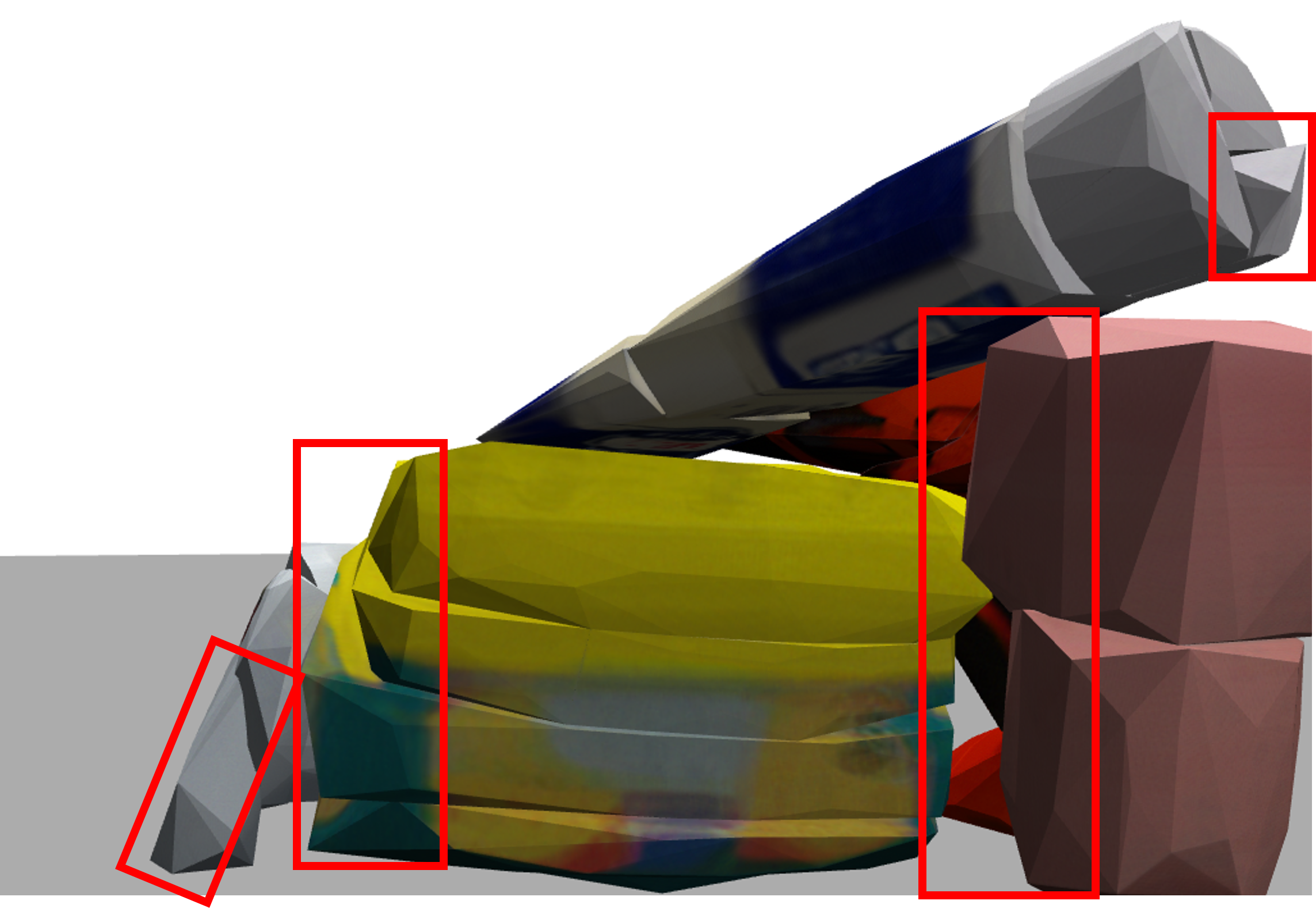}
    \caption{\rev{Shape quality for occluded part, left: w/ regularization, right: w/o regularization, showing unnatural shape result.}}
    \label{fig:ablation}
\end{figure}
Although all the results satisfy our physics constraints, shapes with regularization are more visually consistent, especially at occluded regions. Meanwhile, we keep our shape regularization while removing the primary visual matching term, and we use~\prettyref{table:ablation-objective} to show that the Type II term is necessary.
\begin{table}[H]
\centering
\ifshowchanges\color{blue}\fi
\setlength{\tabcolsep}{4pt}
\begin{tabularx}{0.49\textwidth}{c c *{5}{Y}}
\toprule
Metric & Method & Scene~$1$ & Scene~$2$ & Scene~$3$ & Scene~$4$ & Scene~$5$ \\
\midrule
\multirow{3}{*}{Average}
 & SAM3D (init.) & $2.259$      & $1.414$       & $1.940$       & $1.381$       & $1.927$ \\
 & W/ Type II    & $\E{1.978}$  & $\E{0.967}$   & $\E{1.201}$   & $\E{1.046}$   & $\E{1.651}$ \\
 & W/o Type II   & $3.821$      & $1.887$       & $2.751$       & $3.026$       & $5.981$ \\
\midrule
\multirow{3}{*}{$95\%$ quantile}
 & SAM3D (init.) & $7.459$      & $4.998$       & $6.592$       & $4.636$       & $6.619$ \\
 & W/ Type II    & $\E{6.298}$  & $\E{2.875}$   & $\E{3.265}$   & $\E{3.190}$   & $\E{5.728}$ \\
 & W/o Type II   & $10.923$     & $5.695$       & $9.323$       & $9.490$       & $14.563$ \\
\midrule
\multirow{3}{*}{$99\%$ quantile}
 & SAM3D (init.) & $\E{8.793}$  & $7.822$       & $9.27$       & $7.991$       & $9.893$ \\
 & W/ Type II    & $10.431$      & $\E{5.649}$   & $\E{4.842}$   & $\E{5.685}$   & $\E{9.182}$ \\
 & W/o Type II   & $15.033$     & $8.704$       & $12.085$       & $12.932$       & $20.838$ \\
\bottomrule
\end{tabularx}
\caption{\label{table:ablation-objective} \rev{Whole scene point cloud matching error (mm) $\downarrow$.}}
\end{table}}